\documentclass[10pt,journal,compsoc]{IEEEtran}
\IEEEoverridecommandlockouts
\usepackage{cite}
\usepackage{amsmath,amssymb,amsfonts}
\usepackage{xfrac}
\usepackage{textcomp}
\usepackage{xcolor}
\usepackage{pifont}
\graphicspath{ {imgs/} }

\usepackage{graphicx}

\usepackage{amsmath,amssymb} % amsthm
\usepackage{rotating}

\usepackage{multirow}
\usepackage[usenames,dvipsnames]{pstricks}
\usepackage{url}
\usepackage{enumitem}

\renewcommand{\d}{\mathbf{d}}

\newcommand{\e}{\mathbf{e}}

\renewcommand{\l}{\mathbf{l}}

\newcommand{\p}{\mathbf{p}}

\renewcommand{\t}{\mathbf{t}}

\renewcommand{\u}{\mathbf{u}}

\newcommand{\x}{\mathbf{x}}

\newcommand{\C}{\mathbf{C}}
\newcommand{\D}{\mathbf{D}}

\renewcommand{\P}{\mathbf{P}}

\newcommand{\R}{\mathbf{R}}

\newcommand{\X}{\mathbf{X}}

\newcommand{\vMu}{{\boldsymbol \mu}}

\newcommand{\mA}{\mathtt{A}}
\newcommand{\mB}{\mathtt{B}}

\newcommand{\mI}{\mathtt{I}}

\newcommand{\mK}{\mathtt{K}}

\newcommand{\mM}{\mathtt{M}}

\newcommand{\mR}{\mathtt{R}}

\newcommand{\mSigma}{\mathtt{\Sigma}}

\newcommand{\Real}{\mathbb{R}}

\newcommand{\tP}{\mathcal{P}}

\newcommand{\by}{{\times}} % e.g. $m \by n$ matrix

\newcommand{\norm}[1]{\left\lVert#1\right\rVert} % norm
\newcommand{\eqdef}{\overset{\underset{\mathrm{def}}{}}{=}}

\newcommand{\Comment}[1]{}

\ifCLASSOPTIONcompsoc
\else
\usepackage{cite}
\usepackage{graphicx}
\fi

\begin{document}

\title{ToosiCubix: Monocular 3D Cuboid Labeling via Vehicle Part Annotations}

%\newcommand{\refrencedetail}[1]{
%    \IEEEauthorblockA{
%        \textit{K. N. Toosi University of Technology.} \\
%        Tehran, Iran \\
%        #1
%    }
%}
%\newcommand{\xmark}{\ding{56}}
% this is a x mark the ding macro comes from pifont packae

\author{Behrooz Nasihatkon, Hossein Resani, Amirreza Mehrzadian       

\IEEEcompsocitemizethanks{
	\IEEEcompsocthanksitem Behrooz Nasihatkon, Hossein Resani, and Amirreza Mehrzadian are with the Department of Computer Engineering, K. N. Toosi University of Technology, Tehran, Iran. \protect\\
	E-mail: nasihatkon@kntu.ac.ir, behrooz.nasihatkon@gmail.com; hossein.resani@gmail.com; a.mehrzadian@email.kntu.ac.ir, mehrzadian.amir@gmail.com
}
% \thanks{Manuscript received April 19, 2005; revised August 26, 2015.}}
}

% The paper headers
\markboth{Journal of \LaTeX\ Class Files,~Vol.~14, No.~8, August~2015}%
{Shell \MakeLowercase{\textit{et al.}}: Bare Advanced Demo of IEEEtran.cls for IEEE Computer Society Journals}

\IEEEtitleabstractindextext{%
\begin{abstract}
Many existing methods for 3D cuboid annotation of vehicles rely on expensive and carefully calibrated camera-LiDAR or stereo setups, limiting their accessibility for large-scale data collection. We introduce \emph{ToosiCubix}, a simple yet powerful approach for annotating ground-truth cuboids using only monocular images and intrinsic camera parameters. Our method requires only about 10 user clicks per vehicle, making it highly practical for adding 3D annotations to existing datasets originally collected without specialized equipment. By annotating specific features (e.g., wheels, car badge, symmetries) across different vehicle parts, we accurately estimate each vehicle’s position, orientation, and dimensions up to a scale ambiguity (8 DoF). The geometric constraints are formulated as an optimization problem, which we solve using a coordinate descent strategy, alternating between Perspective-n-Points (PnP) and least-squares subproblems. To handle common ambiguities such as scale and unobserved dimensions, we incorporate probabilistic size priors, enabling 9 DoF cuboid placements. We validate our annotations against the KITTI and Cityscapes3D datasets, demonstrating that our method offers a cost-effective and scalable solution for high-quality 3D cuboid annotation.

\end{abstract}

\begin{IEEEkeywords}
3D Bounding Box, Monocular Images, Cuboid Annotation, Vehicle
	Pose Estimation
\end{IEEEkeywords}}

\maketitle
\IEEEdisplaynontitleabstractindextext

\IEEEpeerreviewmaketitle

\ifCLASSOPTIONcompsoc
\IEEEraisesectionheading{\section{Introduction}\label{sec:introduction}}
\else
\section{Introduction}
\label{sec:introduction}
\fi

\IEEEPARstart{A}{ccurate} 3D vehicle annotation is crucial for the development of autonomous driving
and advanced driver-assistance systems (ADAS). Traditional approaches often rely on
complex setups involving multiple sensors, such as LiDAR and cameras, which require
extensive calibration and synchronization. While effective, these methods necessitate
cost-prohibitive equipment and intricate data preparation procedures. The complexity and cost
associated with these setups limit their accessibility and scalability, especially for
applications needing rapid deployment and cost efficiency.

\begin{figure}[ht]
	\centering
	\includegraphics[width=0.99\columnwidth]{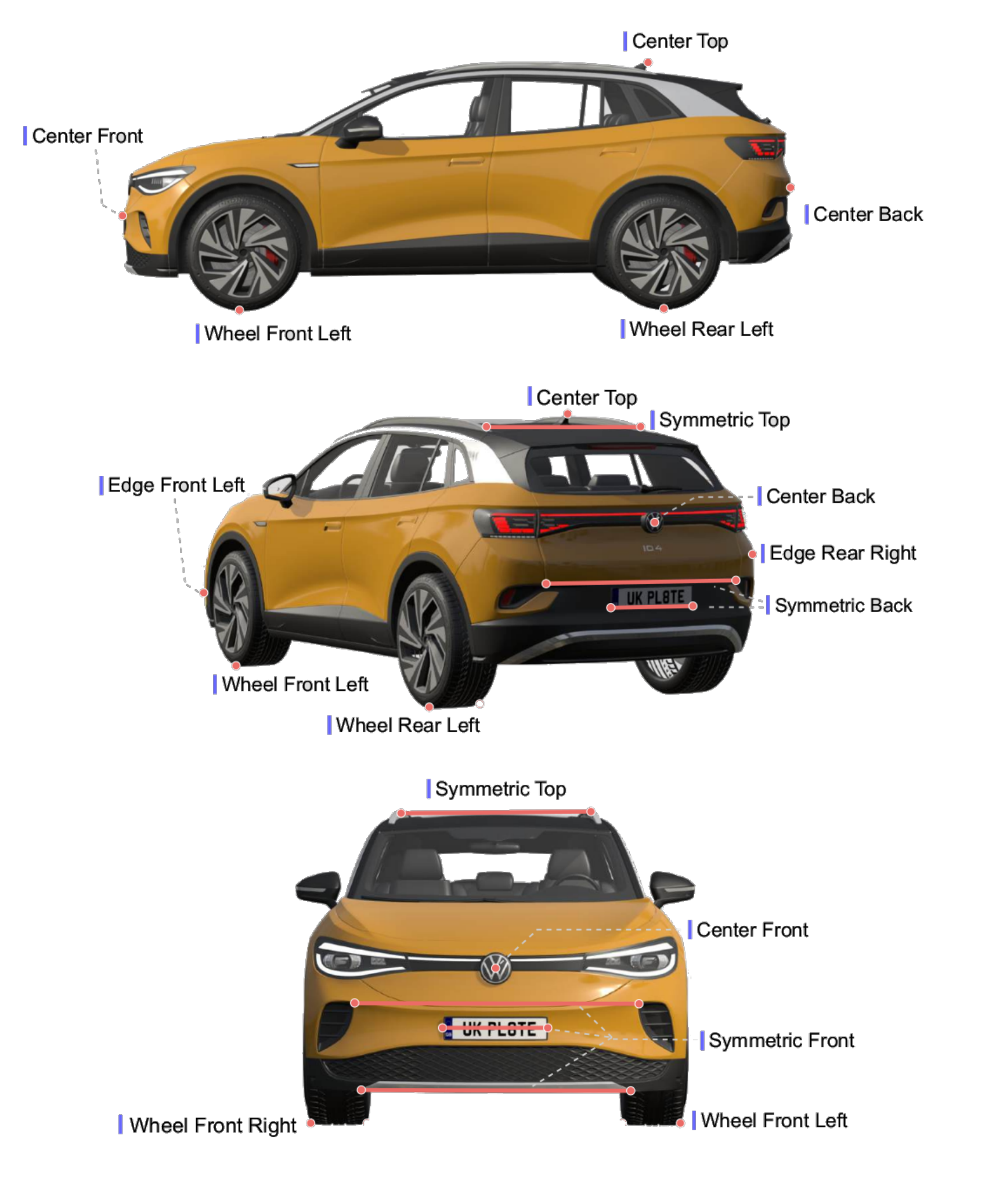}
	\caption{Examples of 2D features annotated on the vehicle.}
	\label{fig:our_model}
\end{figure}

Our motivation is to simplify the data gathering process with a single camera setup while maintaining high annotation
performance. We propose a framework that utilizes monocular data and camera intrinsics,
eliminating the need for external calibration between different sensors.
By relying solely on a single, internally calibrated camera,
our approach significantly reduces the cost and complexity of traditional multi-sensor systems.
This enables easier preparation and faster calibration, making advanced vehicle annotation more
accessible for various applications.

The simplicity and efficiency of our method streamline data collection by avoiding the need for
synchronized LiDAR and camera systems. Our framework involves annotating specific points and symmetries
on vehicles to accurately determine the location, rotation, and dimensions of the vehicle (Fig. \ref{fig:our_model}).
Using probabilistic size priors, we addresses the scale ambiguities inherent in monocular vision and resolves
ambiguities when certain vehicle dimensions are not directly observable.
We validate the effectiveness of our method through rigorous comparisons with the KITTI and Cityscapes3D
datasets, demonstrating its potential to revolutionize 3D vehicle annotation with
minimal equipment and preparation. %Our major contributions include:

% External Calibration between sensors
% LiDAR/Camera syncronization

% We do not need expesive/??? equipments for data gathering/ easy preparation/ fast calibration

% just need a single interanlly calibrated camera

% Simplifies data gathering procedures. 

% Not require beween sensor external calibration

% Does not require extra sensors for data gathering. 

\section{Related Work}
In this section, we review cuboid annotation approaches, with a focus on existing autonomous driving datasets used for monocular 3D detection task.

\subsection{Simulation-Based Annotation}
Several datasets, such as \cite{ros2016_Synthia, richter2016playing, wrenninge2018_Synscapes, cabon2020virtual, pahadia2023_skope3d}, are created using simulation environments. However, due to their synthetic nature, these datasets do not fully reflect real-world appearances of road scenes and vehicles. Nevertheless, efforts continue towards generating more realistic road scenes \cite{yang2023unisim, ljungbergh2024_Neuroncap, du2024_3Drealcar, ge2025unraveling}.

\subsection{LiDAR-Based Annotation}
The predominant approach for providing 3D cuboid annotations relies on LiDAR sensors \cite{geiger2012we, huang2018apolloscape, chang2019argoverse, kesten2019lyft, llc2019waymo, patil2019H3D, mao2021one, ye2022_Rope3d, pham20203d, alibeigi2023_Zenseact}. In these methods, monocular 3D detection data is derived by first annotating cuboids using LiDAR point clouds in the LiDAR coordinate system and then transforming them into the camera coordinate system via LiDAR-camera external calibration.

The authors of Cityscapes3D \cite{gahlert2020cityscapes} highlight several shortcomings of LiDAR-based annotations for monocular 3D detection. First, the LiDAR-camera setup requires precise calibration, which is costly and prone to errors, leading to inaccuracies in cuboid placement within the camera coordinate system. Second, synchronization issues arise due to the rotating nature of LiDAR and differences in capture frequencies between LiDAR and cameras, exacerbating errors, especially for fast-moving objects \cite{gahlert2020cityscapes,huang2018apolloscape}. Our motivation for proposing the current approach is primarily to avoid such errors, reduce costs and leverage the widespread availability of monocular data.

\subsection{Stereo Camera Setup}
Recognizing the limitations of LiDAR-based annotations, Cityscapes3D \cite{gahlert2020cityscapes} argues that cuboids obtained directly from cameras are better suited for monocular 3D detection. Their annotation process involves analyzing point clouds generated from stereo images and refining cuboid projections in the RGB image. To address the scale ambiguity, they complement the stereo data with predefined size prototypes by assigning each vehicle instance to a specific category.

In contrast, our method directly annotates cuboids from a single monocular image, eliminating errors associated with stereo calibration and expanding applicability to monocular datasets. Similar to Cityscapes3D, we utilize size priors to address scale ambiguity, but instead of fixed prototypes, we employ probabilistic priors tailored to each vehicle category.

\subsection{Monocular Pseudo-3D Annotation}
The Boxy dataset \cite{boxy2019} provides pseudo-3D annotations using a single view. Instead of 3D cuboids, vehicles are annotated with multiple quadrilaterals representing their visible sides, giving an impression of 3D structure. In contrast, our approach infers full 3D cuboids, even though annotations occur in the 2D pixel domain.

\subsection{Monocular 3D Shape Matching}
Another approach to monocular 3D detection involves aligning 3D CAD models to 2D vehicle images. In \cite{wang20183d}, the best-matching CAD model is manually selected and adjusted by annotators. The \emph{Pascal3D+} dataset \cite{xiang2014_Pascal3Dplus} refines this process by incorporating 2D landmark annotations, using a Perspective-n-Point (PnP) algorithm to estimate the camera pose and minimize reprojection errors. Similarly, the \emph{ApolloCar3D} dataset \cite{song2019Apollocar3D} utilizes 34 CAD models, each with 66 shared landmark keypoints. Annotators specify 2D image keypoints, enabling template alignment through PnP, with depth data aiding pose accuracy for ground-level keypoints.

While these methods annotate full CAD models instead of cuboids, they require precise CAD representations, making them less effective for unknown vehicle shapes. Additionally, identifying exact keypoints on vehicles can be challenging. Unlike these approaches, our method allows annotators to select flexible keypoint locations (e.g., symmetric points at the rear of a vehicle at any height), simplifying the annotation process. However, this flexibility introduces ambiguities in 3D localization that cannot be resolved with standard PnP algorithms.

\subsection{Monocular 3D Morphable Shape Matching}
This approach extends the fixed-template matching method by replacing the fixed 3D CAD models with morphable 3D shape models \cite{blanz1999morphable}, which are adapted to better fit the 2D image of the object of interest. Although we are not aware of any existing datasets that use morphable car models for generating 3D annotations, several works utilize morphable vehicle models as a means for 3D detection \cite{lin2014jointly, he2019mono3d++, shi2023optimal}, typically by estimating the locations of 2D landmark points.

While this technique differs from our approach in how 3D cuboids are derived, the underlying mathematical formulation is closely related. It can be interpreted as a generalization of the Perspective-n-Point (PnP) problem where the 3D points are no longer fixed, but rather depend linearly or affinely on a set of parameters. In morphable model fitting, these parameters control the shape; in our case, they govern the cuboid dimensions along with some auxiliary variables.

Among the works that fully adopt a projective camera model, Shi et al.~\cite{shi2023optimal} address the problem using a convex relaxation based on a second-order Lasserre’s moment relaxation, resulting in a semidefinite programming (SDP) formulation. They further provide theoretical conditions under which the relaxed solution can be certified as globally optimal.

In our specific problem setting, we observe that it is often feasible to obtain a reliable initial estimate of the rotation matrix in closed-form. Building on this, we adopt an alternating strategy: iteratively solving a least-squares problem for auxiliary variables, followed by pose estimation via the SQPnP algorithm \cite{terzakis_2020_consistently_fast_pnp}. Convergence is typically achieved within a few iterations.

\begin{figure}[ht]
	\centering
	
	\includegraphics[width=0.72\columnwidth]{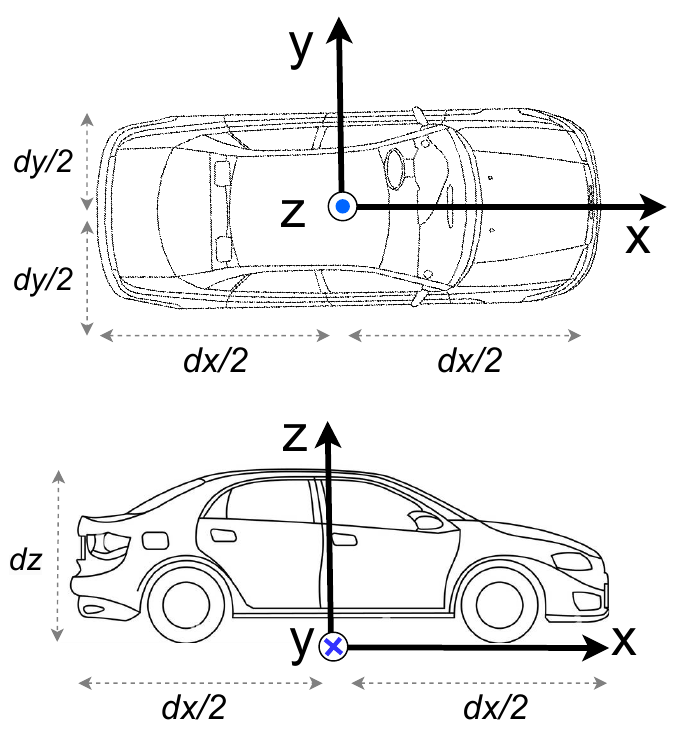}
	
	\caption{Vehicle coordinate system}
	\label{fig:vehicle_coords}
\end{figure}

\section{Method}
\label{sec:method}
We parameterize the 3D cuboid around the vehicle using the triplet  
$(\mR, \t, \d)$, where $\mR \in \Real^{3\by 3}$ and $\t \in \Real^3$ are  
the rotation matrix and the translation vector transferring vehicle  
coordinates to camera coordinates, and  
$\d = (d_x, d_y, d_z)^T \in \Real^3$ represents vehicle dimensions.  

As illustrated in Fig. \ref{fig:vehicle_coords}, we 
choose the vehicle's coordinate system such that the center of  
the coordinate system lies at the bottom center of the vehicle. The $X$  
axis points forward, the $Y$ axis points left, and the $Z$ axis points  
upward. Thus, the top-right-front corner of the cuboid has  
coordinates $(d_x/2, -d_y/2, d_z)$ in the vehicle coordinate system.

We annotate a set of 2D points $(x_i,y_i)$ on specific parts of the  
car (e.g., where the wheels touch the ground or the car emblem at the  
back or front center of the car). Different types of annotations are  
illustrated in Fig. \ref{fig:our_model} and described in the  
following subsections. Let $\X_i = (X_i,Y_i,Z_i)^T$ be the  
corresponding 3D point of $(x_i,y_i)$ in the 3D vehicle coordinate system,  
$\mK$ be the (known) camera intrinsics matrix, and  
$\u_i = \mK^{-1} (x_i,y_i,1)^T$ be the normalized image point.\footnote{If lens distortion is present, it must be accounted for when computing $\u_i$.} Then, we  
have  
\begin{align}  
\label{eq:proj}  
\lambda_i \u_i = \mR \X_i + \t,  
\end{align}  
for some depth $\lambda_i \in \Real$. While $\u_i$ is obtained through  
annotation, the corresponding 3D point $\X_i$ is not fully  
determined in vehicle coordinates. Instead, its form depends on the  
specific part of the vehicle it belongs to, imposing constraints on  
$(\mR, \t, \d)$ along with some auxiliary variables through  
\eqref{eq:proj}. Next, we introduce the different types of annotations  
and describe the corresponding form of $\X_i$ and the constraints associated with each.

\subsection{Single Point Annotations}
\label{sec:point_consts}
In this section, we discuss annotations that involve placing a single  
point on the image.  

\subsubsection{The wheels}
\label{sec:wheels}
We annotate the points where the \emph{outer} part of a wheel touches the
ground (see Fig. \ref{fig:our_model}). To obtain
$\X_i = (X_i, Y_i, Z_i)^T$, note that at these locations, we have
$Z_i = 0$. Ignoring the side mirrors, it is reasonable to assume that
the outer part of the car's wheel lies on the sides of the
cuboid. Thus, $Y_i = \pm d_y/2$. To represent $X_i$, we introduce two
auxiliary variables, $X_{wf}$ and $X_{wr}$, which denote the $X$
coordinates of the \emph{front} and \emph{rear} wheels,
respectively. Consequently, for instance, the label
\emph{Wheel-Front-Right} corresponds to
$\X_i = (X_{wf}, -d_y/2, 0)$, which, when substituted into
\eqref{eq:proj}, imposes two constraints on $(\mR, \t, \d)$ as well as
$X_{wf}$. Since determining $X_{wf}$ consumes one of these
constraints, annotating a single front wheel leaves only one remaining
constraint on $(\mR, \t, \d)$, whereas annotating both front wheels
introduces a total of three constraints. The same applies to the rear
wheels.

% \begin{itemize}
% \item \emph{Wheel-Rear-Left:} $\X_i = (X_{wr},  ~d_y/2, 0)$
% \item \emph{Wheel-Rear-Right:} $\X_i = (X_{wr}, -d_y/2, 0)$
% \item \emph{Wheel-Front-Left:} $\X_i = (X_{wf},  ~d_y/2, 0)$
% \item \emph{Wheel-Front-Right:} $\X_i = (X_{wf}, -d_y/2, 0)$
% \end{itemize}

%\subsubsection{Inner side of the wheels}
%\label{sec:wheels}
%Sometimes, only the inner side of the wheels is visible for
%annotation, while in other cases, both the outer and inner sides can
%be annotated to provide additional constraints (see
%Fig. \ref{fig:our_model}). To account for the inner side, we introduce
%an auxiliary variable $W_\text{tire}$ representing the tire width,
%allowing us to specify its location in vehicle coordinates. For
%example, the label \emph{Wheel-Front-Right-Inside} corresponds
%to $\X_i = (X_{wf}, -d_y/2 + W_\text{tire}, 0)$ in vehicle
%coordinates.

\subsubsection{Points on the centerline}
\label{sec:lateral_centers}
In many situations, we can annotate a visible mark (such as brand badges, antennas, etc.) located at the  
front-center, back-center, or top-center of the vehicle, where $Y_i = 0$ (see Fig. \ref{fig:our_model}).  
Thus, we have:  
\begin{itemize}  
	\item \emph{Front-center:} $\X_i = (d_x/2, 0, Z_i)$.  
	\item \emph{Back-center:} $\X_i = (-d_x/2, 0, Z_i)$.  
	\item \emph{Top-center:} ~~$\X_i = (X_i, 0, d_z)$.  
\end{itemize}  
Here, $X_i$ and $Z_i$ are auxiliary variables specific to each constraint. Since each of these constraints introduces a single auxiliary variable, they leave one constraint on $(\mR, \t, \d)$.

\subsubsection{Vertical Edges}
\label{sec:edges}
We may specify points along the vertical edges of the enclosing
cuboid, positioned at the rear-left, rear-right, front-left, or
front-right edges of the vehicle (see Fig. \ref{fig:our_model}).  For
instance, a point on the rear-left edge is represented as
$\X_i = \left(-\frac{d_x}{2}, \frac{d_y}{2}, Z_i\right)$, where
$Z_i$ is an auxiliary variable specific to this constraint. A single
auxiliary variable leaves one constraint on $(\mR, \t, \d)$.

\subsubsection{Corners}
\label{sec:corners}
For certain vehicles, such as container trucks and buses, some corners  
at the top of the vehicle can be marked. For example, the label  
\emph{Top-Rear-Left-Corner} corresponds to  
$\X_i = \left(\frac{d_x}{2}, -\frac{d_y}{2}, d_z\right)$,  
imposing two constraints on $(\mR, \t, \d)$ without introducing any  
auxiliary variables.

% \subsection{Paired Constraints}
% \label{sec:paired_consts}

\subsection{Left-right Symmetris}
\label{sec:sym_consts}
In many cases, we can determine pairs of points that are laterally symmetric on  
the front, back, or top faces of the vehicle. For example, symmetric points  
can be annotated on the headlights or taillights of the vehicle. This results in a  
pair of corresponding 3D points with shared auxiliary variables:  
\begin{itemize}  
	\item \textbf{Symmetry on the front:}  
	\begin{align}  
	\X_i^\text{left} &= \left(\frac{d_x}{2}, Y_i, Z_i\right), ~  
	\X_i^\text{right} = \left(\frac{d_x}{2}, -Y_i, Z_i\right)  
	\end{align}  
	
	\item \textbf{Symmetry on the back:}  
	\begin{align}  
	\X_i^\text{left} &= \left(-\frac{d_x}{2}, Y_i, Z_i\right), ~  
	\X_i^\text{right} = \left(-\frac{d_x}{2}, -Y_i, Z_i\right)  
	\end{align}  
	
	\item \textbf{Symmetry on the roof:}  
	\begin{align}  
	\X_{i}^\text{left} &= \left(X_i, Y_i, d_z\right), ~  
	\X_i^\text{right} = \left(X_i, -Y_i, d_z\right)  
	\end{align}  
\end{itemize}  
Each pair of equations introduces two auxiliary variables, leaving two constraints  
on $(\mR, \t, \d)$ per pair. Consequently, these symmetric annotations are  
preferred over their single-point counterparts in Sect. \ref{sec:lateral_centers}.

\subsection{Directions}
\label{sec:directions}
\begin{figure}
	\centering
	\begin{tabular}{cc}
		\includegraphics[width=0.48\linewidth]{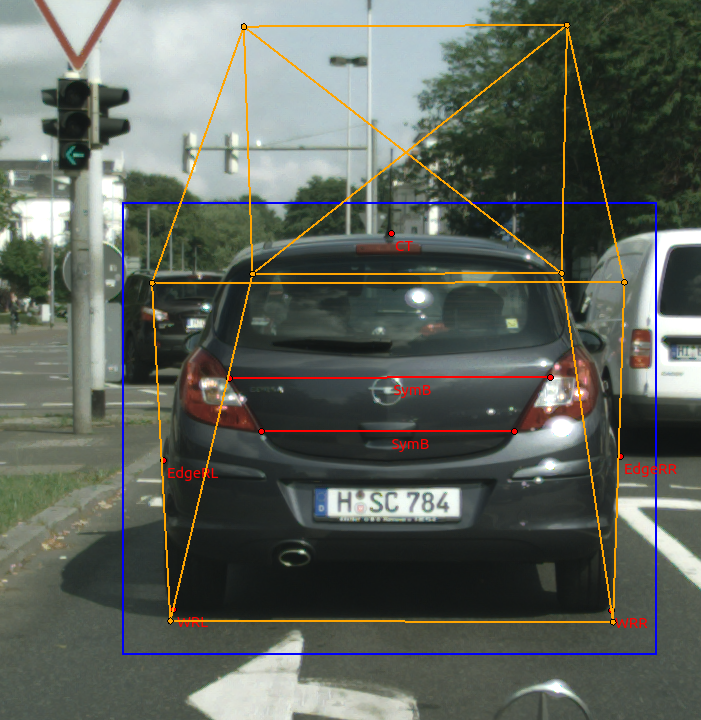}
		&
		\includegraphics[width=0.48\linewidth]{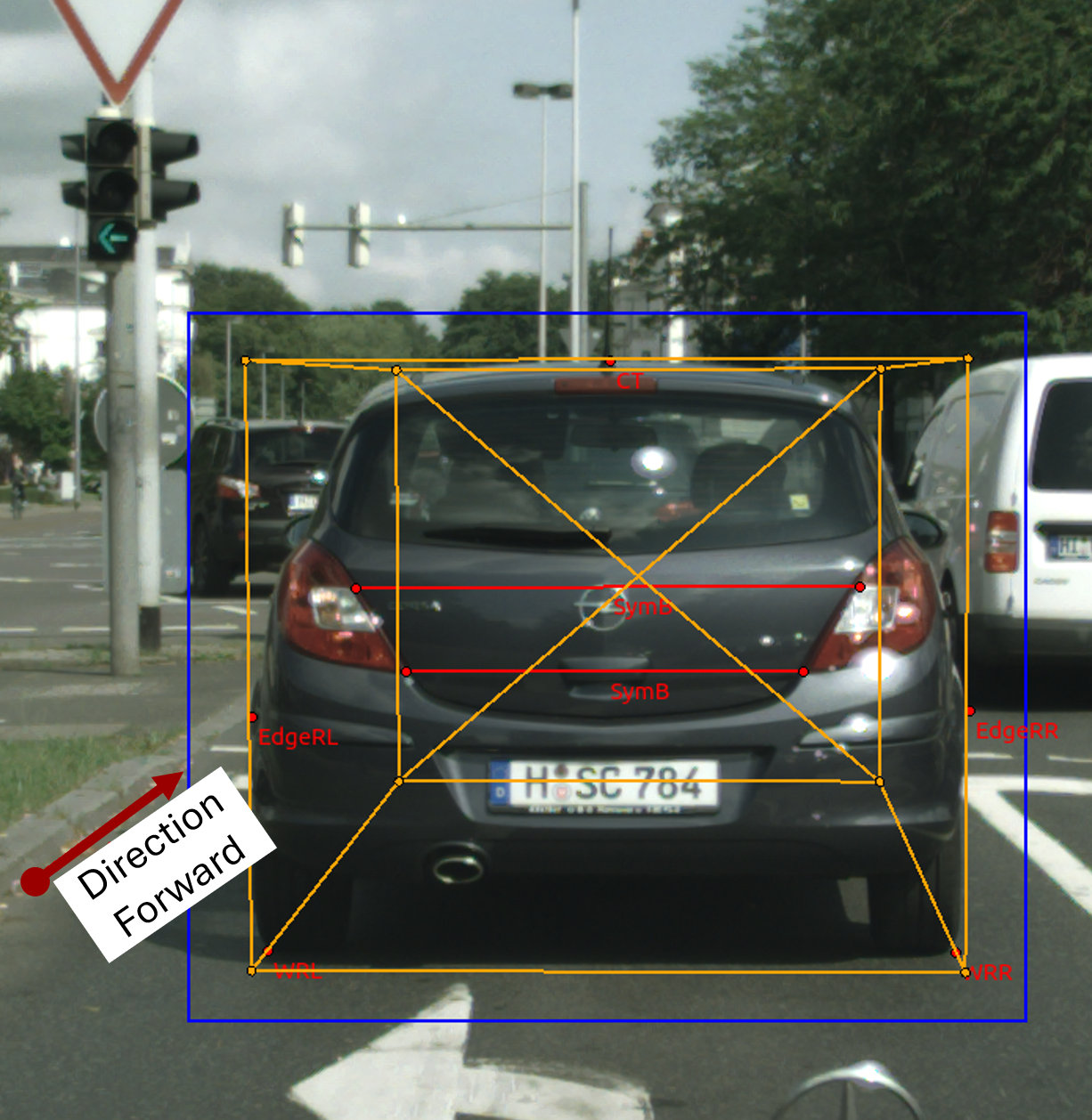}
		\\
		(a) IoU=0.37
		&
		(b) IoU=0.72			
	\end{tabular}
	\caption{Cuboid estimation from a rear view: (a) without, and (b) with directional annotation. The image is from the Cityscapes3D dataset \cite{gahlert2020cityscapes}.}
	\label{fig:withoutdir}
\end{figure}

The above constraints are sufficient for placing the cuboids in most cases. However, there are situations where 
additional geometric clues are needed. One of the most common scenarios is when the vehicle is viewed strictly from the rear or the side. Other cases include when many of the aforementioned features are occluded. To address such cases, we add annotations in the form of arrows that indicate the orientation of the vehicle. These include \emph{Forward}, \emph{Upward}, and \emph{Sideways} directions. A similar annotation was used in \cite{tao2023weakly} as a form of weak supervision for the 3D detection task. The annotations can be placed on the faces of the vehicle’s enclosing cuboid or on the ground next to the vehicle.
 An example of using a forward direction annotation on the ground is shown in Fig.~\ref{fig:withoutdir}.

\begin{figure}
	\centering
	\begin{tabular}{cc}
		\includegraphics[width=0.56\linewidth]{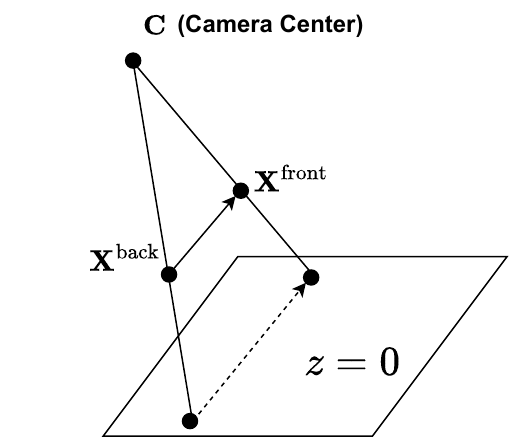}
		&
		\includegraphics[width=0.44\linewidth]{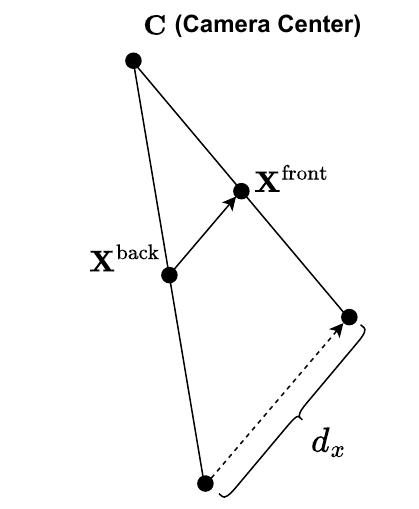}
		\\
		(a) & (b)
	\end{tabular}
	\caption{(a) Instead of parameterizing the vector from $\X_i^\text{back}$ to $\X_i^\text{front}$, we can parameterize its projection onto the ground plane ($Z = 0$). This approach fails when the plane passing through the two points and the camera center is parallel to the ground plane. (b) We can always parameterize an equivalent vector with length $d_x$, the length of the vehicle.}
	\label{fig:direction_ambg}
\end{figure}

We demonstrate how the \emph{Forward} direction is formulated within our framework. A forward-pointing vector in the vehicle coordinate system can be represented using a pair of points that share the same $Y$ and $Z$ coordinates:
\begin{align}
\X_i^\text{back} &= (X_i^\text{back}, Y_i, Z_i),~ \nonumber \\
\X_i^\text{front} &= (X_i^\text{front}, Y_i, Z_i). \label{eq:direction1}
\end{align}
However, this introduces four constraints and four auxiliary variables: $X_i^\text{front}$, $X_i^\text{back}$, $Y_i$, and $Z_i$, leaving no constraints on $(\mR,\t,\d)$. 
The key insight is that there is an ambiguity due to projective geometry. Let $\C = -\mR^{-1} \t$ be the camera center in the vehicle coordinate system. For any scalar $\alpha$, the point $\X + \alpha (\C - \X)$ lies along the line connecting $\C$ and $\X$, and therefore projects to the same image point as $\X$. Since $\X_i^\text{back}$ and $\X_i^\text{front}$ in \eqref{eq:direction1} share the same $Y$ and $Z$ coordinates, for a given $\alpha$, 
the new points $\X^\text{back} \leftarrow \X^\text{back} + \alpha (\C - \X^\text{back})$ and 
$\X^\text{front} \leftarrow \X^\text{front} + \alpha (\C - \X^\text{front})$ 
also have the same $Y$ and $Z$. Thus, we can formulate the constraints using these new points. An immediate way to reduce one auxiliary variable is to choose $\alpha$ such that the $Z$ (or the $Y$) coordinates of these new points become zero, as shown in Fig.~\ref{fig:direction_ambg}(a). Our experiments showed that this formulation makes the algorithm unstable in some situations. Note that when the plane defined by $\C$, $\X_i^\text{back}$, and $\X_i^\text{front}$ is parallel to the $XY$ plane, no choice of $\alpha$ can set the $Z$ coordinates to zero. This approach also becomes unstable when the two planes are nearly parallel. Thus, we use another alternative by choosing $\alpha$ such that the vector $\X^\text{front} - \X^\text{back}$ has a specific length (Fig.~\ref{fig:direction_ambg}(b)). Particularly, we use
\begin{align}
\X_i^\text{back}  &= (X_i^\text{back}, Y_i, Z_i),~ \nonumber \\
\X_i^\text{front} &= (X_i^\text{back} + d_x, Y_i, Z_i), \label{eq:direction_Z0}
\end{align}
thus reducing the number of auxiliary variables to 3. In the above, we chose the length of $\X^\text{front} - \X^\text{back}$ to be equal to the length of the vehicle $d_x$. This choice makes $\X_i^\text{back}$ and $\X_i^\text{front}$ linear in the dimensions and auxiliary variables, keeping the formulation consistent with other annotations. Having three auxiliary variables, \eqref{eq:direction_Z0} introduces one constraint on $(\mR,\t,\d)$.

Likewise, we can define similar annotations for the \emph{Upward} and \emph{Sideways} directions.

\subsection{Solving the equations}
\label{sec:solve}
Having $n$ point annotations, we need to solve the following set of equations:
\begin{align}
\label{eq:equations2}
\lambda_i \u_i = \mR \X_i + \t, \quad i=1,2,\ldots,n
\end{align}
for $(\mR, \t, \d)$ and the auxiliary variables. Notice that the vectors $\X_i$  
are functions of the dimension variables  
$\d = (d_x, d_y, d_z)$, the wheel locations $X_{wf}, X_{wr}$, and other  
constraint-specific auxiliary variables. From Sections  
\ref{sec:point_consts} and \ref{sec:sym_consts}, it is clear that, in all cases,
$\X_i$ is a linear function of these variables. Thus, we can rewrite  
\eqref{eq:equations2} as  
\begin{align}
\label{eq:equations_pt}
\lambda_i \u_i = \mR\, \mA_i\, \p + \t, \quad i=1,2,\ldots,n
\end{align}
where $\p \in \mathbb{R}^P$ is a vector containing all or a subset of  
the variables $d_x, d_y, d_z, X_{wf}, X_{wr}$, and the constraint-specific  
auxiliary variables, and $\mA_i$-s are $3 \times P$ matrices. Our goal is  
to solve \eqref{eq:equations_pt} for $\mR$, $\t$, and $\p$.  
Notice that the above equations are homogeneous in $(\p, \t)$,
which follows from the inherent scale ambiguity in a single-view setting.  
We will propose methods to resolve this ambiguity along with other sources  
of indeterminacy.

To solve the system \eqref{eq:equations_pt}, first observe that if $\p$ is known,  
the problem reduces to the well-studied \emph{Perspective-n-Point (PnP)} problem.  
Thus, we adopt a coordinate descent approach, alternating between estimating  
$(\mR, \t)$ and estimating $\p$ (or $\p, \t$). To ensure that both stages optimize  
a common cost function, we employ the error function used in the \emph{SQPnP} method  
\cite{terzakis_2020_consistently_fast_pnp}
\begin{align}
\mathrm{Err}(\mR,\t,\p)&=\sum_{i=1}^n \norm{\e_3^T (\mR \X_i + \t) \, \u_i  - (\mR \X_i + \t)}^2  \label{eq:sqpnp_cost} \\
&=\sum_{i=1}^n \norm{(\mI - \u_i\,\e_3^T) (\mR \X_i + \t)}^2,   \label{eq:sqpnp_cost2}
\end{align}
where $\e_3 = (0,0,1)^T$. Since the final coordinate of  
$\u_i = \mK^{-1} (x_i,y_i,1)^T$ is equal to 1,  
$\e_3^T (\mR \X_i + \t)$ corresponds to the depth $\lambda_i$ in  
\eqref{eq:equations2}. Therefore, \eqref{eq:sqpnp_cost} represents the  
error in 3D space rather than image space, as discussed in  
\cite{terzakis_2020_consistently_fast_pnp}. The SQPnP method solves  
\eqref{eq:sqpnp_cost} for $(\mR, \t)$, leaving us with the task of  
solving for $\p$ or $(\p, \t)$.  

Defining $\mM_i \eqdef \mI - \u_i\,\e_3^T$ and substituting  
$\X_i = \mA_i \p$ into \eqref{eq:sqpnp_cost2}, we obtain  
\begin{align}
\label{eq:sqpnp_cost3}
\mathrm{Err}(\mR,\t,\p) 
&= \sum_{i=1}^n \norm{\mM_i \, (\mR \mA_i \p + \t)}^2  
= \norm{\mB \p + \mM \t}^2,
\end{align}
where $\mM$ and $\mB$ are vertical concatenations of the matrices
$\mM_i$ and $\mM_i\mR\mA_i$, respectively. The above is a least
squares problem when solving for $\p$. To solve for $(\p, \t)$, as the
problem is homogeneous in these variables, a constraint must be
imposed on $\p$ alone to ensure that both stages of the coordinate
descent algorithm minimize the same cost function. Given that the
height of vehicles, $d_z$, is typically visible in driving scenarios,
we may set $d_z = 1$ to transform the problem into a standard least
squares form. Alternatively, we can solve \eqref{eq:sqpnp_cost3} in a
homogeneous manner using Singular Value Decomposition and then
renormalize the solution to enforce $\norm{\p} = 1$.

\subsection{Initialization}
\label{sec:init}
\begin{figure}[ht]
	\centering
	\includegraphics[width=0.7\columnwidth]{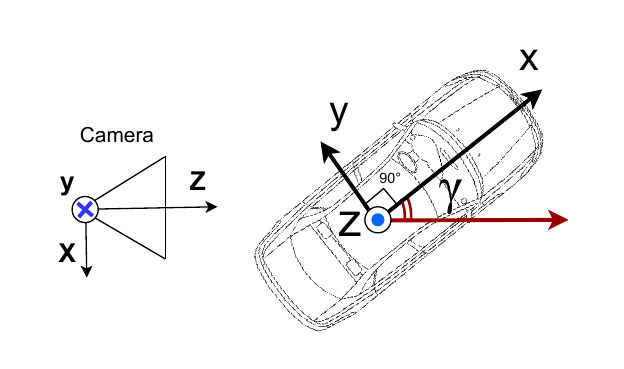}
	\caption{A bird’s-eye view illustrating the case where the vehicle has zero pitch and roll angles relative to the camera.}
	\label{fig:init_R}
\end{figure}

Our coordinate descent algorithm requires an initial value for either
$\mR$ or $\p$. Here, we choose to initialize $\mR$ by setting the
pitch and roll angles of the vehicle relative to the camera to zero
and estimating only the yaw angle $\gamma$. We assume that when the
pitch, yaw, and roll angles are all zero, the $X$, $Y$, and $Z$ axes
of the vehicle coordinate system align with the $Z$, $-X$, and $-Y$
axes of the camera coordinate system, respectively. A yaw angle of
$\gamma$ corresponds to a rotation of the vehicle by $\gamma$ radians
about its $Z$ axis, as illustrated in Fig.~\ref{fig:init_R}. Under
this assumption, the rotation matrix $\mR$ takes the form  
\begin{align}
\label{eq:init_R}
\mR = \begin{bmatrix} 
-\sin(\gamma) & -\cos(\gamma) & 0 \\ 
0 & 0 & -1 \\ 
\cos(\gamma) & -\sin(\gamma) & 0 
\end{bmatrix}.
\end{align}

To determine $\gamma$, we use line correspondences. Consider a 3D line
$\P + \alpha\, \D$ in the vehicle coordinate system defined by a point
$\P \in \mathbb{R}^3$ and a direction vector $\D \in \mathbb{R}^3$.
Let $\l \in \mathbb{P}^2$ denote the projection of the 3D line into the image’s pixel coordinate space, represented in homogeneous coordinates. Then we have the constraint \cite{xu2016_PnL}  
\begin{align}
\label{eq:line_const_rotation}
\l^T \mK \mR \,\D = 0,
\end{align}
where $\mK$ is the camera calibration matrix. This provides an equation
in terms of $\cos(\gamma)$ and $\sin(\gamma)$, as defined in \eqref{eq:init_R}.

In many cases, the direction $\D$ is known. If the front and rear
wheels on either side of the vehicle are annotated, then for the line
connecting them, we have $\D = (1,0,0)^T$. Similarly, for the line
connecting the left and right wheels of the front or back of the vehicle,
we have $\D = (0,1,0)^T$. Additionally, each of the \emph{left-right
	symmetry} constraints discussed in Sect.~\ref{sec:sym_consts} corresponds
to a line with $\D = (0,1,0)^T$. The \emph{Forward} and \emph{Sideways} directional constraints in Sect.~\ref{sec:directions} correspond to lines with  
$\D = (1,0,0)^T$ and $\D = (0,1,0)^T$, respectively.\footnote{The \emph{Upward} direction provides no information about rotation when the pitch and roll angles are zero.}
Notably, for all these cases, the third
component of $\D$ is zero. Thus, substituting these constraints into
\eqref{eq:line_const_rotation} results in equations of the form  
\begin{align}
a_i \cos(\gamma) + b_i \sin(\gamma) = 0.
\end{align}
Each individual equation provides a direct estimate via
$\tan(\gamma) = -a_i/b_i$. However, we find it more robust to consider
all available equations simultaneously. Given $m$ such equations, let
$\mA \in \mathbb{R}^{m \times 2}$ be a matrix whose rows consist of
the vectors $(a_i, b_i)$. We then compute its second (final) right
singular vector, denoted as $(u,v)$, and obtain two possible solutions
$\gamma = \tan^{-1}(v/u)$ and $\gamma = \pi + \tan^{-1}(v/u)$.  We run
the coordinate descent algorithm starting from both initial values and
select the one that yields the lower cost. Additionally, to account
for potential estimation errors in $\gamma$, we further test
$\gamma \pm 10^\circ$ and $\gamma \pm 20^\circ$ for each
case. Empirically, these adjustments suffice for convergence to a good
solution.

In very rare cases where line constraints are unavailable, we perform a brute-force
search over 36 candidate angles $\gamma = 0^\circ,10^\circ,\dots,350^\circ$
and select the angle that results in minimum cost. Nevertheless, we have not encountered such cases in our 
experiments, since the direction constraints can almost always be annotated.

\subsection{Resolving the Ambiguities}
\label{sec:ambiguities}
There are two primary types of ambiguities in this problem. The first is \emph{scale ambiguity}, as discussed earlier. In a single view setting, the dimension and location vectors $(\d, \t)$, can only be determined up to a common scale factor. This ambiguity extends to $(\p, \t)$ in our formulation. 

The second type of ambiguity arises when the length $d_x$ or width $d_y$ of the vehicle cannot be inferred. For example, when only the rear view of a car is available, there is no geometric clue to determine the length $d_x$.

Various methods have been proposed to handle scale ambiguity. One common approach is to assume a planar ground surface and use the camera's height and orientation relative to the ground to resolve the scale. However, this method can result in inaccuracies, especially for distant vehicles. An alternative approach is to ask annotators to specify the size of some \emph{features} on the vehicle’s enclosing cuboid (such as the width of the license plate). However, in some cases, it might be difficult to accurately determine those features or their size.

In our work, we leverage prior knowledge about vehicle sizes. In the Cityscapes3D dataset \cite{gahlert2020cityscapes}, prototype sizes for different vehicle categories (e.g., Sedan, Station Wagon, Mini Truck) are provided. We adopt the same prototypes as Cityscapes3D and ask annotators to assign a prototype category to each vehicle. However, instead of using a deterministic prototype size, we fit a Gaussian distribution to the vehicle dimensions for each prototype class $c$. To do this, we need to compute a mean vector $\vMu_c \in \mathbb{R}^3$ and a covariance matrix $\mSigma_c \in \mathbb{R}^{3 \times 3}$, which are estimated by computing the mean and covariance of the vehicle sizes for each  category in the Cityscapes3D training data. 

Unfortunately, for some vehicle classes, the data may not follow a Gaussian distribution due to the presence of outliers. To address this, we compute a robust estimate of $\vMu_c$ and $\mSigma_c$ using the following formulas:
\begin{align}
\label{eq:robust_mu_sigma}
\vMu_c &= \mathrm{median}(\{\d_i\}_{y_i = c}) \\
\mSigma_c &= \mathrm{median}(\{(\d_i - \vMu_c)(\d_i - \vMu_c)^T\}_{y_i = c})
\end{align}
where $y_i$ is the prototype class of the $i$-th vehicle in the Cityscapes3D training dataset, $\d_i \in \mathbb{R}^3$ is the dimension of the $i$-th vehicle, and $\mathrm{median}$ represents the geometric median \cite{drezner2002_geo_median} of a set of vectors or matrices.

Once we have the mean vector and covariance matrix for each prototype class, we incorporate a Gaussian prior into the cost function from \eqref{eq:sqpnp_cost3}, yielding the following regularized cost for a vehicle of class $c$:
\begin{align}
\label{eq:sqpnp_cost_prior} 
%\norm{\mB \p + \mM \t}^2 + \lambda (\d - \vMu_c)^T \mSigma_c^{-1} (\d - \vMu_c).
\sum_{i=1}^n \norm{\mM_i \, (\mR \mA_i \p + \t)}^2  + \lambda (\d - \vMu_c)^T \mSigma_c^{-1} (\d - \vMu_c).
\end{align}
Note that $\d$ is a subvector of $\p$. The prior term does not affect the PnP stage of the coordinate descent algorithm. In the subsequent stage for estimating $(\p, \t)$, this regularization term introduces an inhomogeneous term, which can be solved as a least-squares problem.

Choosing the appropriate value for $\lambda$ is crucial. The first term in \eqref{eq:sqpnp_cost_prior} is not scale-invariant, and it tends to shrink $(\p, \t)$. A small value for $\lambda$ may lead to underestimating the scale, whereas a large $\lambda$ will excessively rely on the prior distribution, neglecting the geometric cues from the first term.

One advantage of using a prior distribution is that it resolves not only the scale ambiguity but also the ambiguity regarding missing dimensions discussed earlier.

However, priors may introduce errors if the assumed sizes are inaccurate or if annotators incorrectly assign the prototype category for certain vehicles. In such cases, it is beneficial to allow the annotator to specify the size of certain \emph{features} on the vehicle or on the ground near the vehicle, thereby adjusting the scale of $(\p, \t)$ accordingly.

\subsection{Pixel-domain Fine-tuning}
\label{sec:pixel_domain}
As discussed in Sect. \ref{sec:solve}, the SQPnP cost defined in \eqref{eq:sqpnp_cost3} minimizes error in 3D space. We introduced a fine-tuning stage to minimize the geometric reprojection cost in the pixel domain, as the errors in point annotations typically occur in pixel space. The cost function is defined as follows:
\begin{align}
\label{eq:cost_pixel} 
\sum_{i=1}^n \norm{\x_i - \mK_{:2} \,\tP(\mR \mA_i \p + \t)}^2 + \lambda_p (\d - \vMu_c)^T \mSigma_c^{-1} (\d - \vMu_c),
\end{align}
where \( \tP: \mathbb{R}^3 \rightarrow \mathbb{R}^2 \) is the canonical projection operator, \( (X,Y,Z) \mapsto (X/Z, Y/Z) \), the matrix \( \mK_{:2} \in \mathbb{R}^{2 \times 3} \) consists of the first two rows of the camera calibration matrix \( \mK \), and \( \x_i = (x_i, y_i)^T \) represents the 2D annotated point in pixel coordinates.

We formulate the above as a nonlinear least squares problem by utilizing the Cholesky decomposition of \( \mSigma_c^{-1} \), and optimize for \( (\mR, \t, \p) \) using the Levenberg–Marquardt (LM) algorithm, initialized from the solution obtained via our coordinate-descent method. 
To update \( \mR \in \mathrm{SO}(3) \), we follow the common practice of computing the LM step in the Lie algebra tangent space at the current point, taking a step in the tangent space, and projecting back to \( \mathrm{SO}(3) \) using the exponential map \cite{eade2013gauss}.
It is important to note that, unlike the cost function in \eqref{eq:sqpnp_cost_prior}, the first term in \eqref{eq:cost_pixel} is invariant to scaling of \( (\p, \t) \).
Therefore, \( \lambda_p \) is not directly comparable to \( \lambda \) in \eqref{eq:sqpnp_cost_prior}.

\section{Evaluation}
\label{sec:eval}
We evaluate our method using two datasets: KITTI \cite{geiger2012we} and
Cityscapes3D \cite{gahlert2020cityscapes}. These datasets were chosen because they
represent two distinct modalities for generating 3D bounding boxes:
LiDAR and stereo vision, respectively. To assess the performance of
our approach, we compare our annotated 3D cuboids against the ground
truth annotations provided by these datasets. However, since both our method and the dataset annotations are subject to their own sources of error, the comparisons should be interpreted with caution; some discrepancies may stem from inaccuracies in the original dataset annotations rather than our method.

\subsection{Evaluation Metrics}
\label{sec:metrics}
Since we are dealing with an annotation task rather than a detection task, there is no matching issue—we do not need to handle many-to-one or one-to-many correspondences between detected and ground-truth objects. Therefore, object detection metrics like Average Precision (AP) or Average Orientation Similarity (AOS) may not be suitable for our evaluation. Instead, we propose the following metrics to assess the quality of 8DoF and 9DoF bounding box annotations.

\paragraph*{Intersection over Union (IoU)} 
We compute the 3D IoU \cite{ravi2020IoU3d} between the cuboid derived from our estimation \((\R, \t, \d)\) and the ground-truth cuboid \((\R_\text{gt}, \t_\text{gt}, \d_\text{gt})\), and report the average IoU across all annotated samples.

We believe that IoU alone is not an ideal metric for evaluating cuboid placement in road scenes. Notice that the translation error \(\t\) naturally grows linearly with the distance of the object, while the rotation \(\mR\) and dimensions \(\d\) remain independent of distance. Thus, even a small \emph{relative} translation error can cause the IoU to approach zero as the object moves farther from the camera, despite being acceptable in many autonomous driving applications. Another limitation of IoU is its insensitivity to orientation errors. To address these shortcomings, we also evaluate the following additional metrics.

\paragraph*{Rotation Error}
\label{sec:rot_error}
We measure the rotation error as the angular distance \cite{Hartley2013_rotation_averaging} between the estimated rotation \(\mR\) and the ground-truth rotation \(\mR_\text{gt}\):
\begin{align}
	\label{eq:rot_err}
	\text{E}_\mR = \frac{180}{\pi} \cdot \angle(\mR, \mR_\text{gt}) = \frac{180}{\sqrt{2} \pi } \norm{\log(\mR \mR_\text{gt}^{-1})}_F
\end{align}
where \(\log\) denotes the matrix logarithm and \(\norm{\cdot}_F\) is the Frobenius norm. This the angle of the relative rotation \(\mR \mR_\text{gt}^{-1}\) in axis-angle form. We report the angle in degrees for better interpretability.
 The maximum error, \(\text{E}_\mR = 180^\circ\), can occur, for example, when a vehicle moving away is estimated as moving toward.

\paragraph*{Relative Translation and Dimensions Errors} 
We report the relative translation error as:
\begin{align}
	\label{eq:trans_err}
	\text{E}_\t = {\norm{\t_\text{gt} - \t}}/{\norm{\t_\text{gt}}},
\end{align}
which is particularly meaningful for autonomous driving scenarios, where relative positional accuracy is crucial. Similarly, we compute the relative error for dimensions as:
\begin{align}
	\label{eq:dim_err}
	\text{E}_\d = {\norm{\d_\text{gt} - \d}}/{\norm{\d_\text{gt}}}.
\end{align}

\paragraph*{Combined Error}
\label{sec:com_err}
To provide a comprehensive evaluation, we define a combined error metric that averages the relative translation error, relative dimensions error, and rotation error:
\begin{align}
	\text{E}_\text{comb} = \frac{1}{3} (\text{E}_\t + \text{E}_\d + \text{E}_\mR/180).
\end{align}
This combined error balances the contributions of translation, dimensions, and rotation components, providing a holistic measure of annotation quality.

\paragraph*{Scaled IoU (sIoU)}
\label{sec:sIoU}
To assess 8DoF cuboids (translation and dimensions up to scale, plus rotation) while accounting for scale ambiguity, we compute the IoU between the scaled solution \((\R, s^* \t, s^* \d)\) and the ground truth \((\R_\text{gt}, \t_\text{gt}, \d_\text{gt})\), where the scaling factor $s^* = {\norm{\t_\text{gt}}}/{\norm{\t}}$.
%We also report a scaled version of the dimensions error:
%\begin{align}
%	\label{eq:dim_err_scaled}
%	\text{E}_\d^\text{scaled} = \frac{\norm{\d_\text{gt} - s^* \d}}{\norm{\d_\text{gt}}}.
%\end{align}

\subsection{Priors and Hyperparameters}
We obtain the priors $\vMu_c$ and $\mSigma_c$ from the training set of the Cityscapes3D dataset, as described in Sect.~\ref{sec:ambiguities}. 
The optimal values for the hyperparameters $\lambda$ and $\lambda_p$ are selected by annotating cuboids for 57 vehicles in the Cityscapes3D training set, 
and performing a search to find the values that yield the highest \emph{sIoU}. 

\subsection{Labeling Procedure}
\label{sec:labeling_procedue}
To obtain 3D annotations, we first select an image from either the KITTI or the Cityscapes3D dataset. 
Next, we identify vehicles for which ground truth cuboids are available. Vehicles are excluded if sufficient features 
for labeling cannot be identified due to occlusion, truncation, or excessive distance.

For each selected vehicle, the user draws a 2D bounding box and selects a vehicle prototype as described in Sect.~\ref{sec:ambiguities}. 
For Cityscapes3D, we use the prototypes provided by the dataset unless the annotator believes another category better matches the vehicle. 
The 2D bounding box is used only to retrieve the ground truth annotations and is not involved in estimating the 3D cuboid. 
Subsequently, 2D vehicle part annotations are added one by one until the estimated cuboid appears visually reasonable—without looking at 
metrics such as IoU or sIoU. The software measures the annotation time from the moment the user begins drawing the 2D bounding box until the final 2D feature is annotated.

 \subsection{Results}
 \label{sec:results}
 To evaluate our approach, we estimate 3D bounding boxes for 145 and 135 vehicles (across 81 and 76 images) from the KITTI dataset 
 and the Cityscapes3D validation dataset, respectively. We exclude images from the Cityscapes3D training set, 
 as they were used to obtain $\vMu_c$, $\mSigma_c$, $\lambda$, and $\lambda_p$. 
 
\begin{figure}
 	\begin{tabular}{cc}
 		\includegraphics[width=0.48\linewidth]{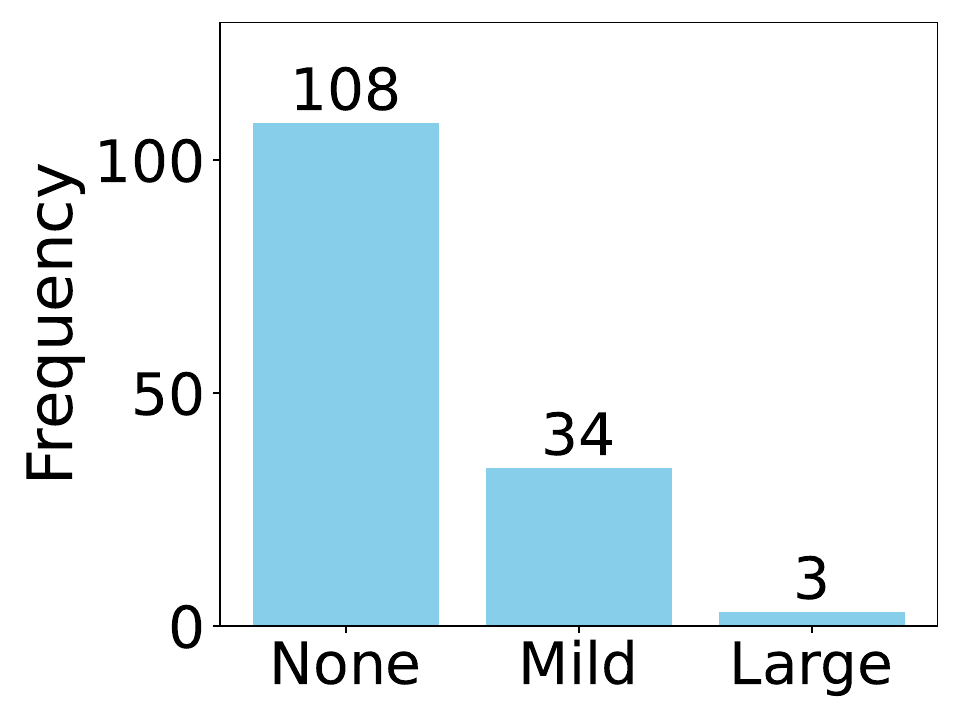}
 		&
 		\includegraphics[width=0.48\linewidth]{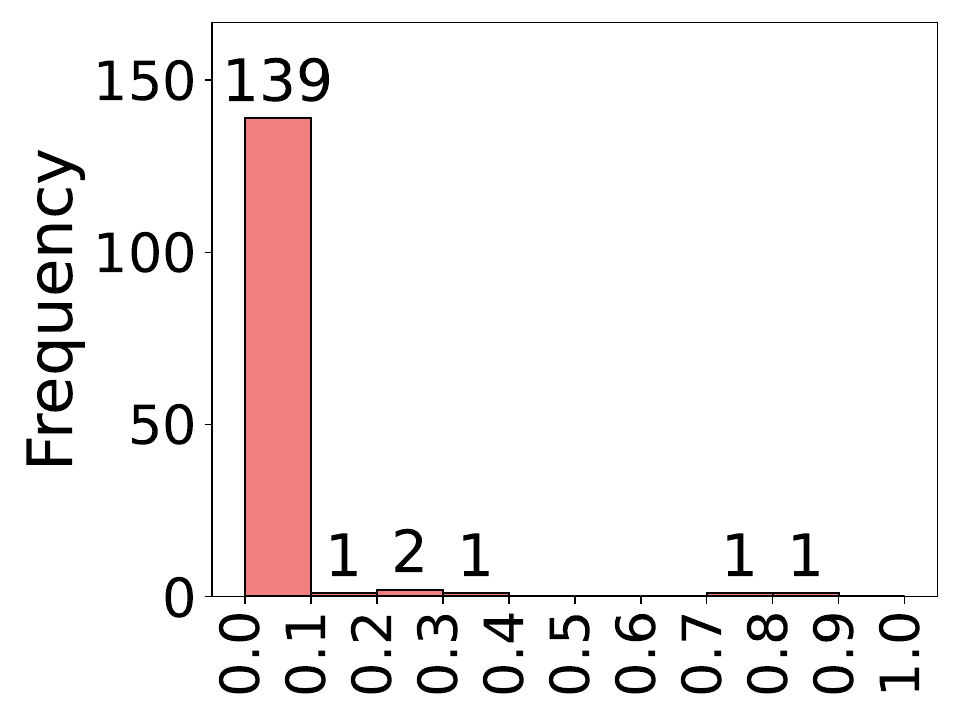}
 		\\
 		~~~~~~~~~KITTI occlusion  & ~~~~~~~~~KITTI truncation 
 		\\
 		\\
 		\includegraphics[width=0.48\linewidth]{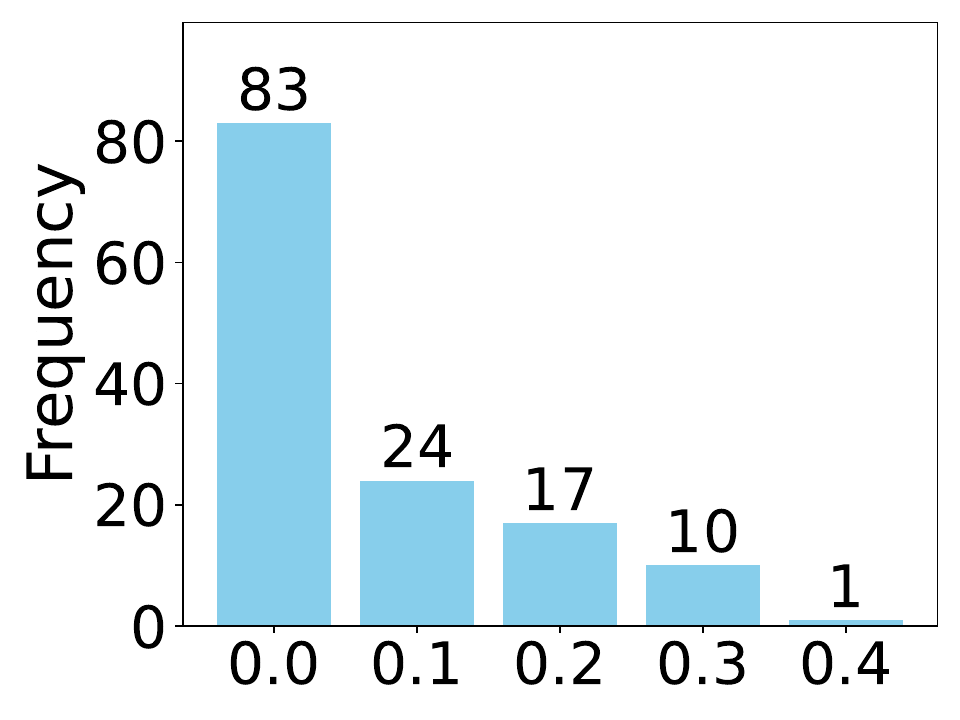}
 		&
 		\includegraphics[width=0.48\linewidth]{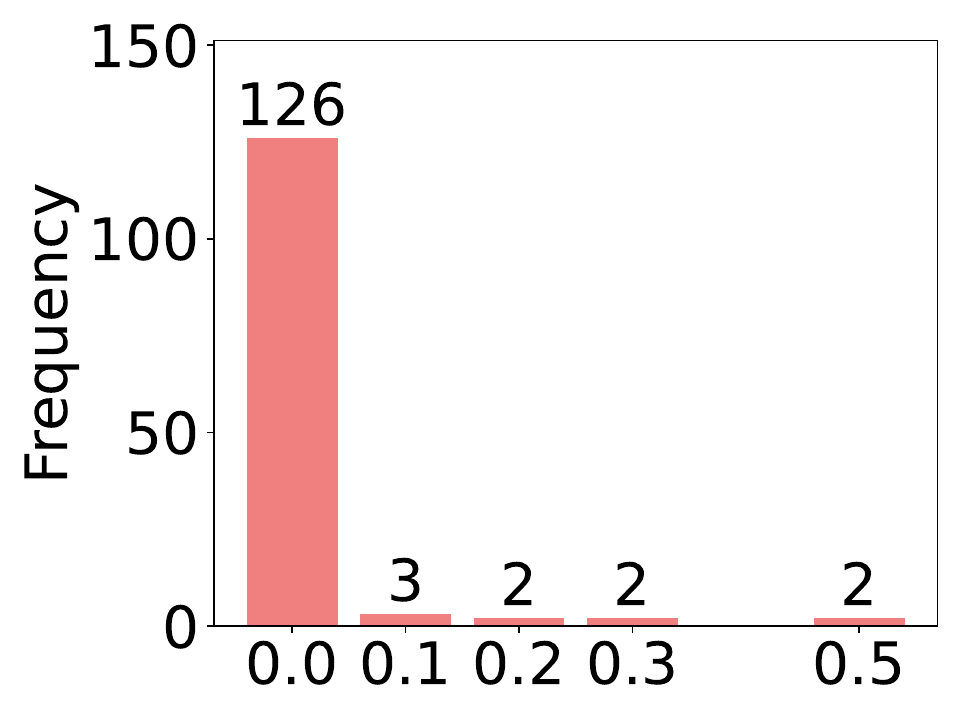}
 		\\
 		~~~~~~~~Cityscapes occlusion  & ~~~~~~~~Cityscapes truncation 
 	\end{tabular}
 	\caption{Distribution of labeled objects according to occlusion and truncation levels.\label{fig:occ_distro}}
 \end{figure}

\begin{figure}
	\centering
	\begin{tabular}{cc}
		\includegraphics[width=0.48\linewidth]{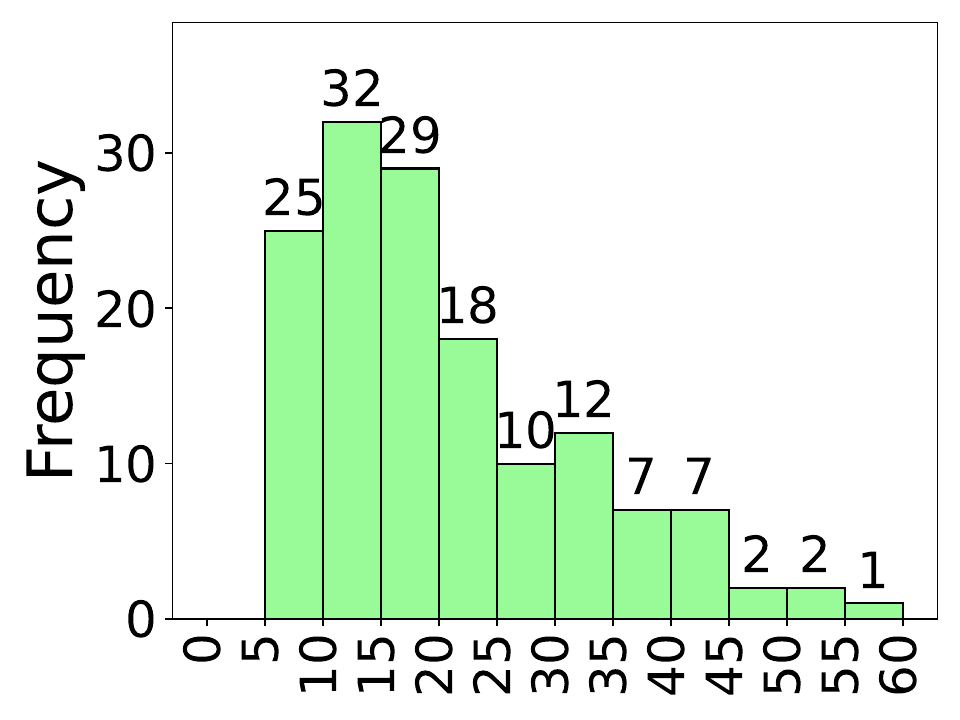} &
		\includegraphics[width=0.48\linewidth]{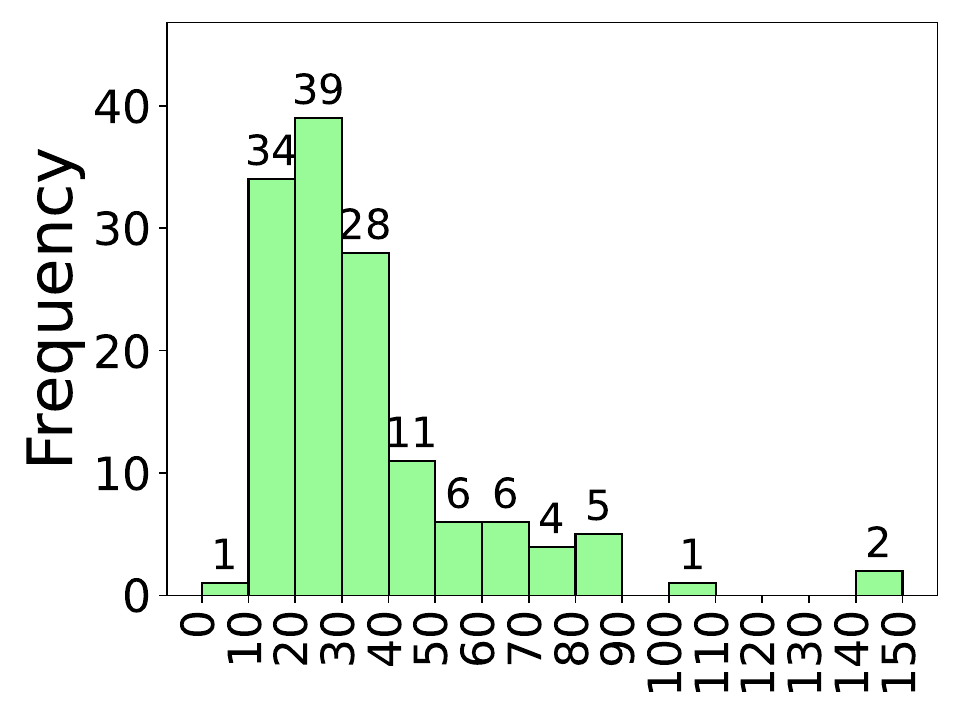} \\
		\small KITTI depth (meters) & \small Cityscapes depth (meters)
	\end{tabular}
	\caption{Distribution of labeled vehicles according to their depth.\label{fig:depth_distro}}
\end{figure}

 Figure~\ref{fig:occ_distro} shows the distribution of the labeled objects with respect to their occlusion and truncation levels. 
 Additionally, Fig.~\ref{fig:depth_distro} illustrates the distribution of the labeled vehicles based on their depths 
 (i.e., the third component of $\t_\text{gt}$).  The results are summarized in Table~\ref{tab:eval}.

 \begin{table}[h!]
 	\centering
 	\begin{tabular}{|l|cccccc|}
 		\hline
 		\textbf{Dataset / Metric} & \textbf{IoU} & \textbf{sIoU} & $\mathbf{E}_\R$ & $\mathbf{E}_\mathbf{t}$ & $\mathbf{E}_\mathbf{d}$ & $\mathbf{E}_\text{comb}$ \\
 		\hline
 		KITTI~\cite{geiger2012we}           & 0.29 & 0.77  & $3.16^\circ$ &  0.10 & 0.14 & 0.10  \\
 		Cityscapes3D~\cite{gahlert2020cityscapes} & 0.32 & 0.82  & $2.95^\circ$ &  0.06 & 0.04 & 0.05  \\
 		\hline
 	\end{tabular}
 	\caption{Evaluation metrics averaged over the KITTI and Cityscapes3D datasets.}
 	\label{tab:eval}
 \end{table}
 
 \paragraph*{8 DoF}
 \label{sec:results_8DoF}
 To assess how well our method estimates 8DoF bounding boxes (rotation, position, and dimensions up to a scale ambiguity), 
 we examine the rotation error $\mathbf{E}_\R$ and the scaled IoU (sIoU). 
 Our method performs well, achieving an average rotation error of approximately $3^\circ$ on both datasets, 
 with \emph{sIoU} scores of $0.77$ and $0.82$ for the KITTI and Cityscapes3D datasets, respectively. 
 Note that part of the error may be attributable to inaccuracies in the original ground truth annotations.
 
 \paragraph*{9 DoF}
 \label{sec:results_9DoF}
The 9 DoF task refers to full 3D pose estimation, encompassing translation, rotation, and object dimensions, and is substantially more challenging in monocular vision. In terms of IoU, our method achieves average scores of $0.29$ and $0.32$ on the KITTI and Cityscapes3D datasets, respectively. However, additional evaluation metrics offer a more encouraging perspective: we obtain a relative translation error of $\mathbf{E}_\mathbf{t} = 0.10$ and a dimension error of $\mathbf{E}_\mathbf{d} = 0.14$ on KITTI. These values improve on Cityscapes3D to $\mathbf{E}_\mathbf{t} = 0.06$ and $\mathbf{E}_\mathbf{d} = 0.04$. While these results, particularly for Cityscapes3D, may be acceptable for downstream tasks such as distance estimation in autonomous driving, they are not sufficient for accurate cuboid annotation. This indicates that size priors alone do not fully resolve the scale ambiguity inherent in monocular 9 DoF estimation. Additional geometric cues are therefore necessary to achieve high-quality annotations, as we discuss next.

 Overall, our method performs better on Cityscapes3D. This may be due to the looser bounding boxes in KITTI, 
 the assumption of zero roll and pitch angles in KITTI, and the use of dataset-provided prototype labels in Cityscapes3D.
 
 \subsection{Annotation Time}
 On average, annotating each vehicle takes 70 and 67 seconds for the KITTI and Cityscapes3D datasets, respectively. 
 Most of this time is spent selecting the correct features and assigning appropriate labels (e.g., \emph{Wheel-Rear-Left}). 
 This suggests that annotation time could be significantly reduced by incorporating semi-automatic techniques 
 and improving the annotation interface.
 
\subsection{Qualitative Assessment}
\label{sec:qualitative}
\begin{figure*}
	\centering
	\begin{tabular}{cc}
		Ours & KITTI \\
		\includegraphics[width=0.40\linewidth]{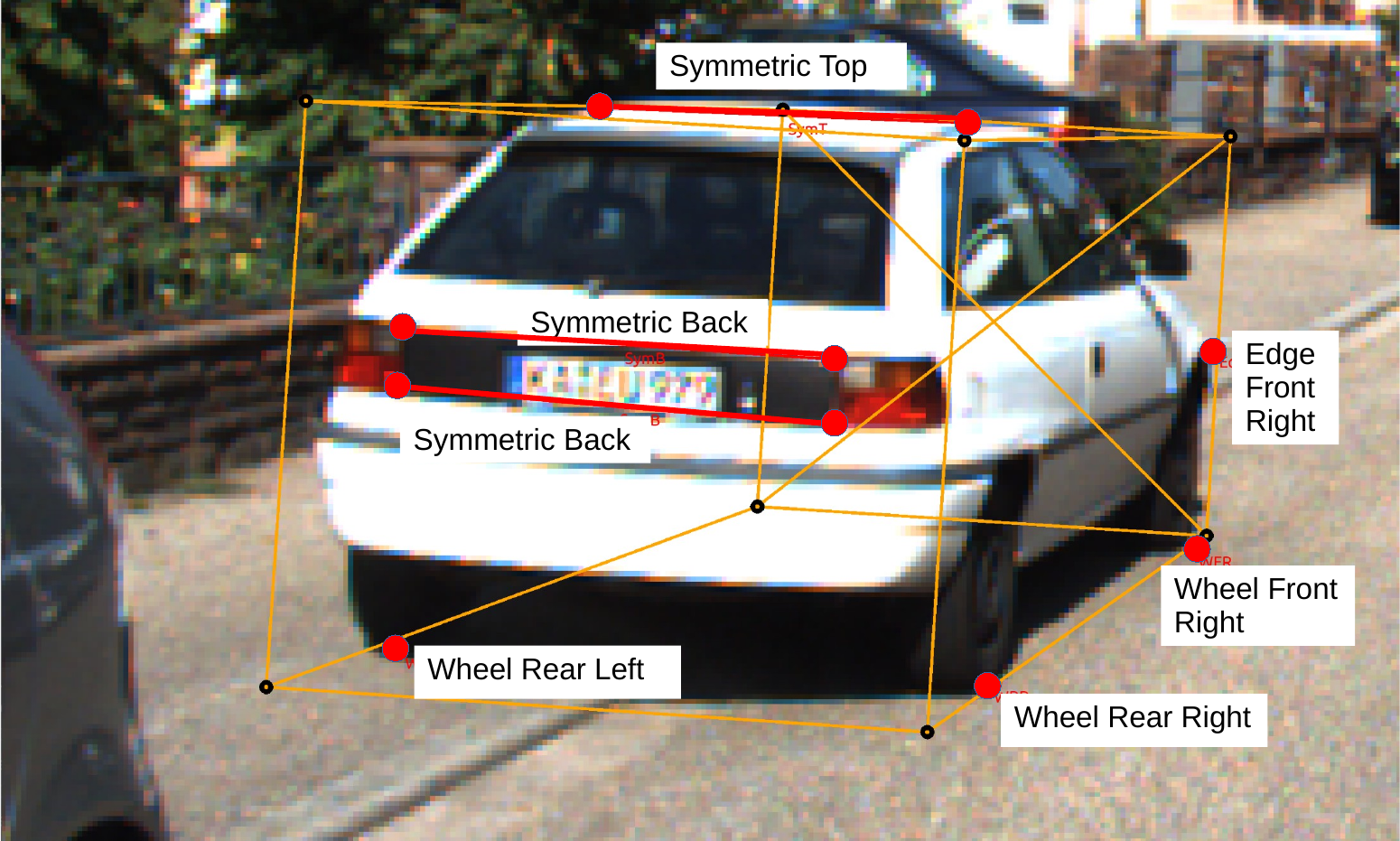} &
		\includegraphics[width=0.40\linewidth]{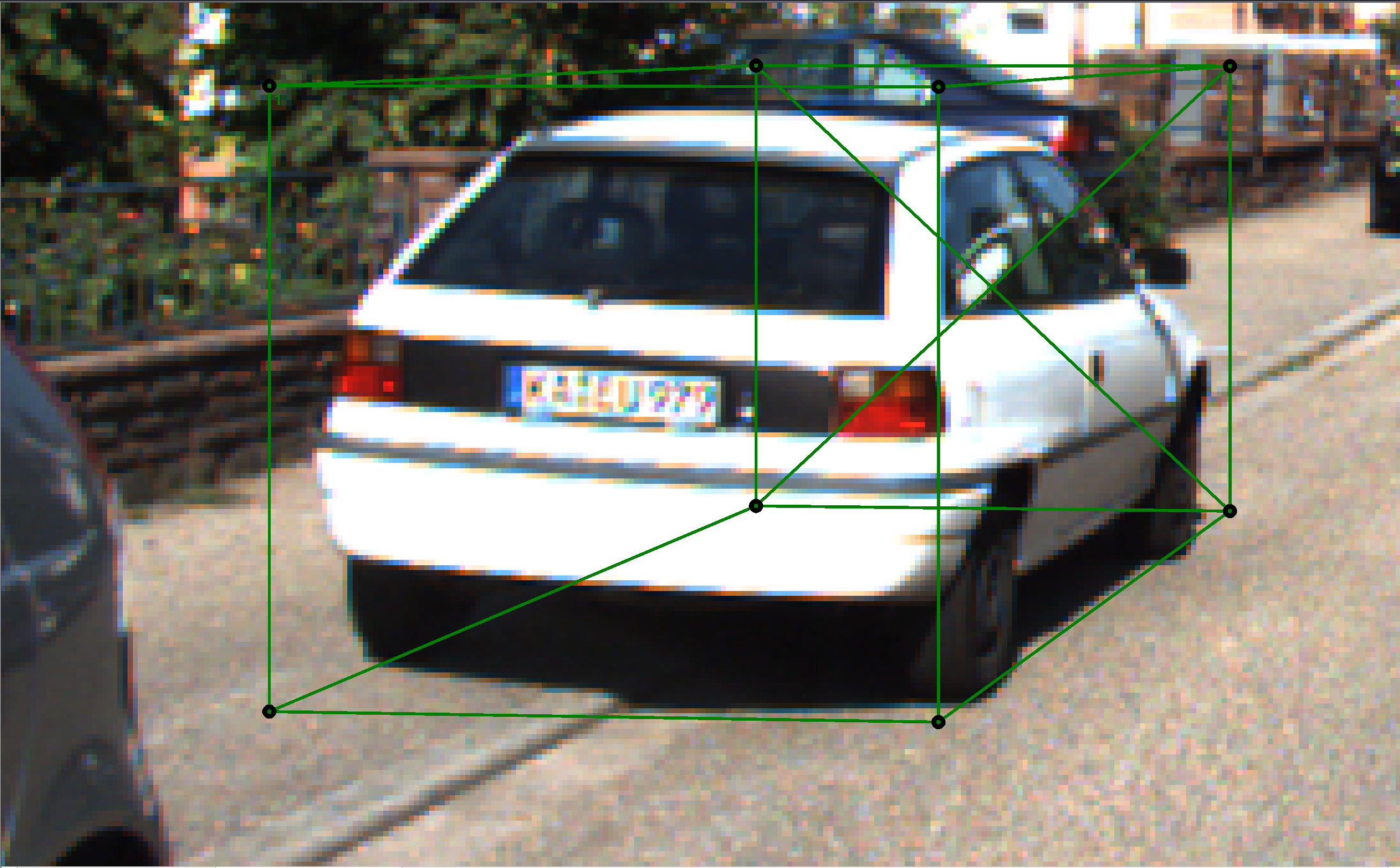} \\
		\includegraphics[width=0.40\linewidth]{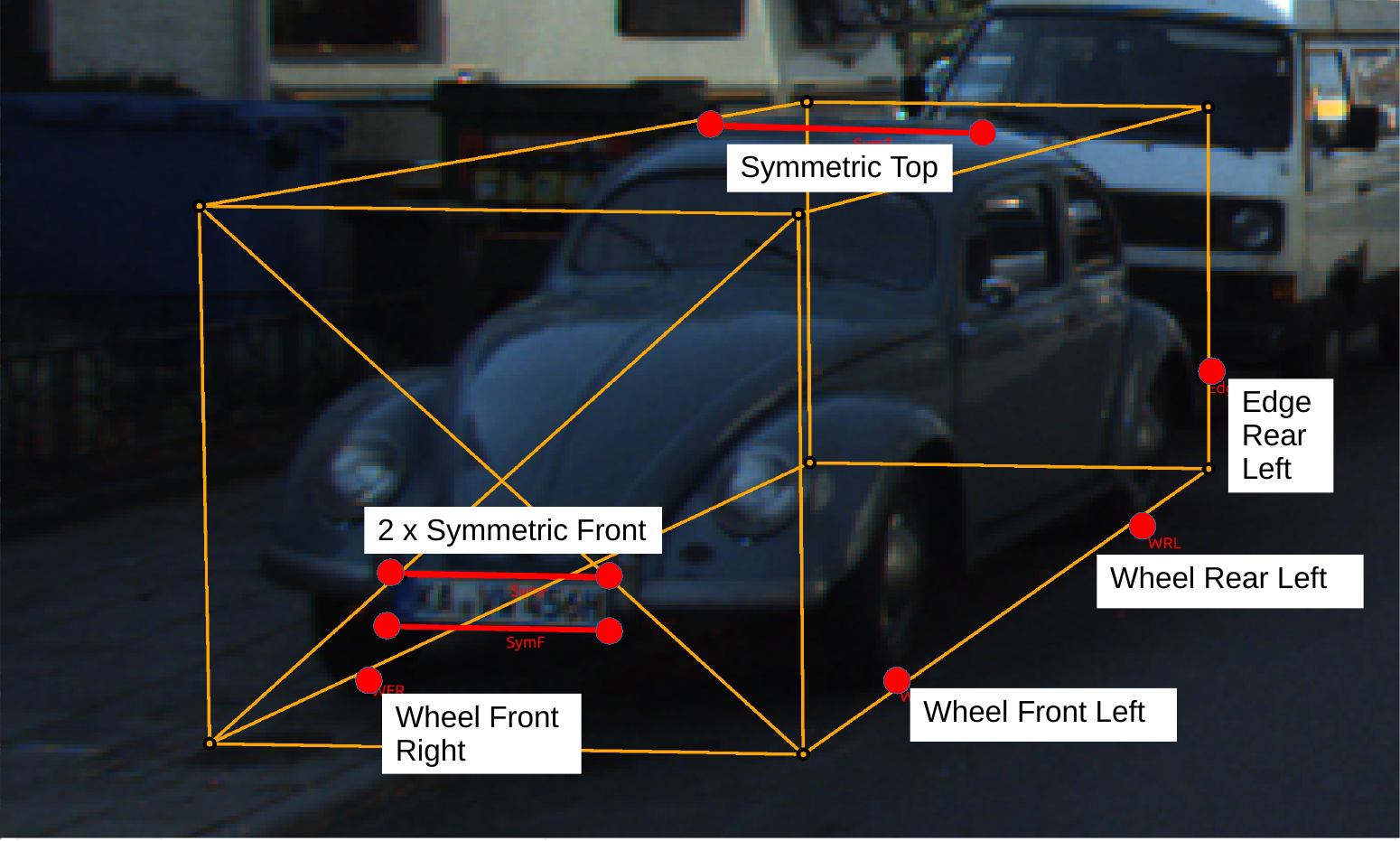} &
		\includegraphics[width=0.40\linewidth]{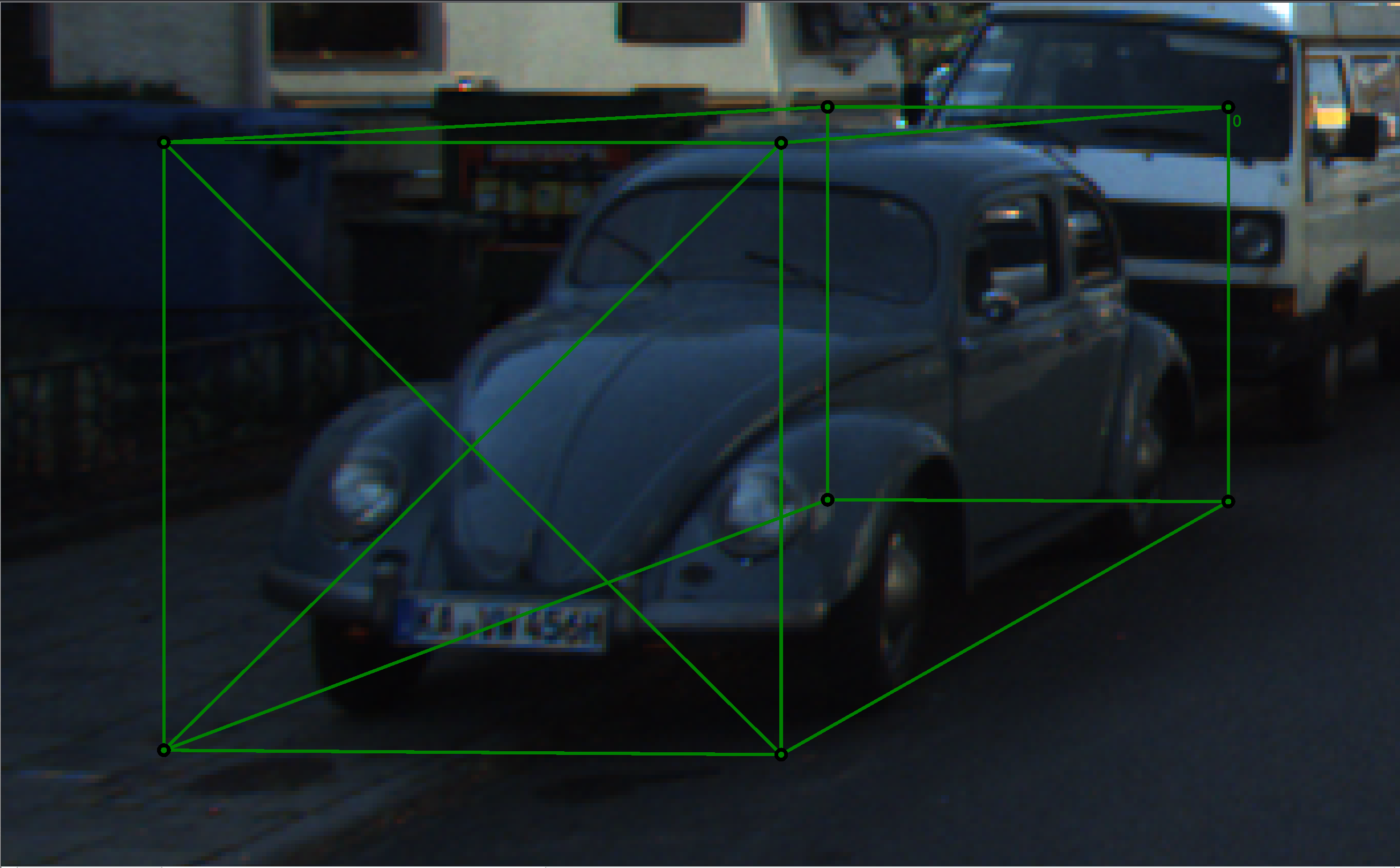} \\
		\includegraphics[width=0.40\linewidth]{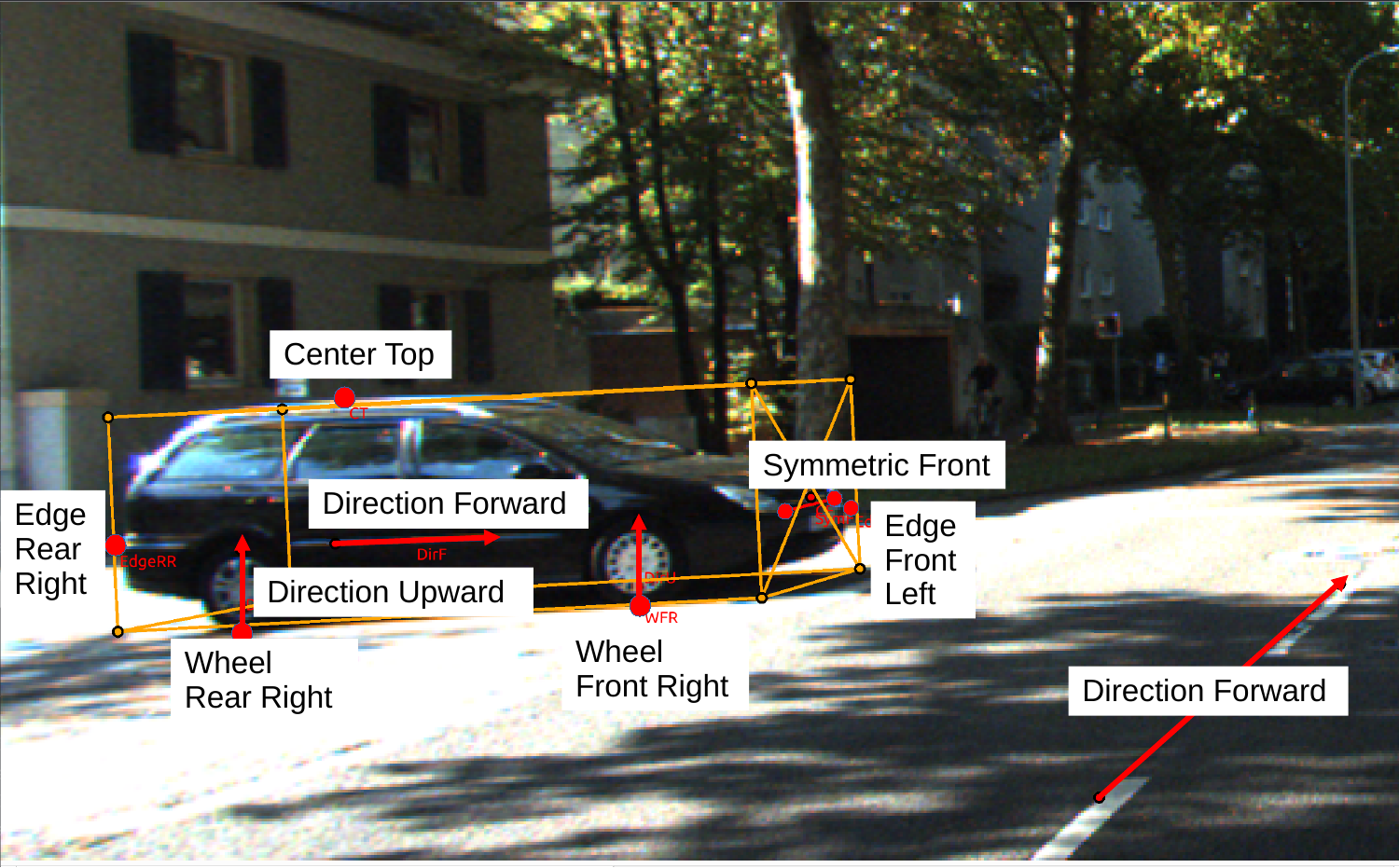} &
		\includegraphics[width=0.40\linewidth]{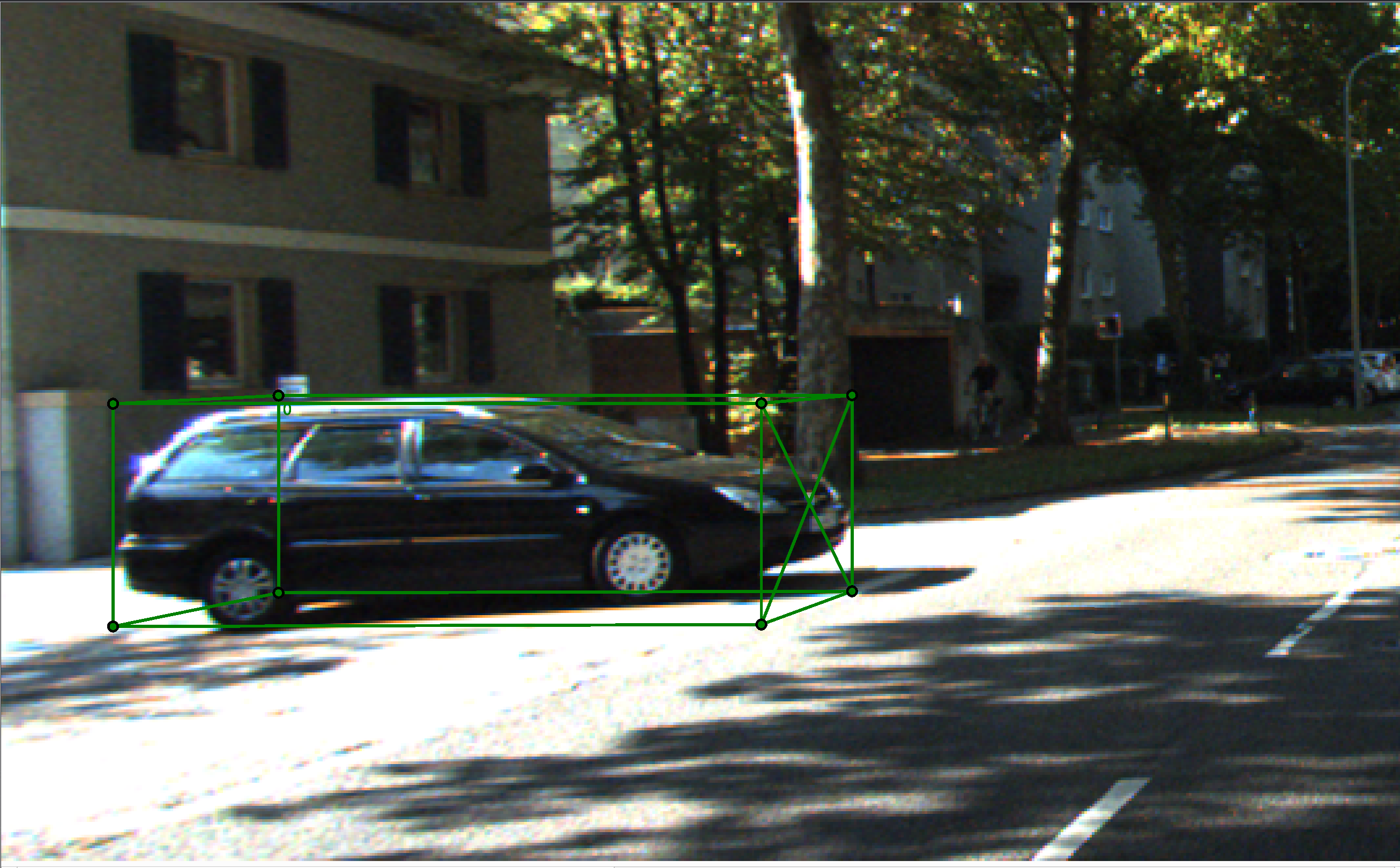} \\
		\includegraphics[width=0.40\linewidth]{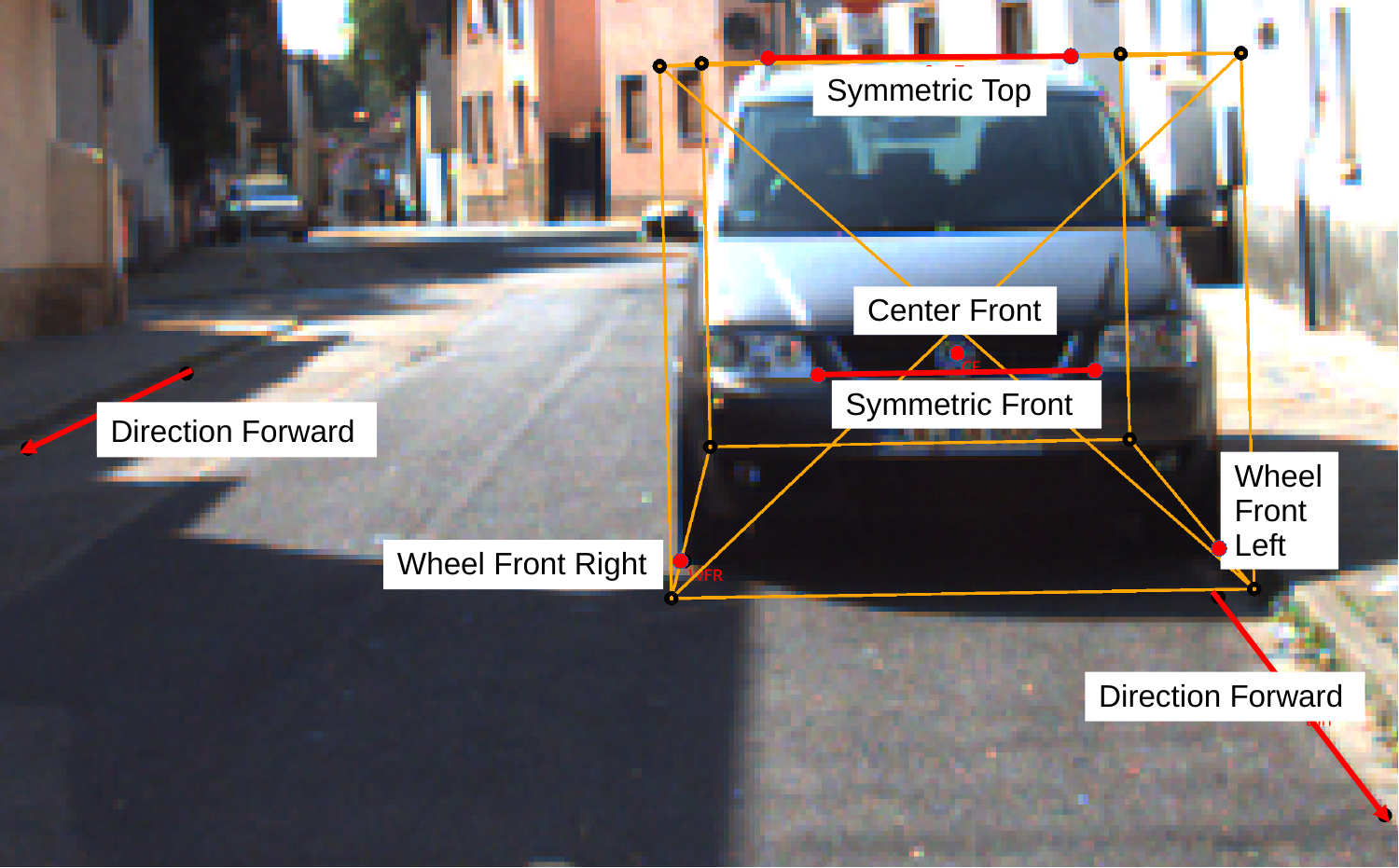} &
		\includegraphics[width=0.40\linewidth]{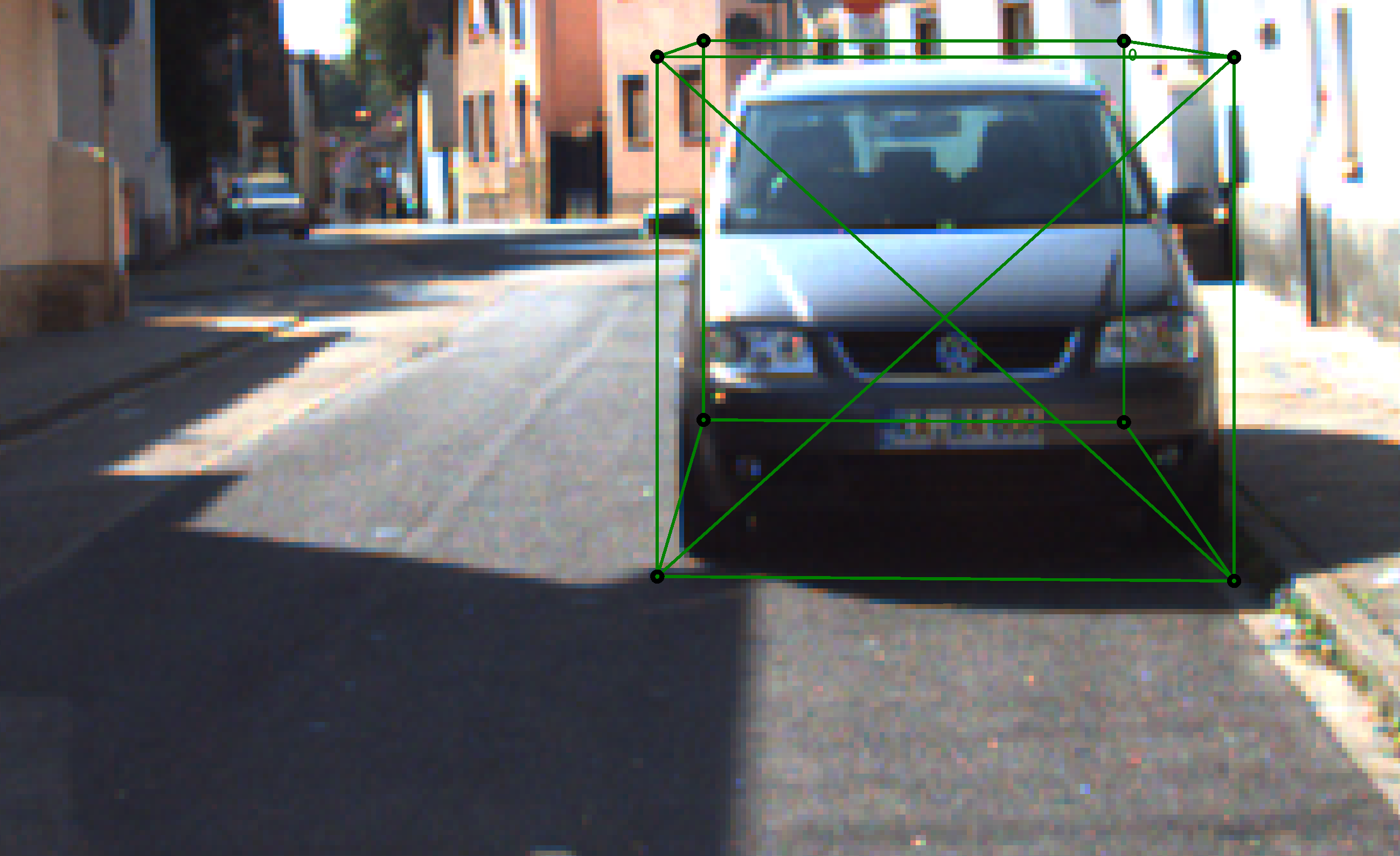} \\
		\includegraphics[width=0.40\linewidth]{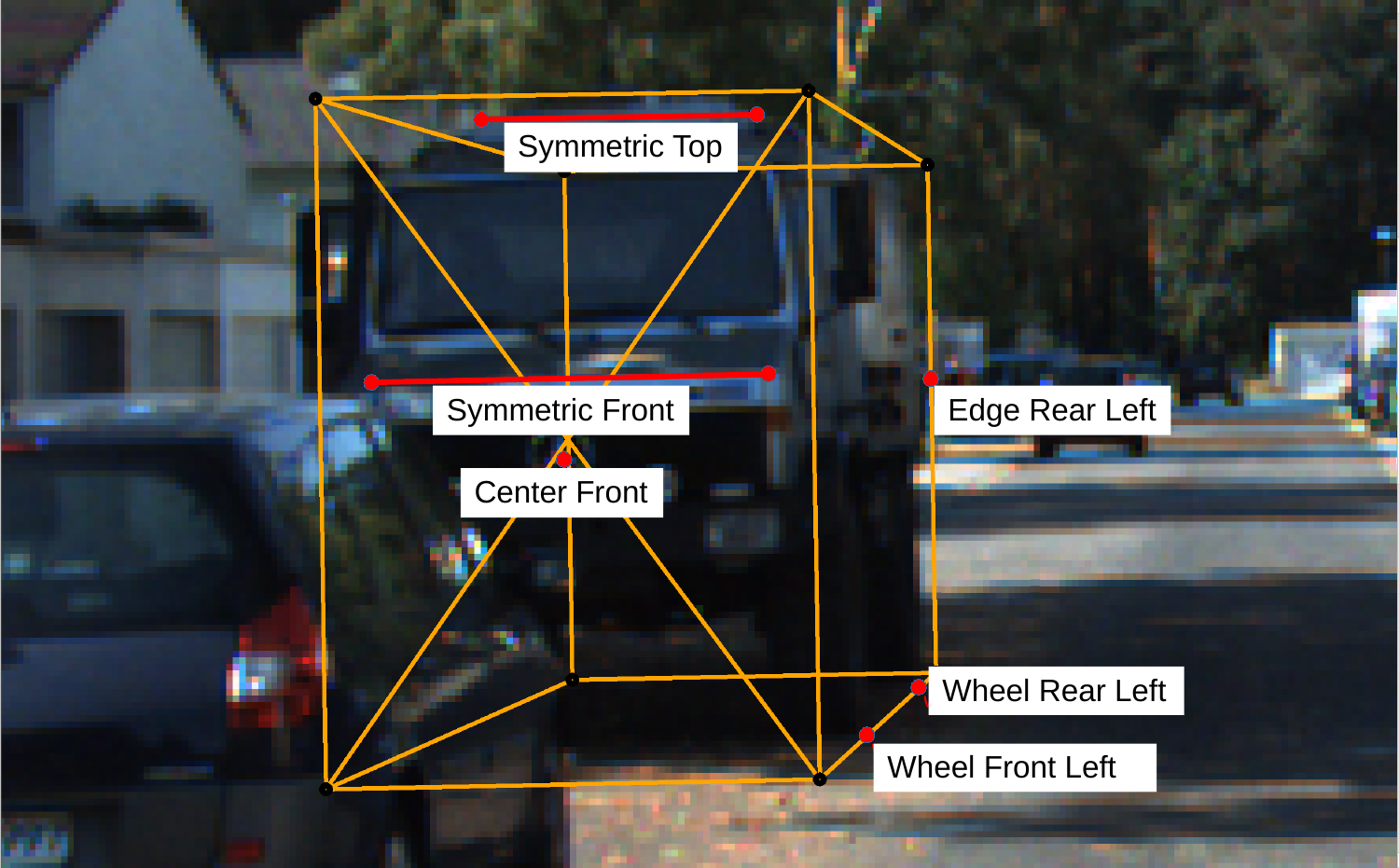} &
		\includegraphics[width=0.40\linewidth]{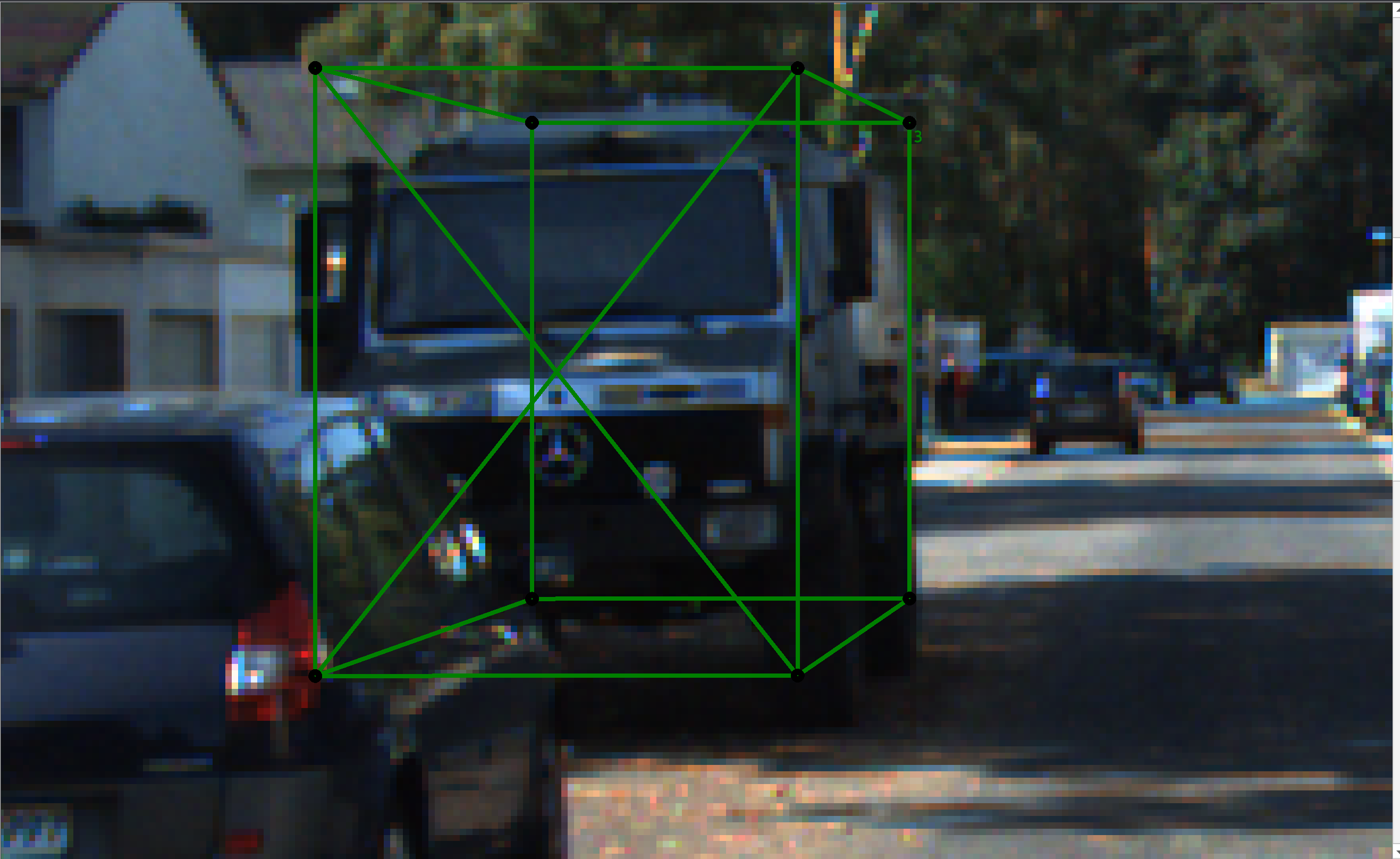}
	\end{tabular}
	\caption{Examples of cuboid annotations in the KITTI dataset. Left: Our annotations. Right: KITTI annotations. Note the 3DoF rotation estimation in our approach versus 1DoF in KITTI. Our cuboids are generally tighter. For example, the left wheels of the truck in the last row show that KITTI’s label is inaccurate.}
	\label{fig:visual_KITTI}
\end{figure*}

\begin{figure*}
	\centering
	\begin{tabular}{cc}
		Ours & Cityscapes \\
		\includegraphics[width=0.40\linewidth]{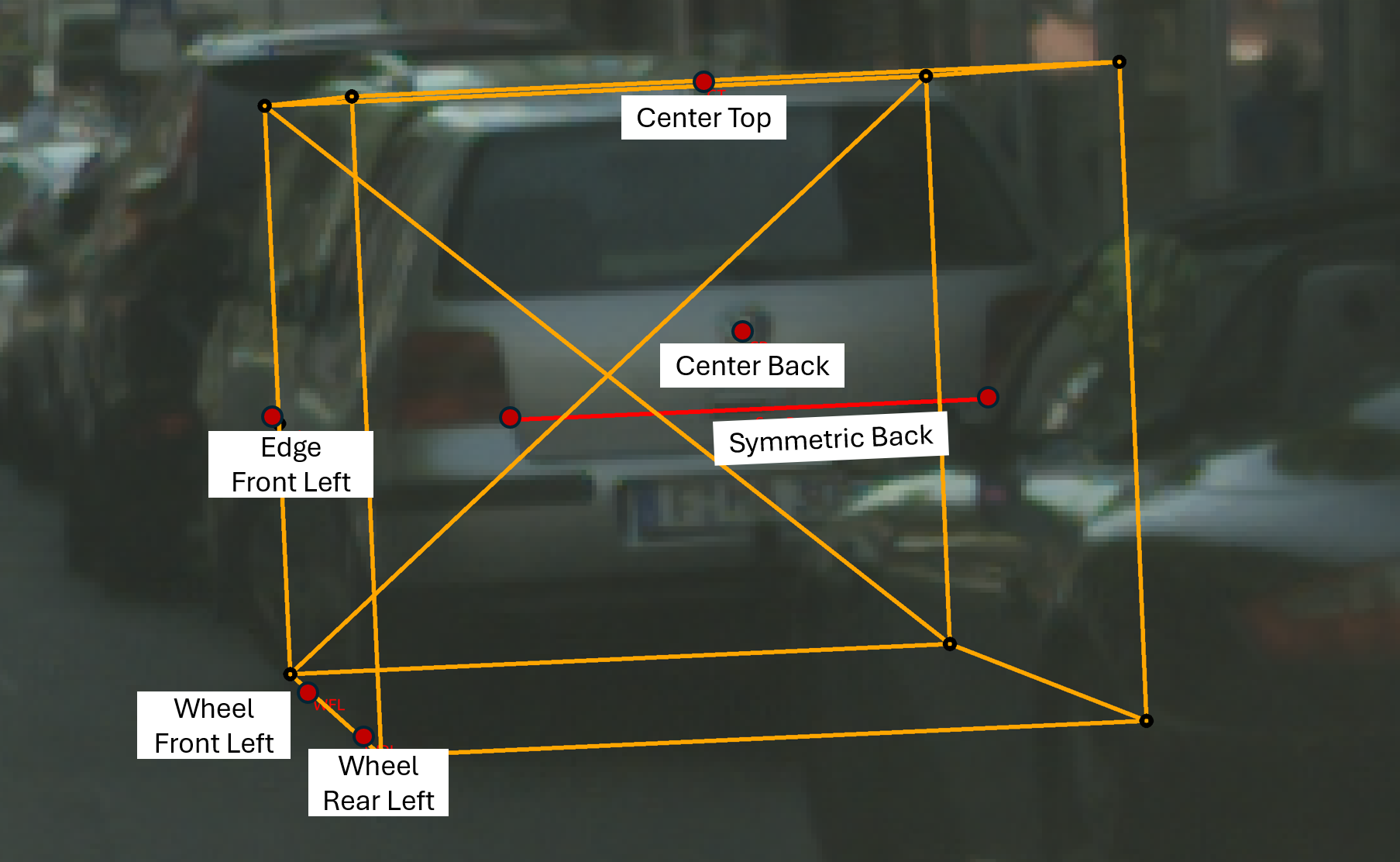} &
		\includegraphics[width=0.40\linewidth]{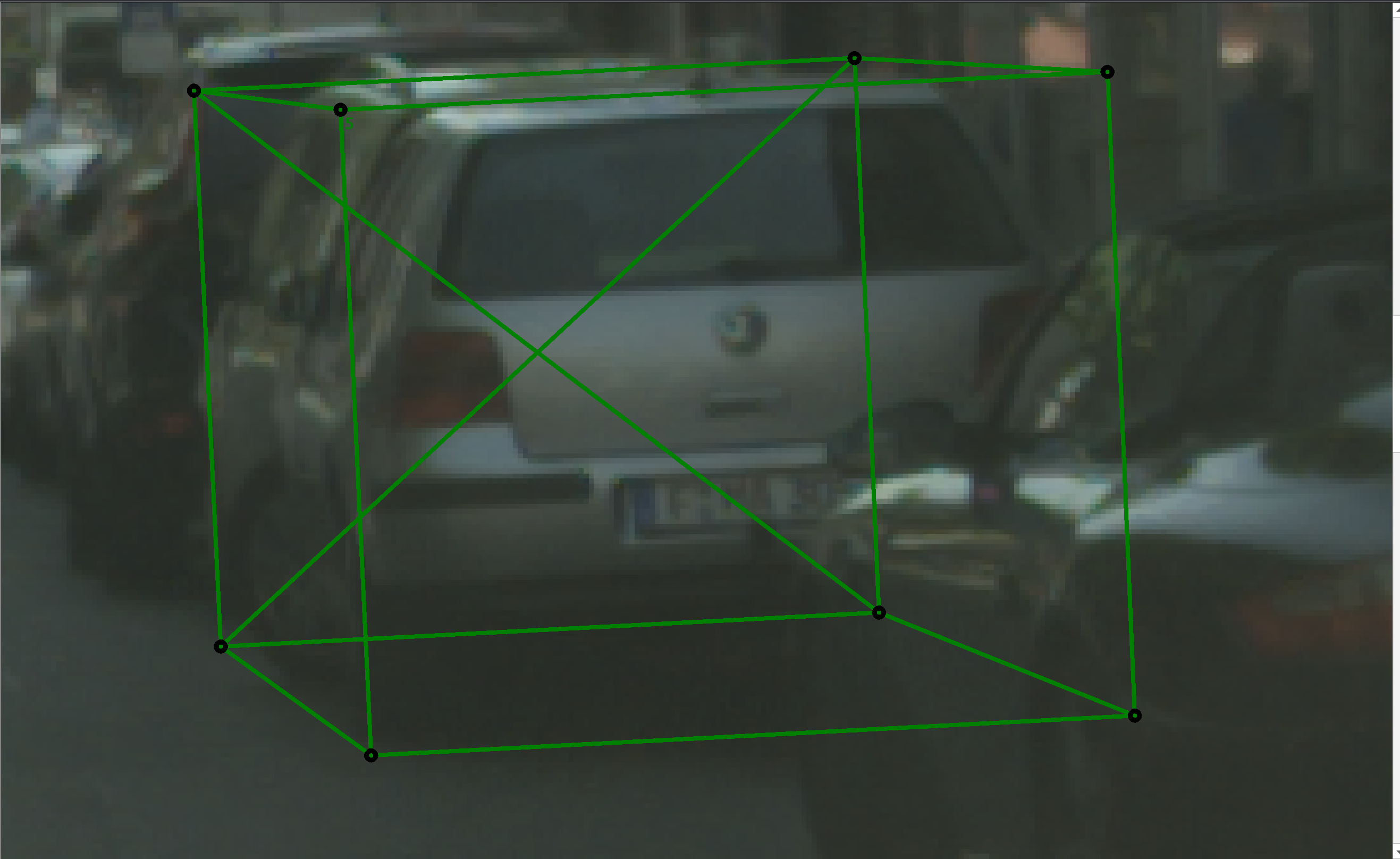} \\
		\includegraphics[width=0.40\linewidth]{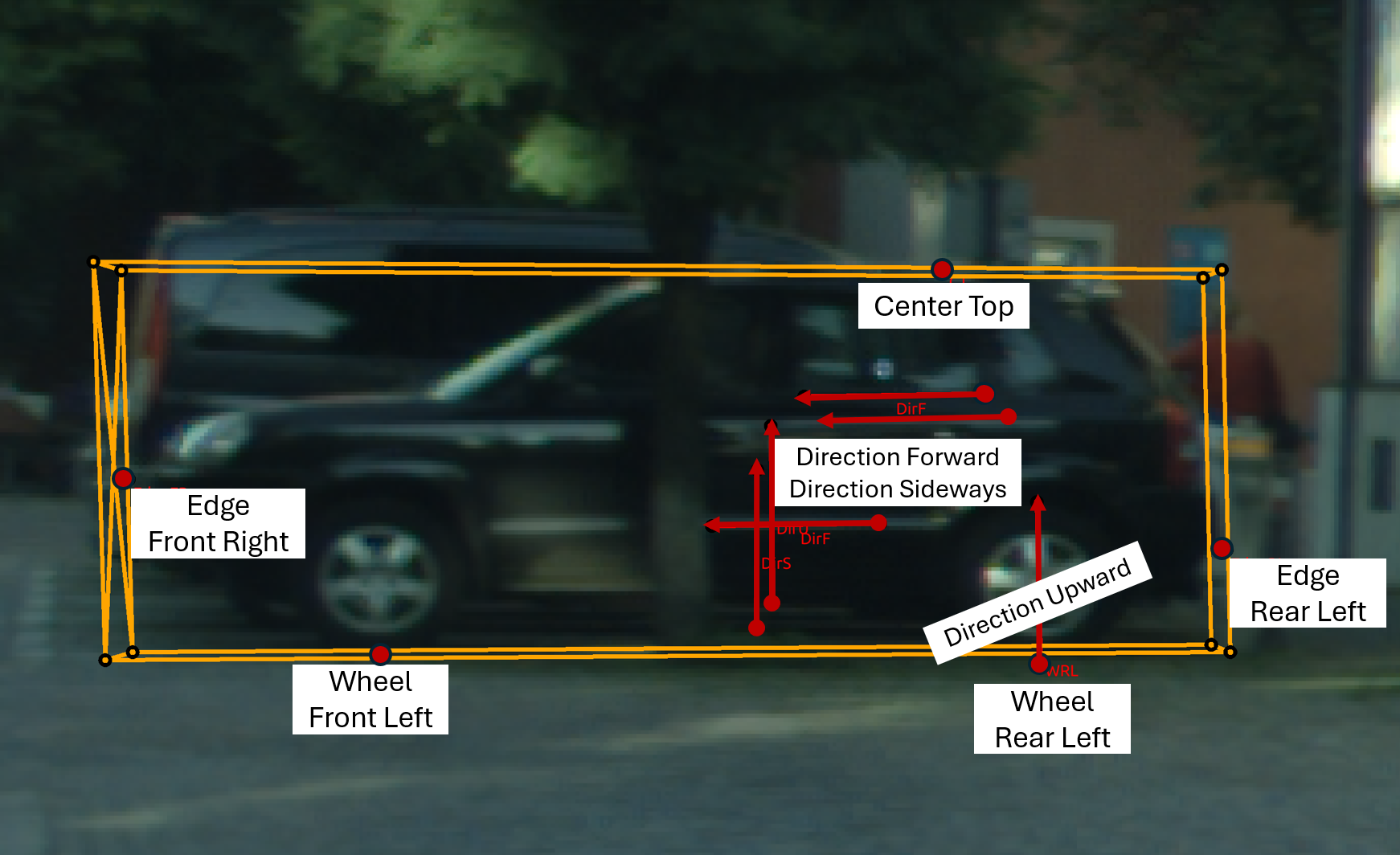} &
		\includegraphics[width=0.40\linewidth]{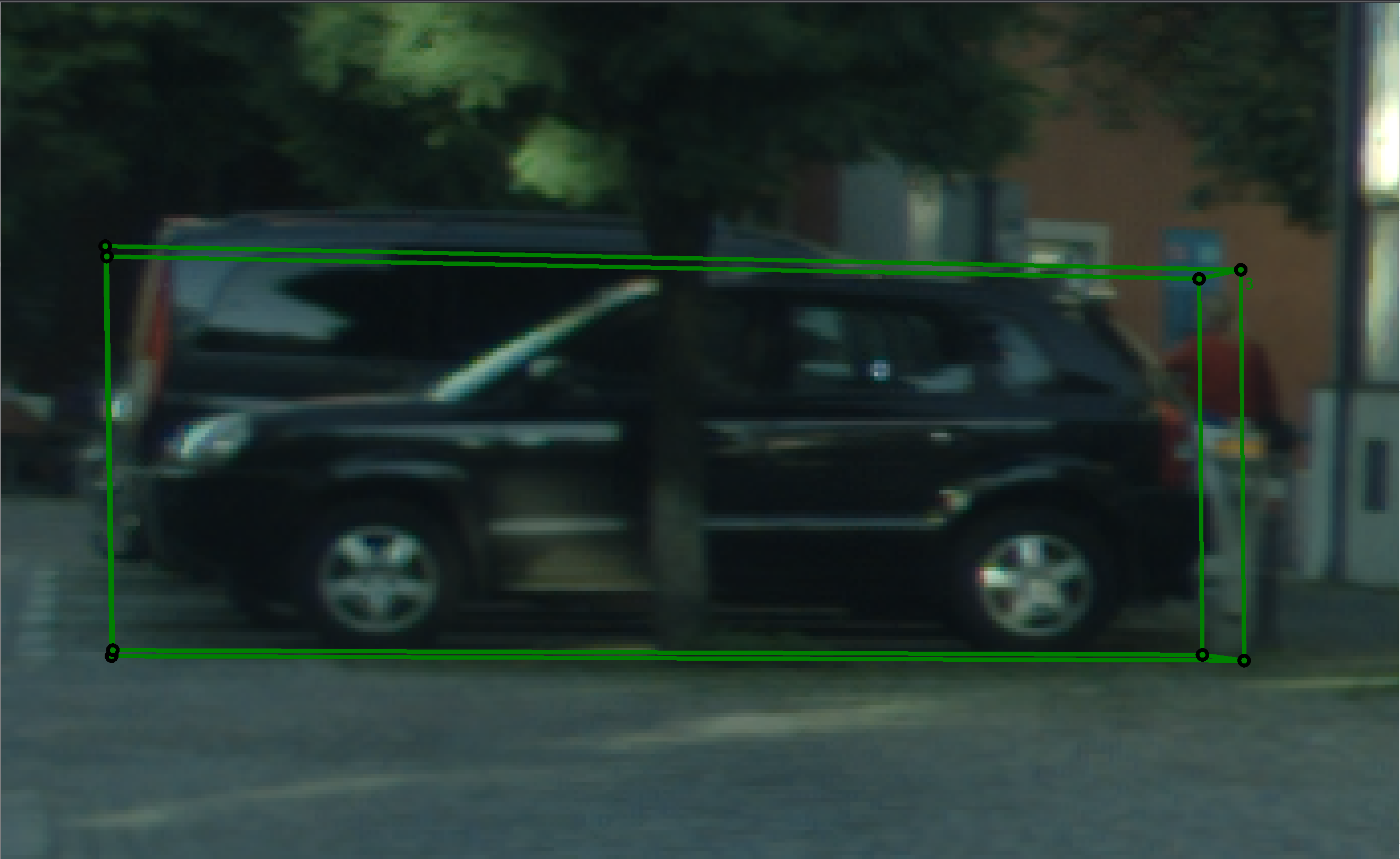} \\
		\includegraphics[width=0.40\linewidth]{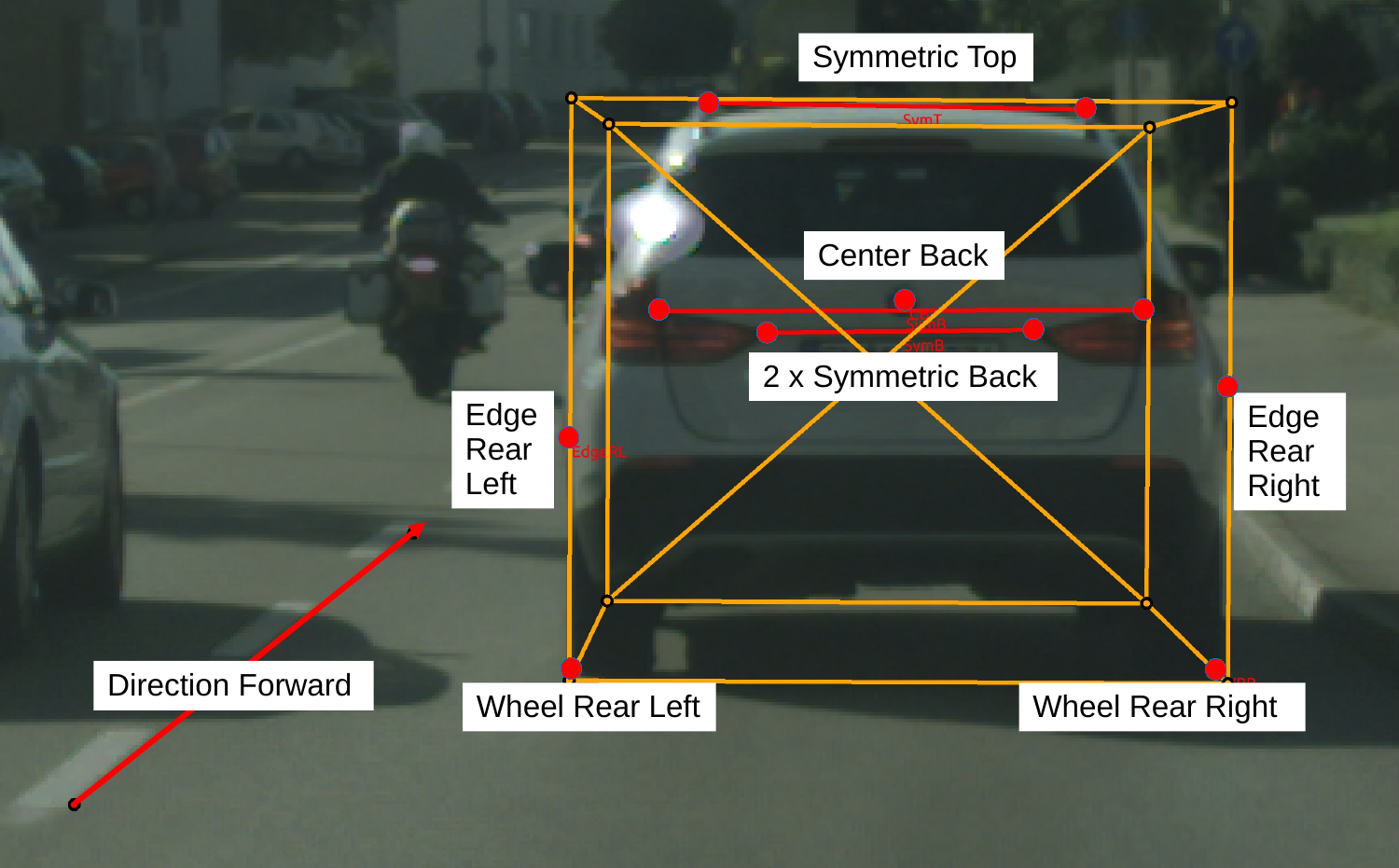} &
		\includegraphics[width=0.40\linewidth]{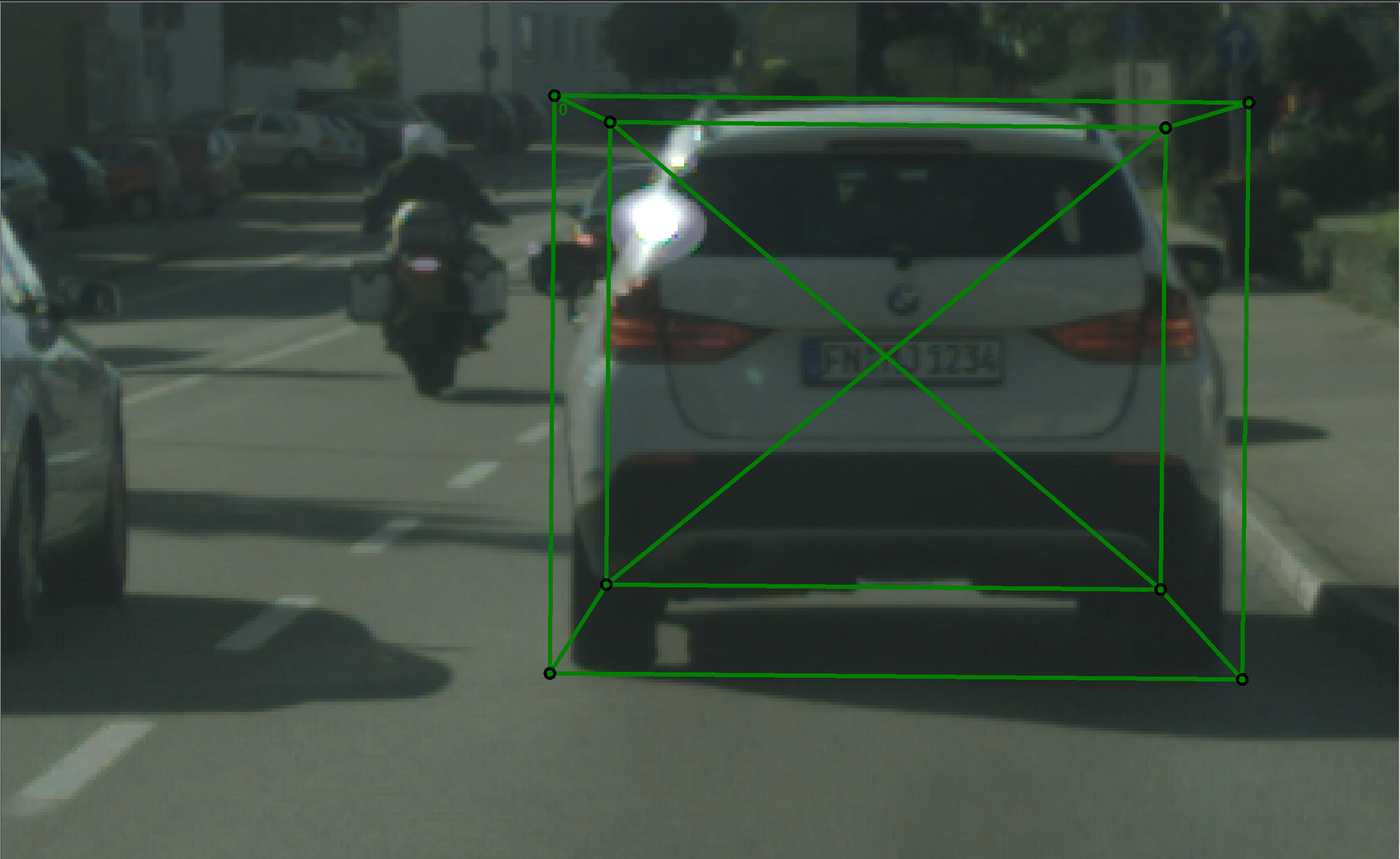} \\
		\includegraphics[width=0.40\linewidth]{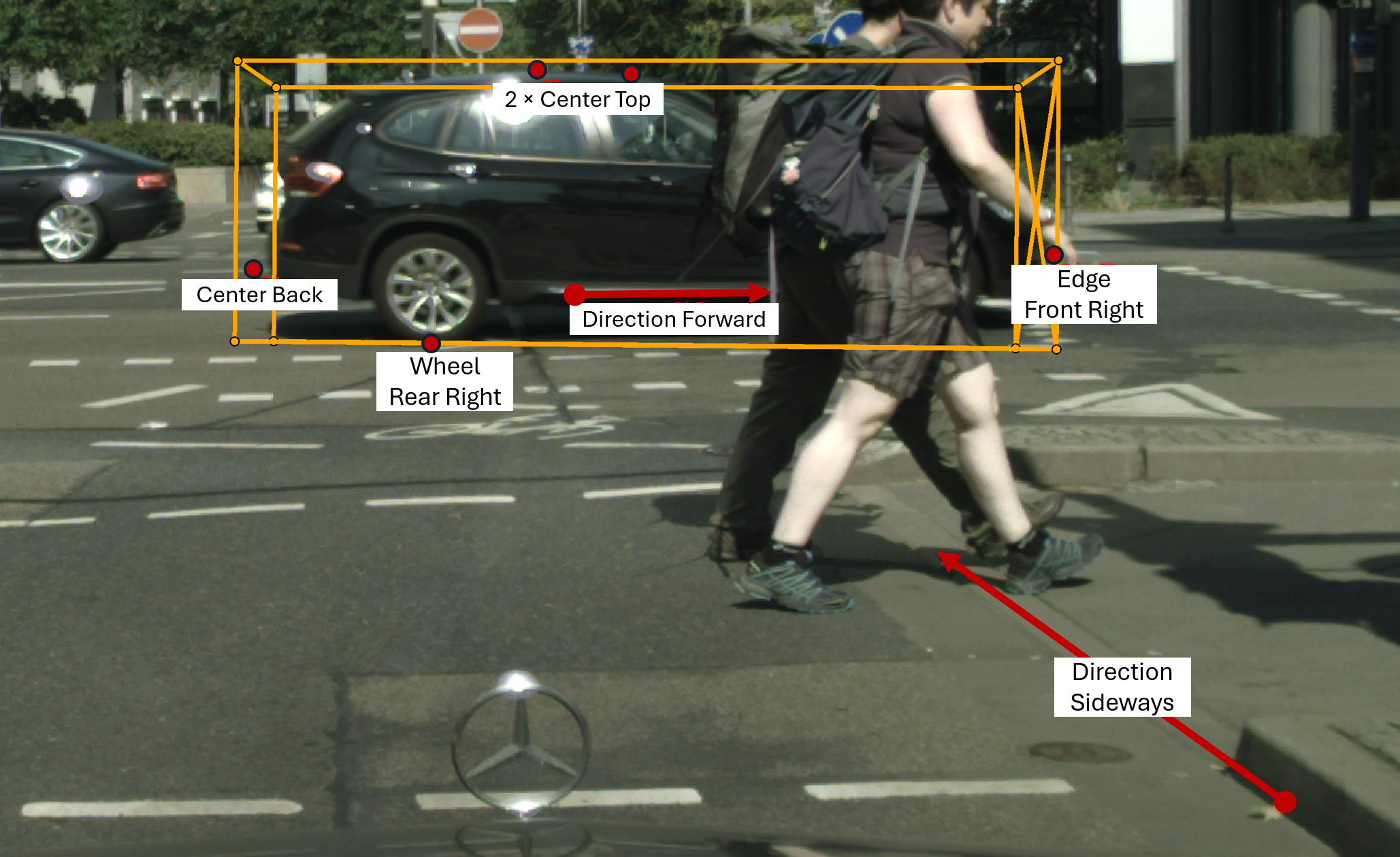} &
		\includegraphics[width=0.40\linewidth]{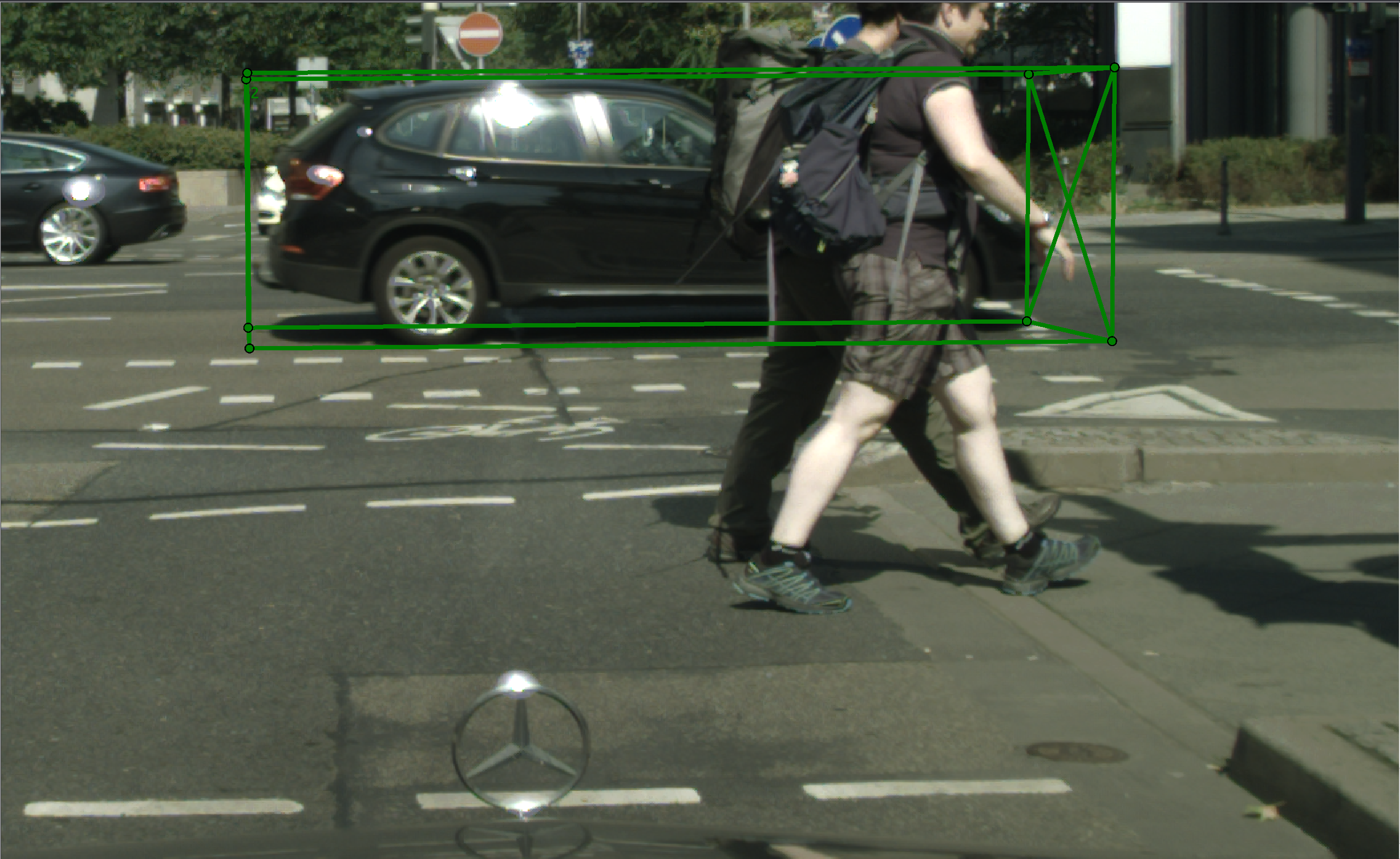} \\
		\includegraphics[width=0.40\linewidth]{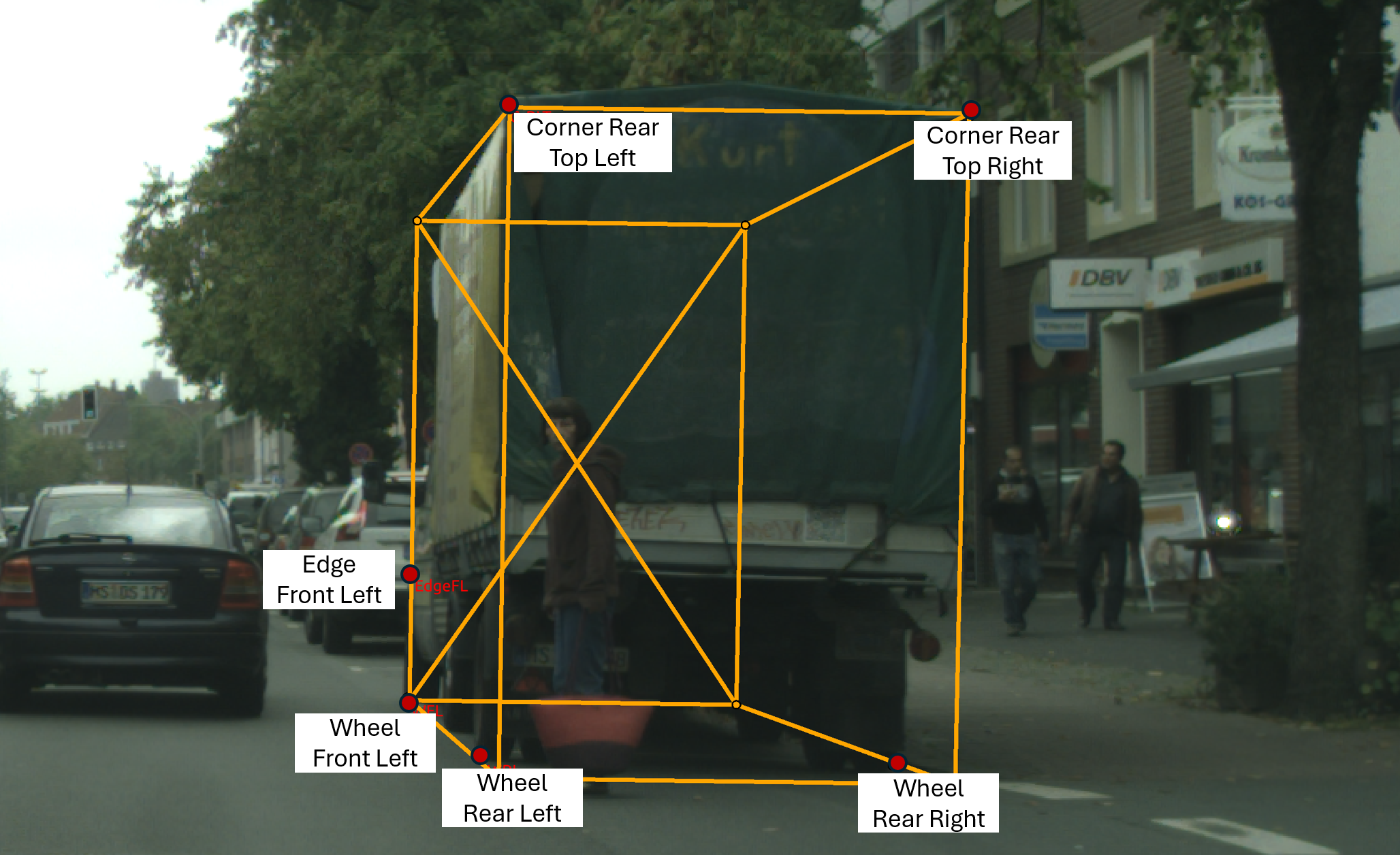} &
		\includegraphics[width=0.40\linewidth]{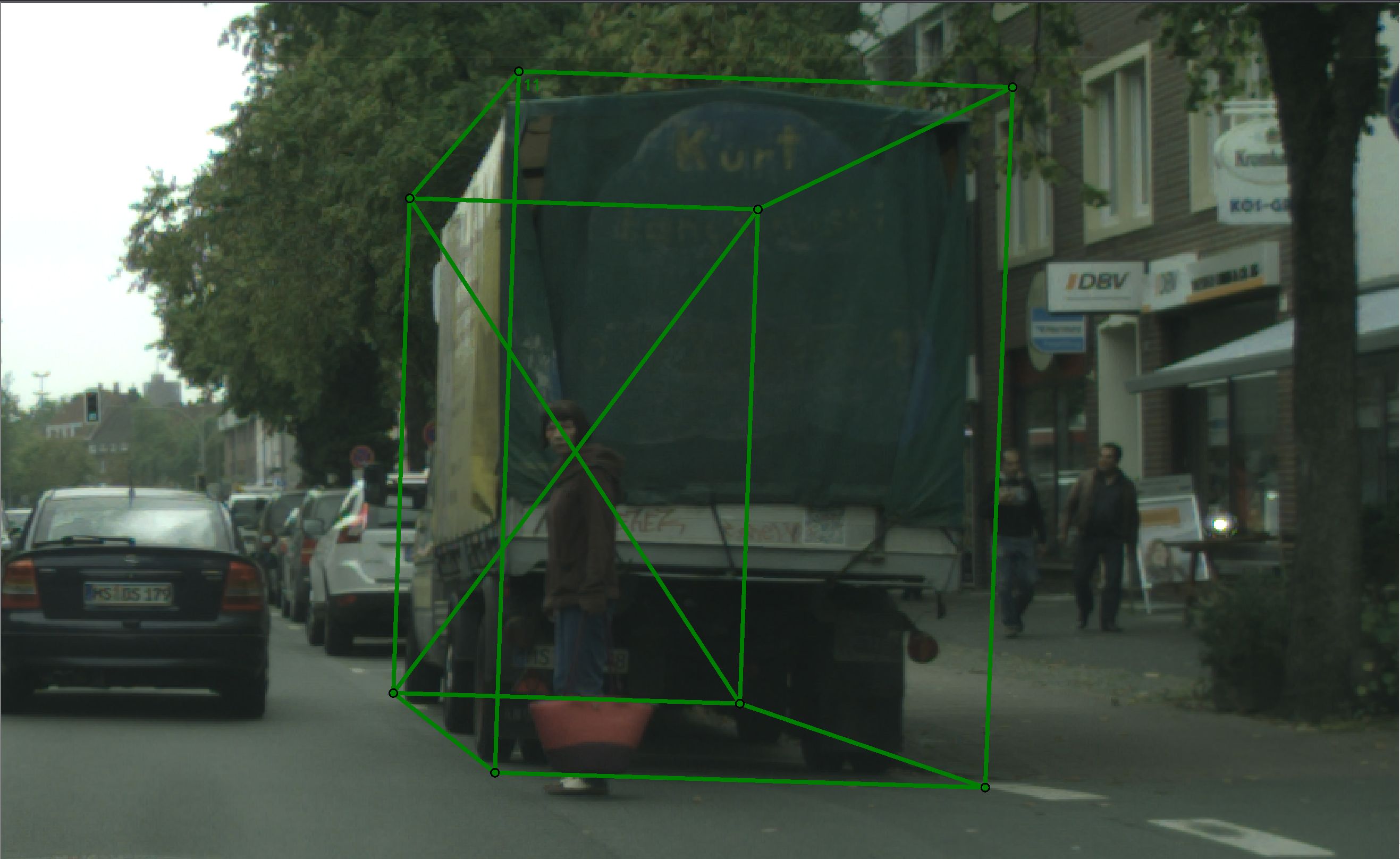}
	\end{tabular}
	\caption{Examples of cuboid annotations in the Cityscapes3D dataset. Left: Our annotations. Right: Cityscapes ground truth.}
	\label{fig:visual_Cityscapes}
\end{figure*}

Figs.~\ref{fig:visual_KITTI} and \ref{fig:visual_Cityscapes} compare the cuboids produced by our method against the KITTI and Cityscapes3D ground truth annotations, respectively.  
As shown in the images, our cuboids appear visually satisfactory across various vehicle types and viewing angles. 
Note that visual assessment primarily reflects the quality of 8DoF estimation. Compared to KITTI, our cuboids are generally tighter. 
Additionally, unlike KITTI, our method does not assume zero pitch and roll angles, which is evident in Fig.~\ref{fig:visual_KITTI}.

\begin{figure*}
	\centering
	\begin{tabular}{cc}
		Ours & KITTI \\
		\includegraphics[width=0.40\linewidth]{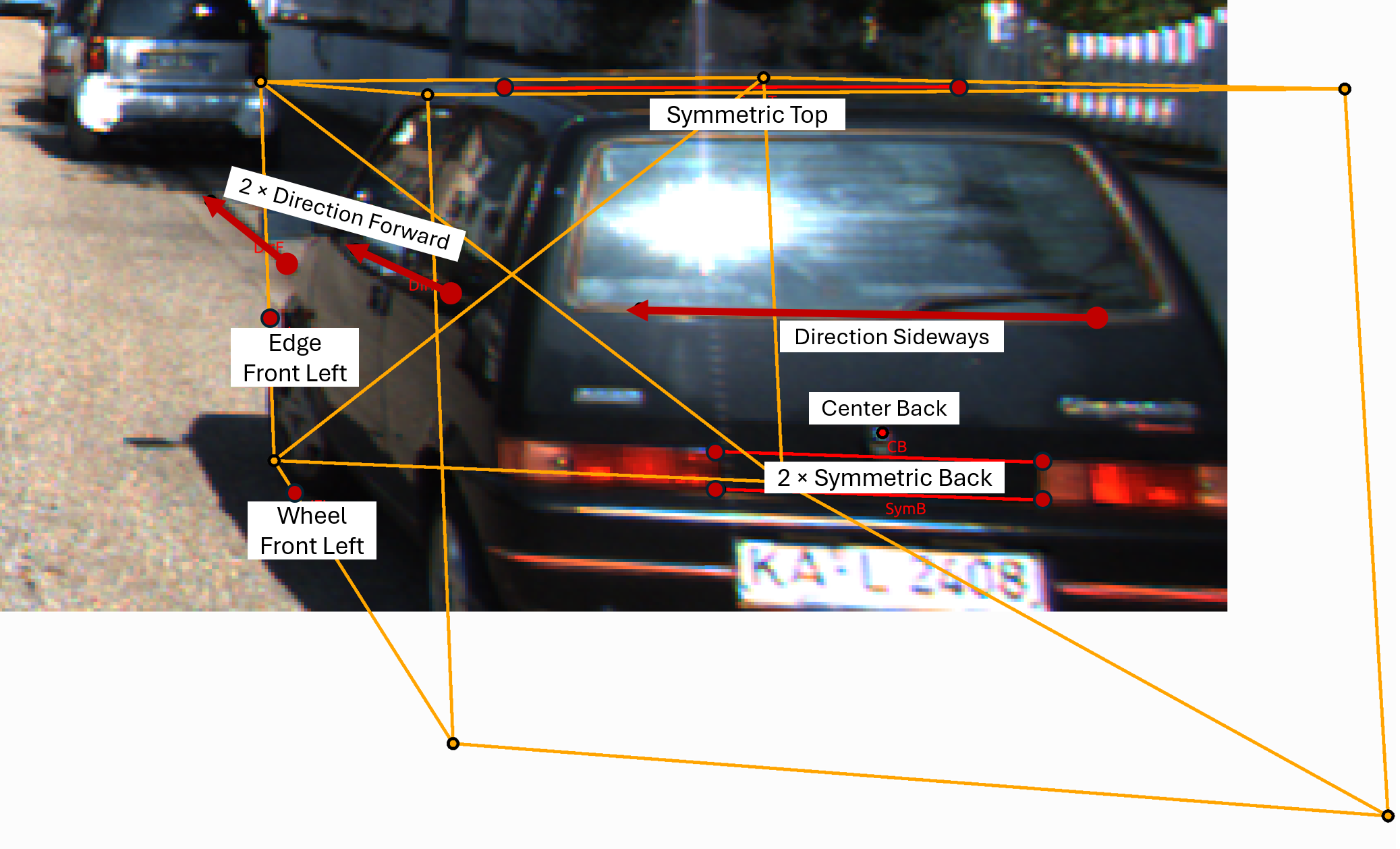} &
		\includegraphics[width=0.40\linewidth]{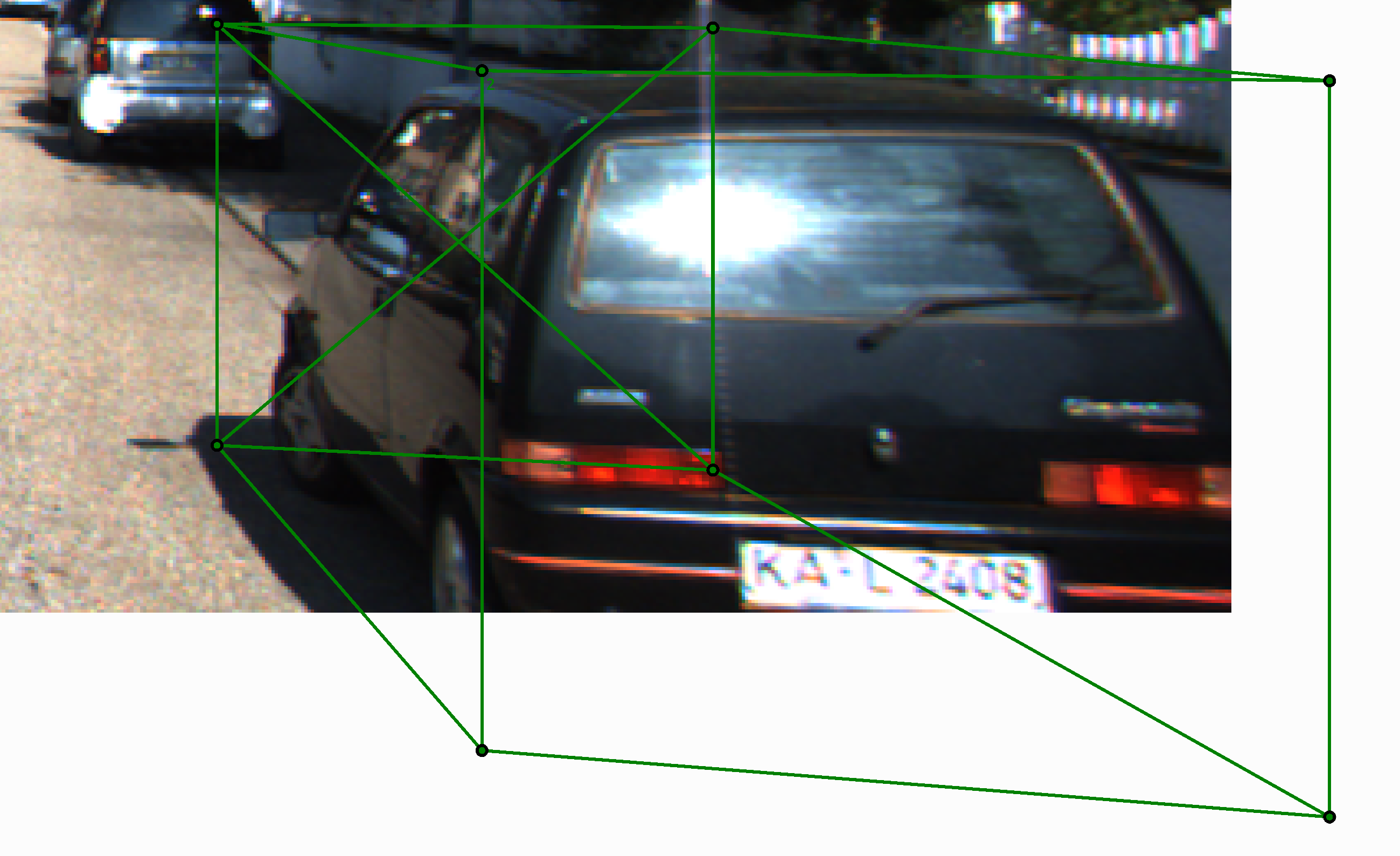} \\
		\includegraphics[width=0.40\linewidth]{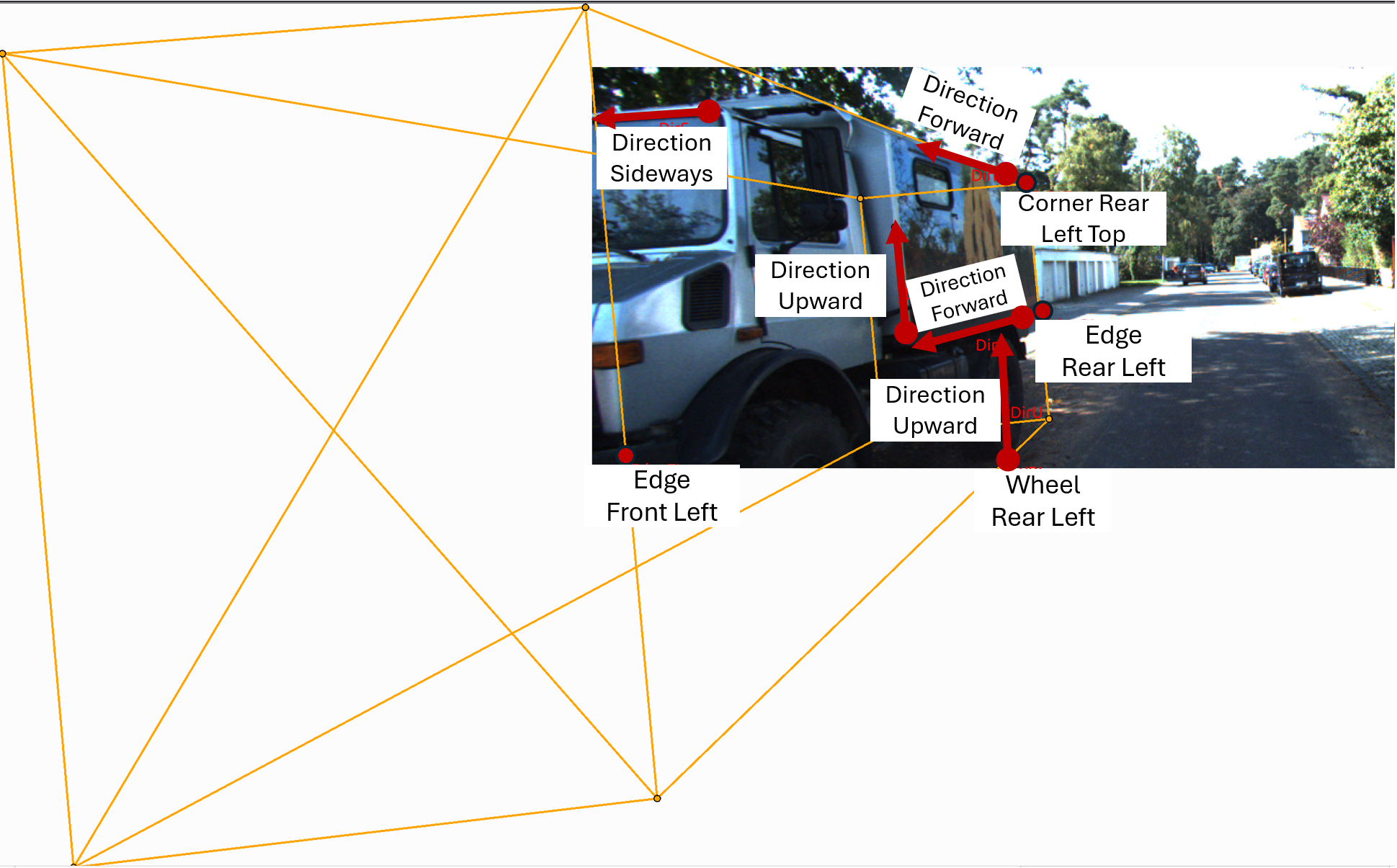} &
		\includegraphics[width=0.40\linewidth]{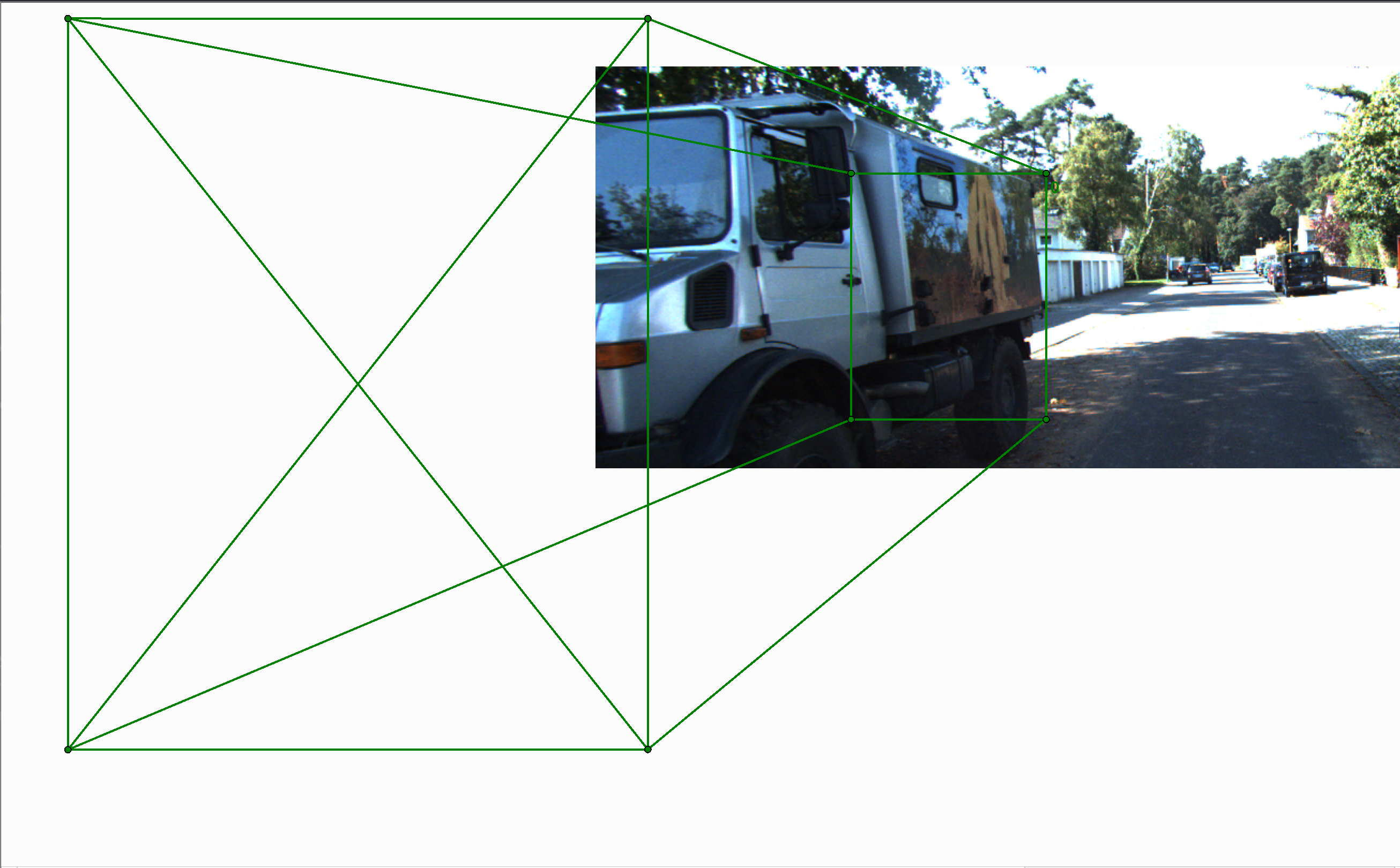} \\
		\includegraphics[width=0.40\linewidth]{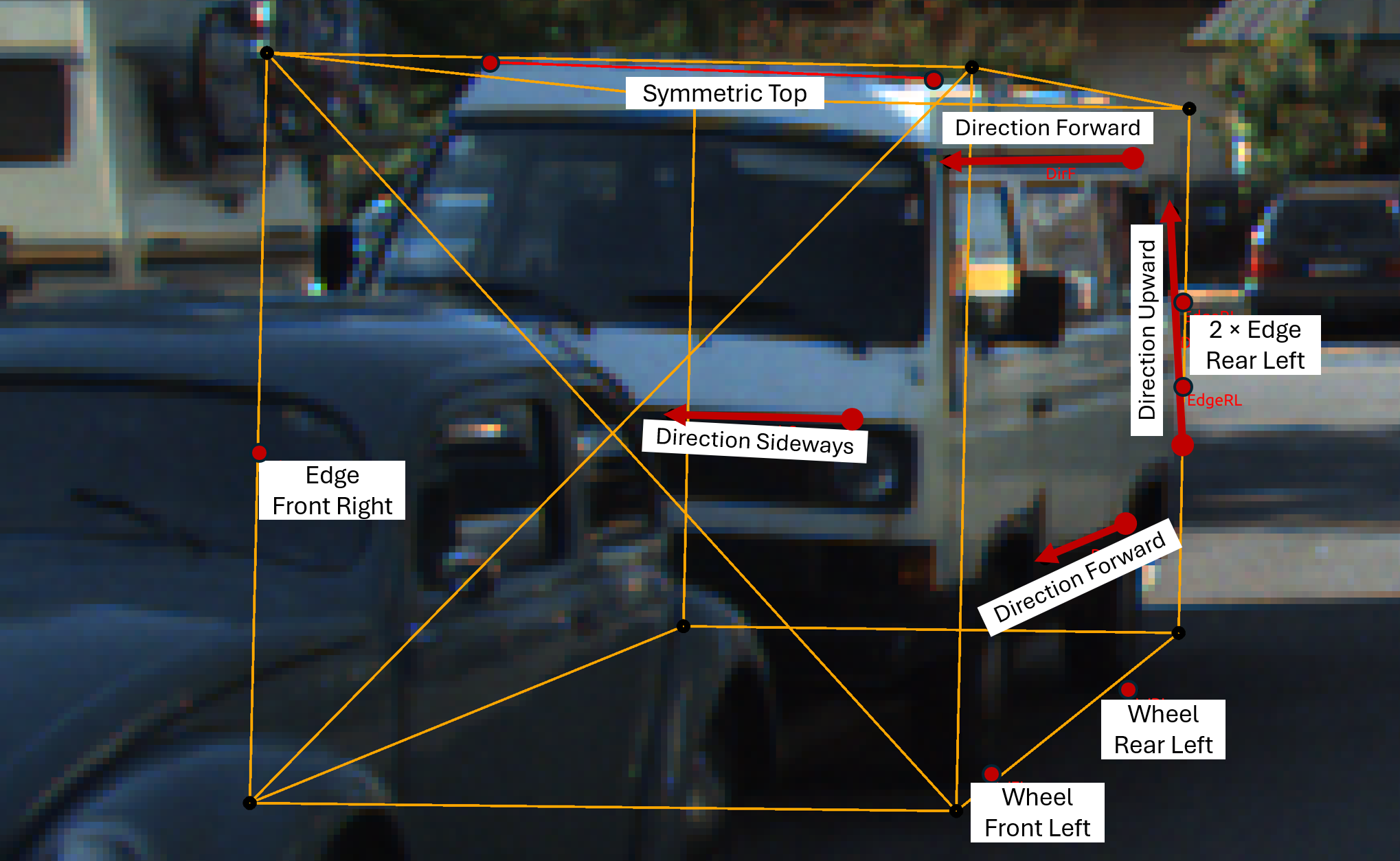} &
		\includegraphics[width=0.40\linewidth]{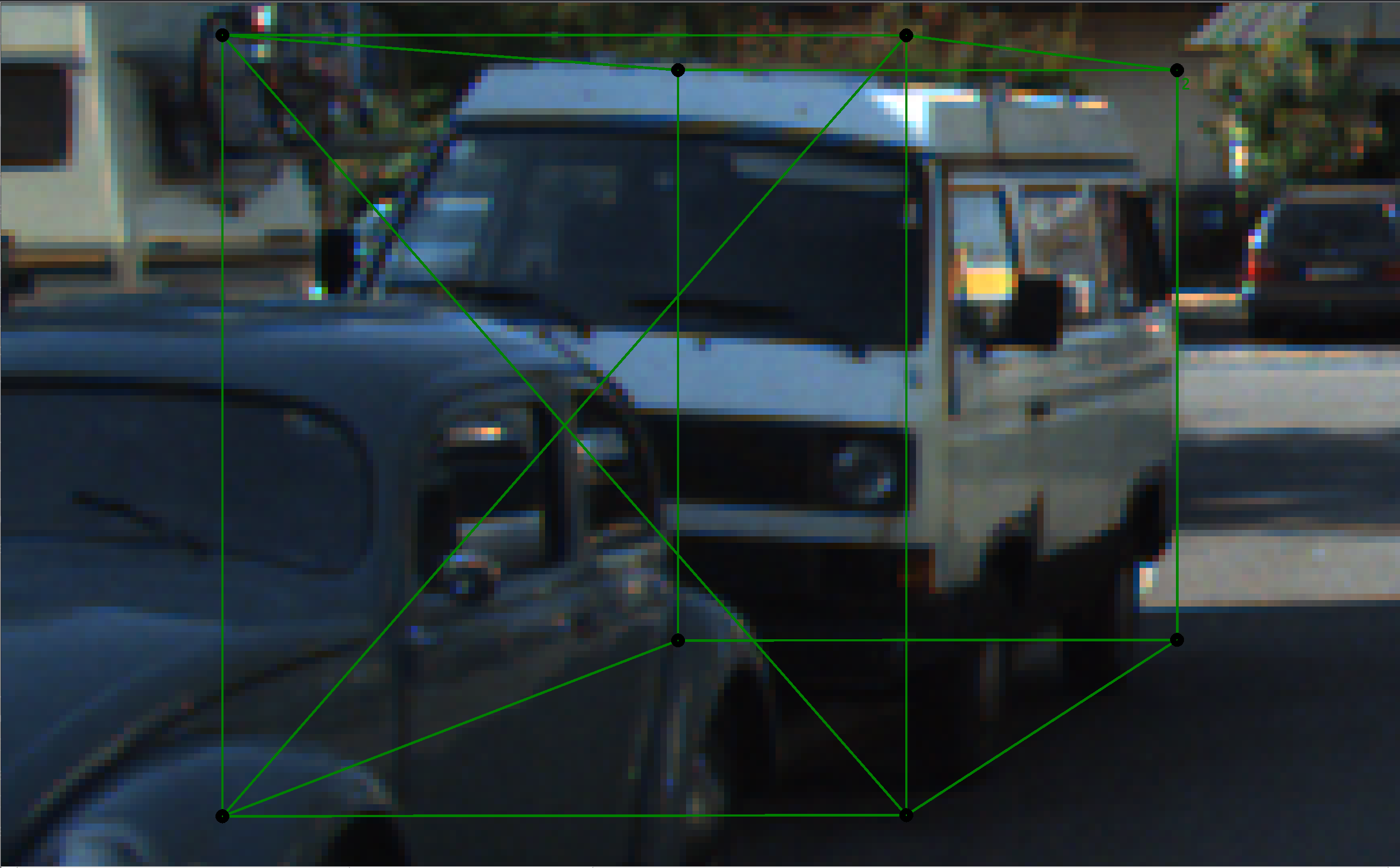} \\
		\includegraphics[width=0.40\linewidth]{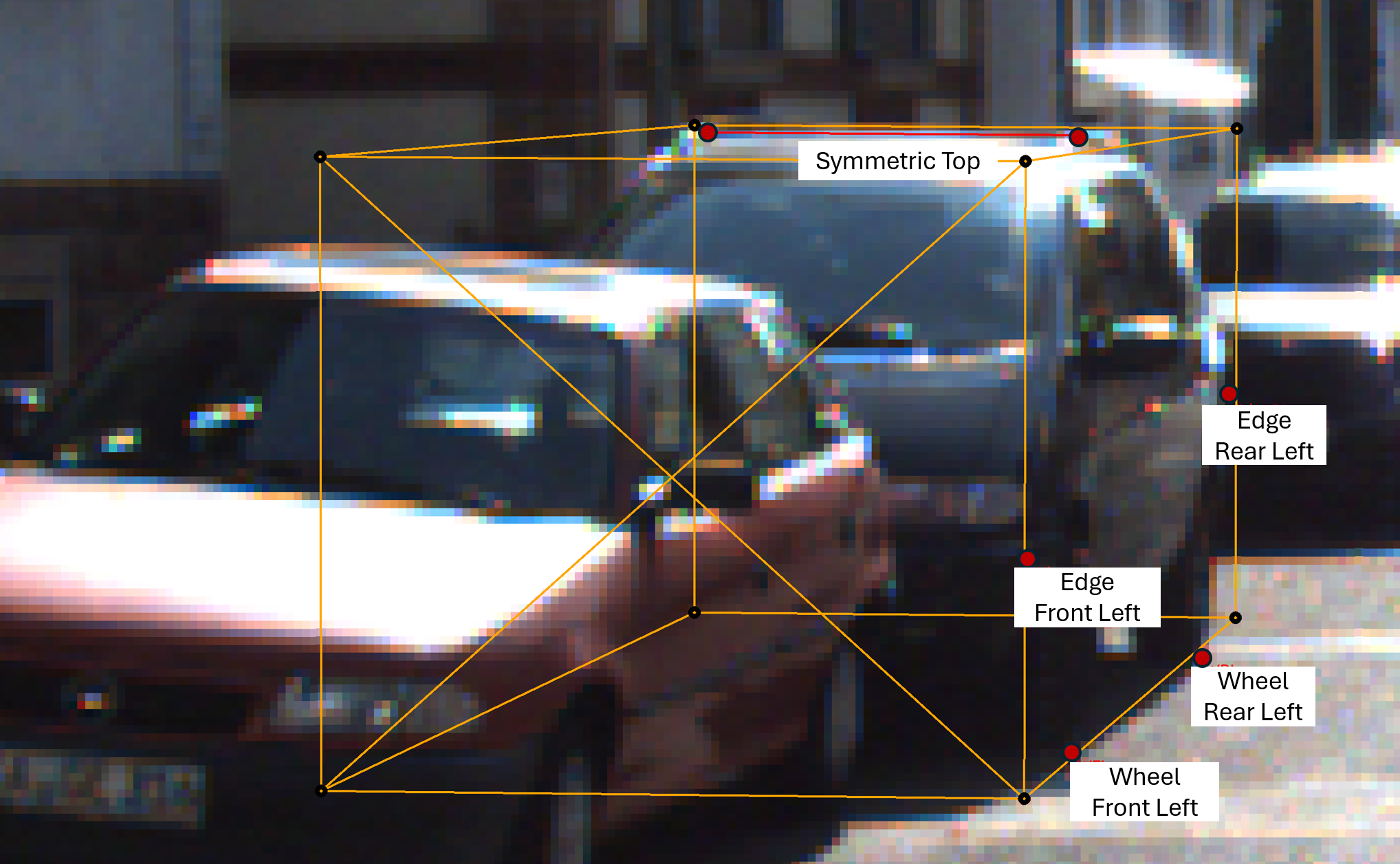} &
		\includegraphics[width=0.40\linewidth]{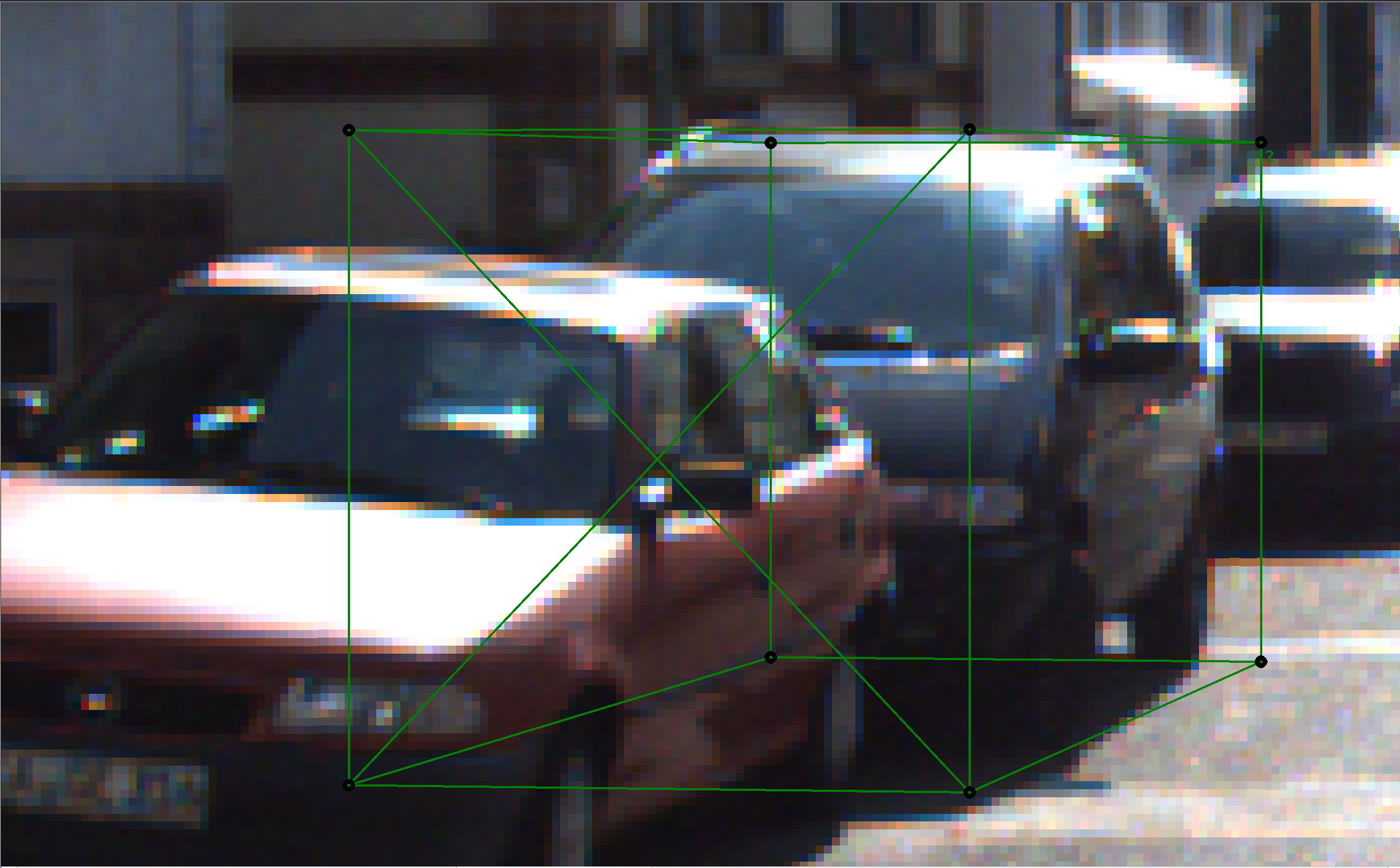} \\
		\includegraphics[width=0.40\linewidth]{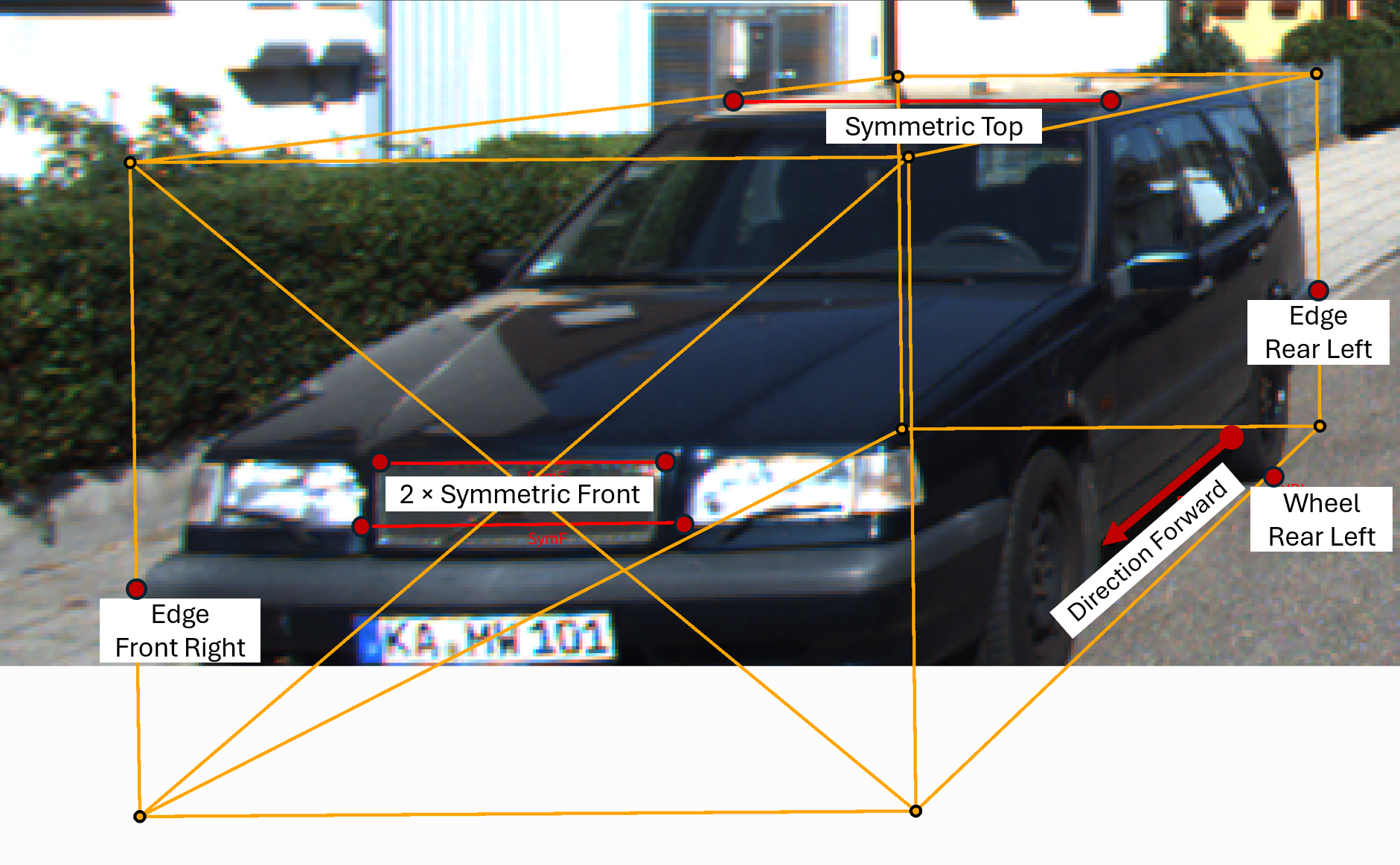} &
		\includegraphics[width=0.40\linewidth]{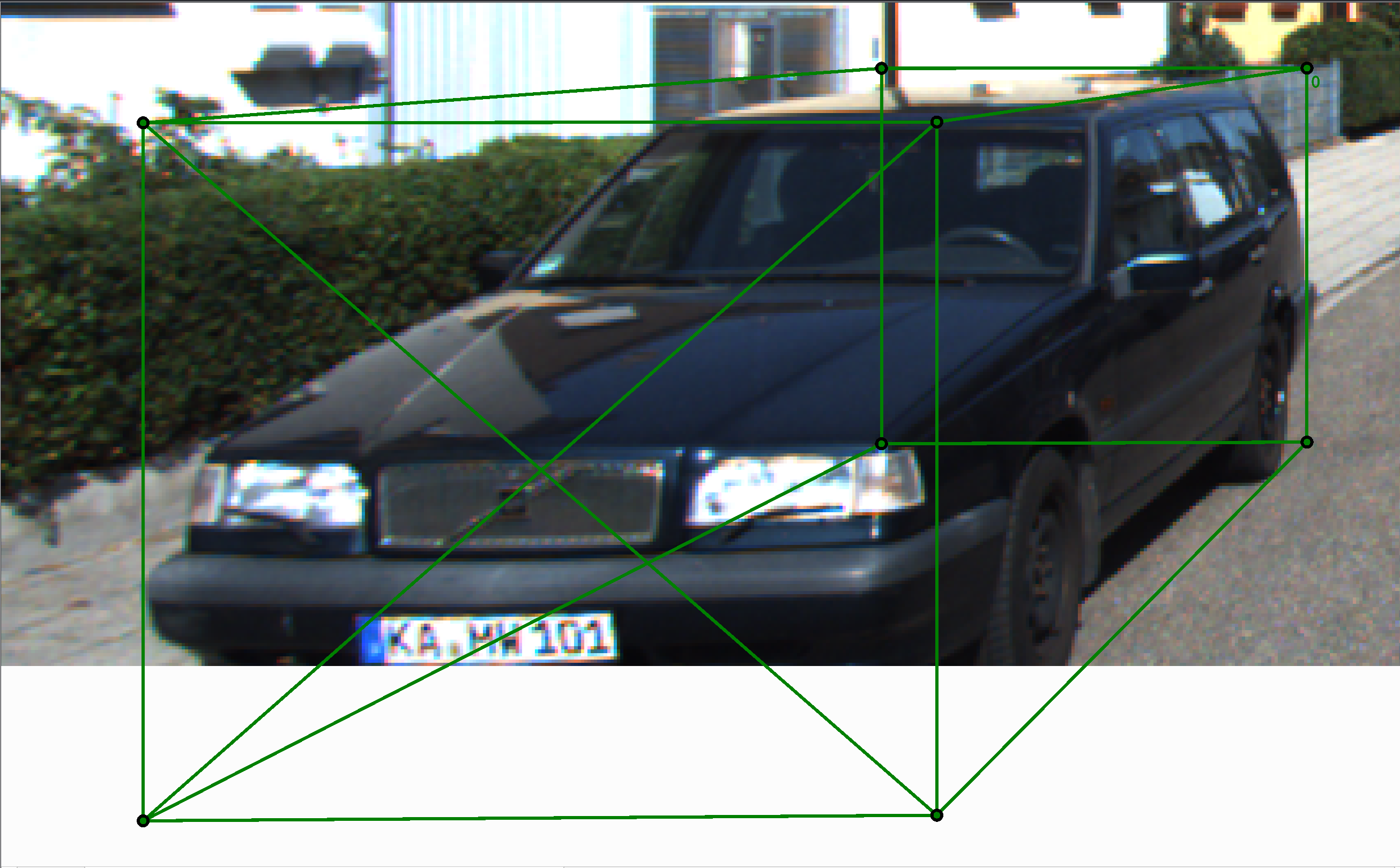}
	\end{tabular}
	\caption{Examples of cuboid annotations for occluded and truncated vehicles in the KITTI dataset. Left: Our annotations. Right: KITTI annotations.}
	\label{fig:occlusion_KITTI}
\end{figure*}

\begin{figure*}
	\centering
	\begin{tabular}{cc}
		Ours & Cityscapes \\
		\includegraphics[width=0.40\linewidth]{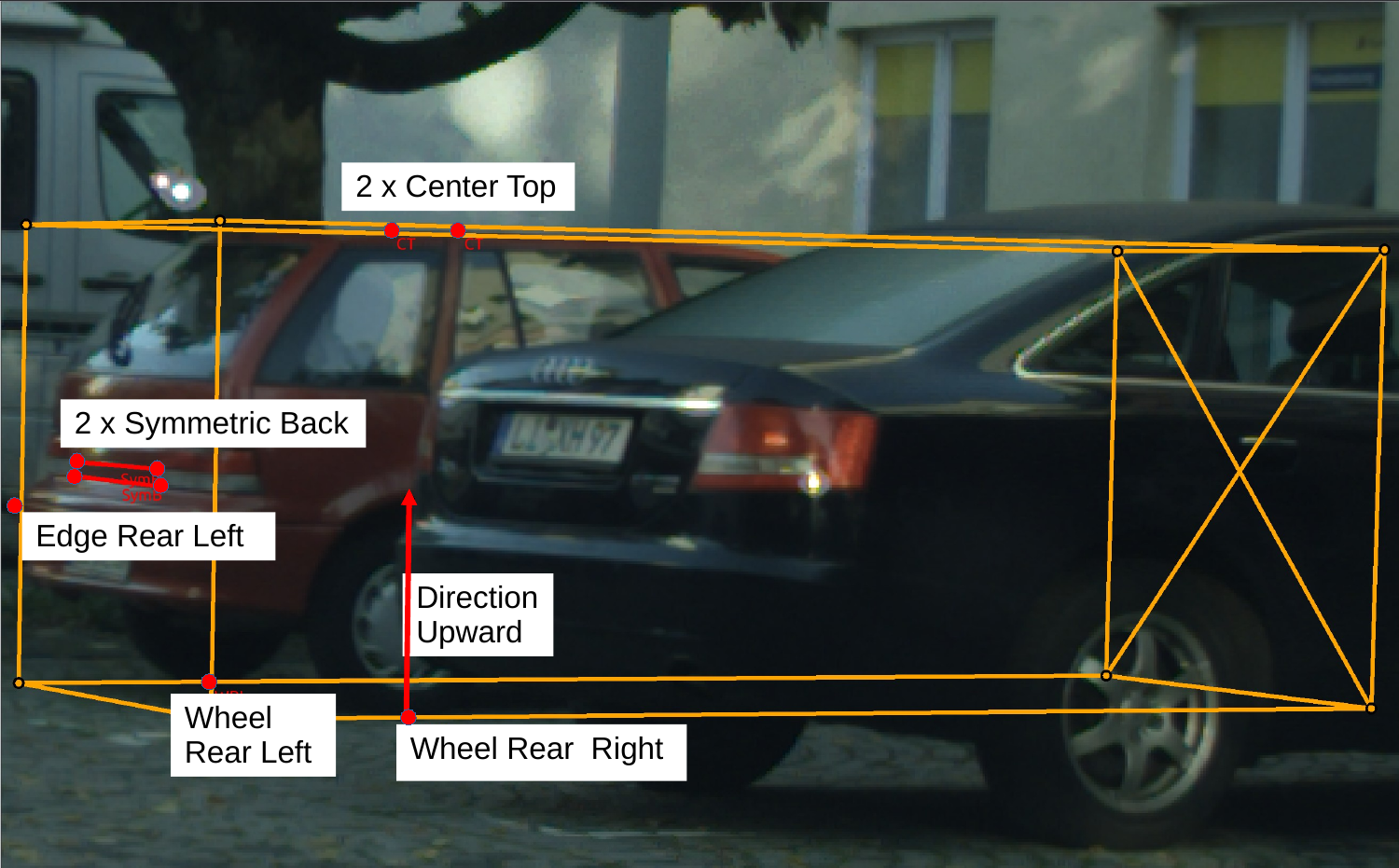} &
		\includegraphics[width=0.40\linewidth]{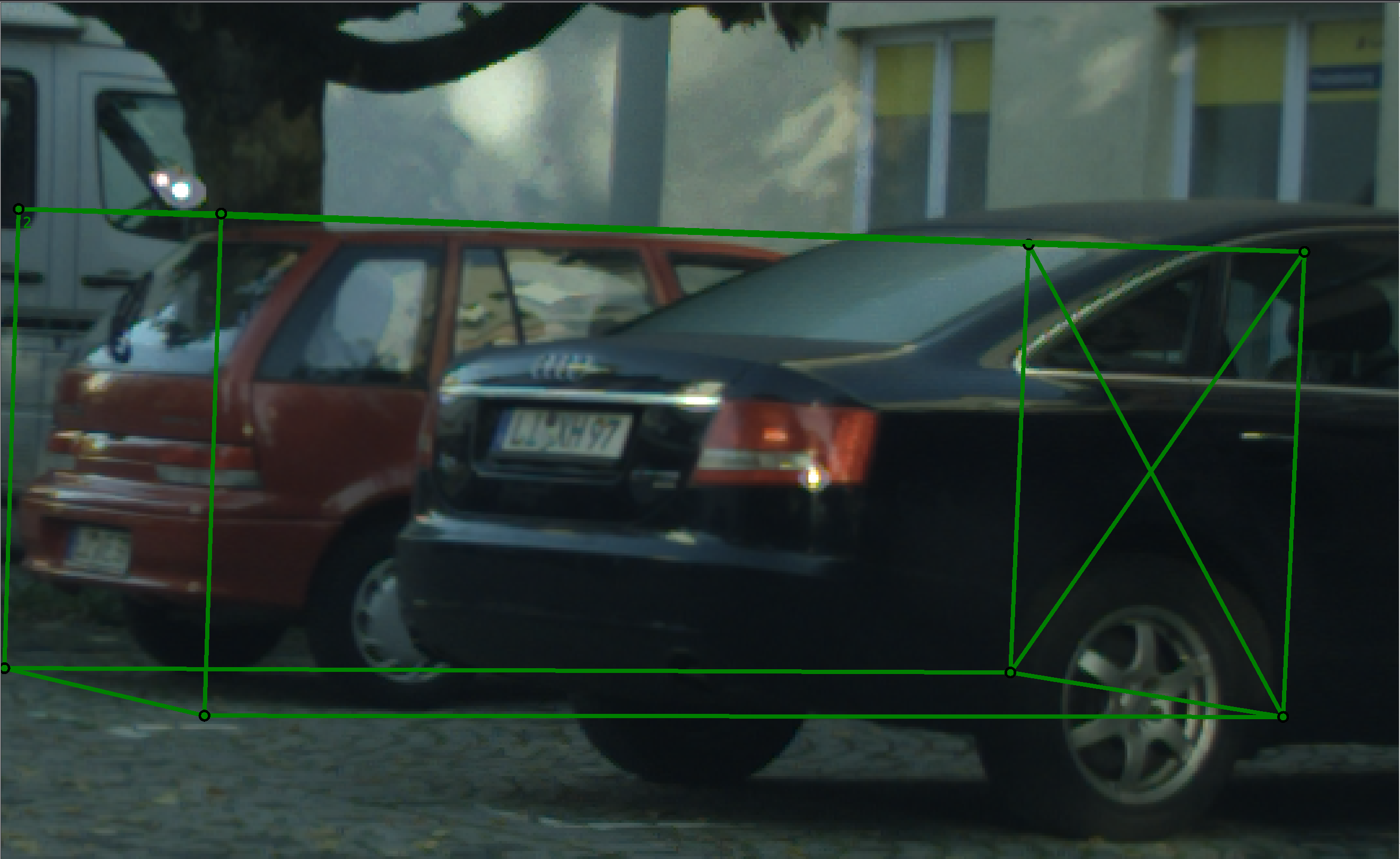} \\
		\includegraphics[width=0.40\linewidth]{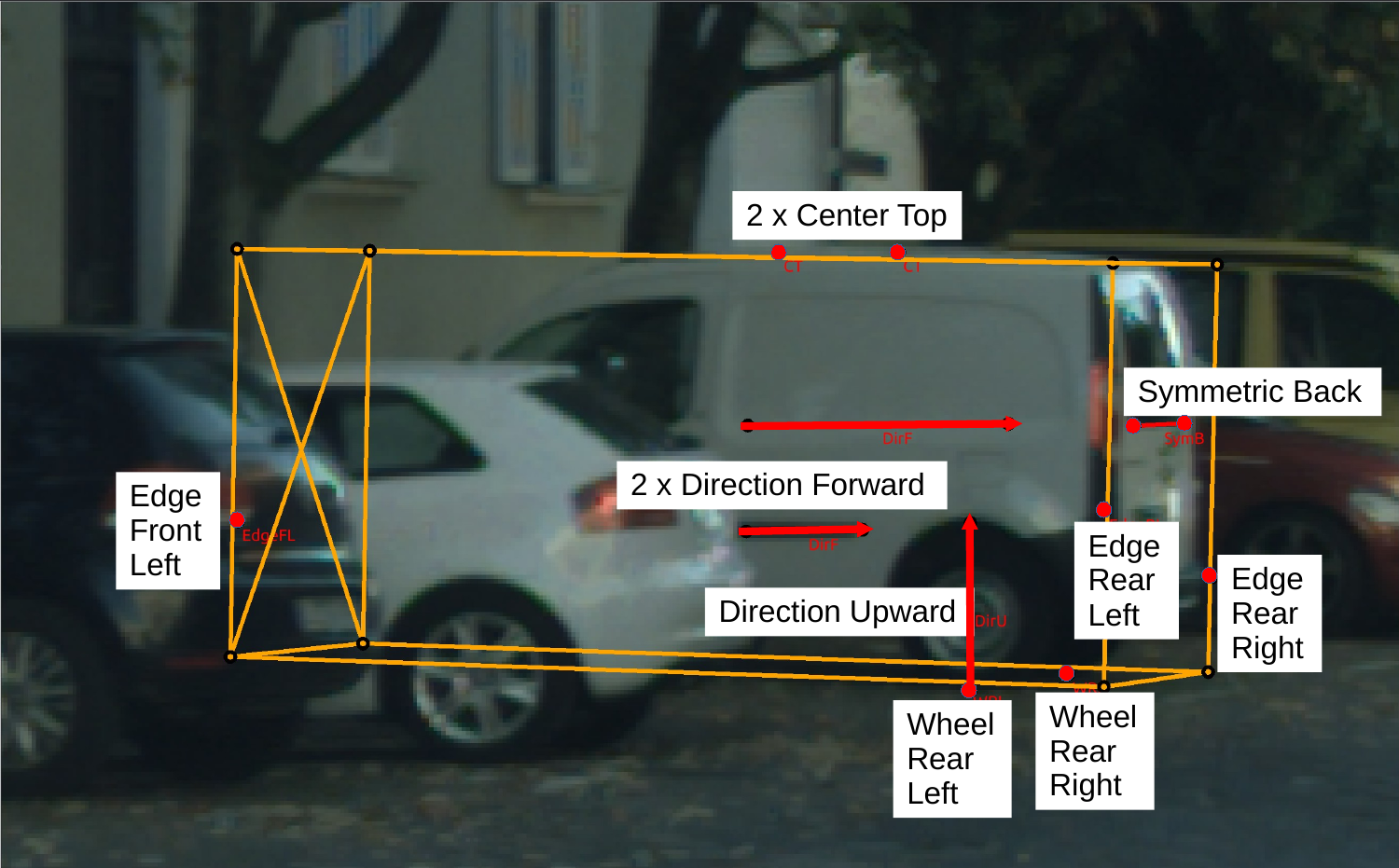} &
		\includegraphics[width=0.40\linewidth]{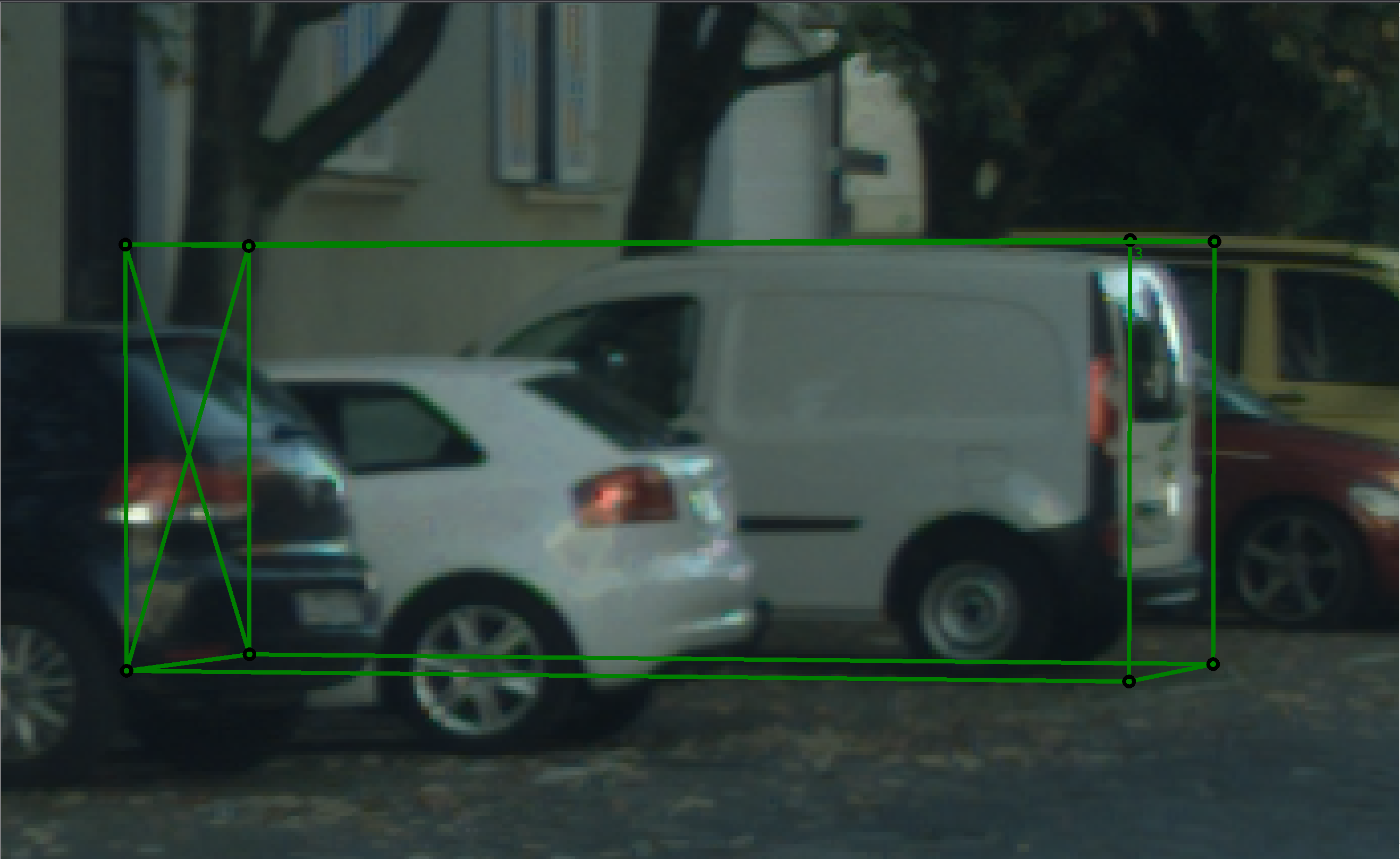} \\
		\includegraphics[width=0.40\linewidth]{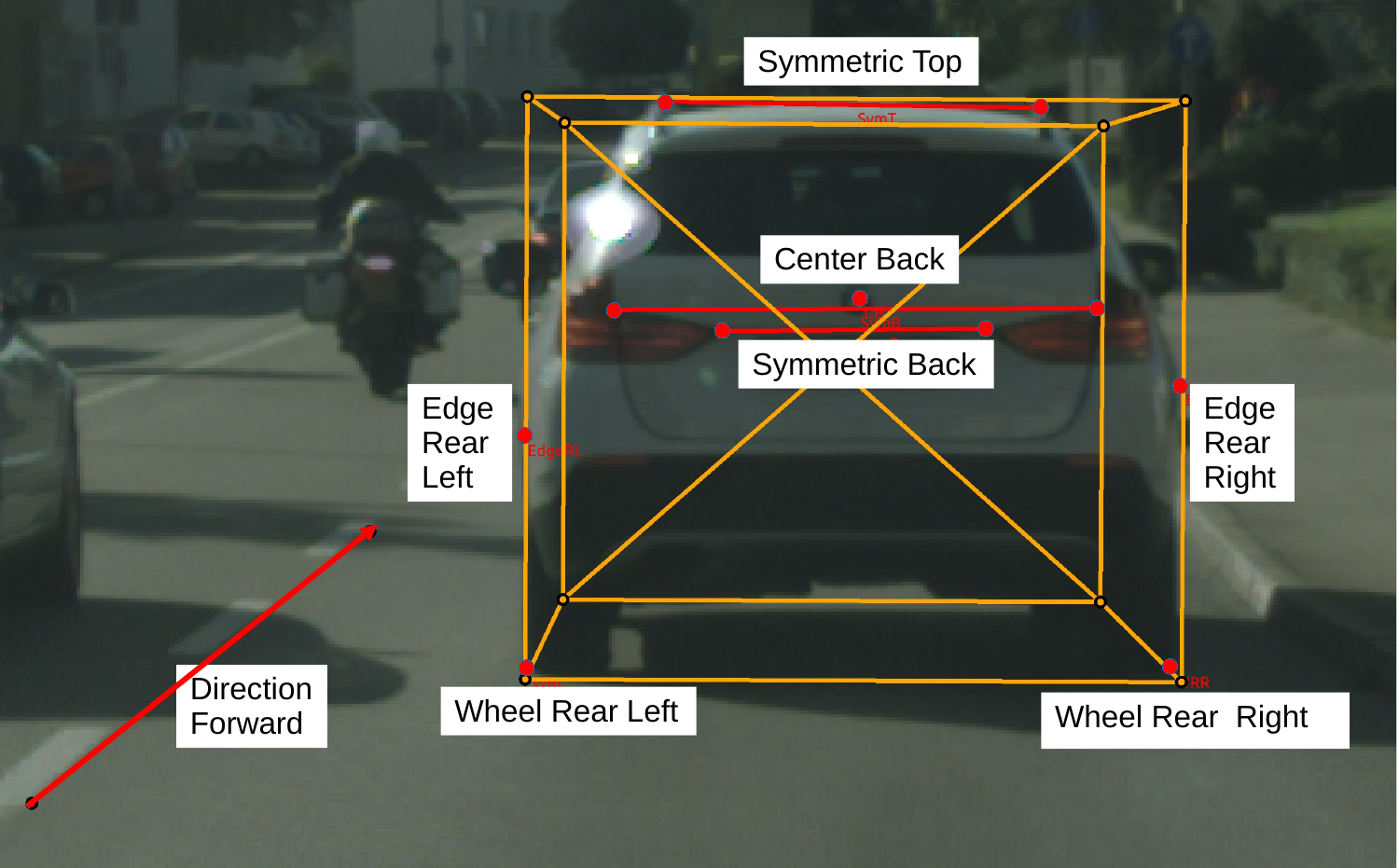} &
		\includegraphics[width=0.40\linewidth]{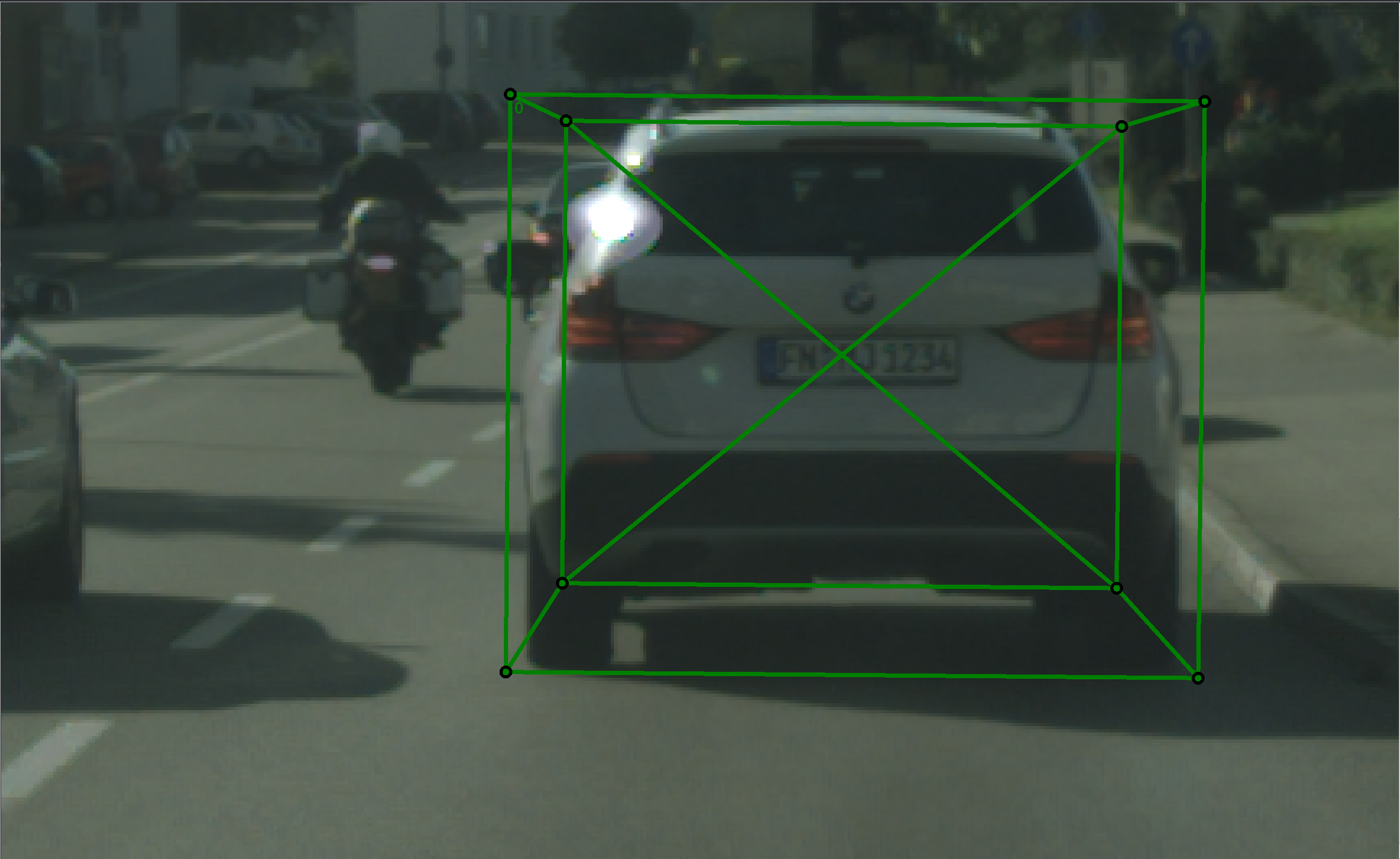} \\
		\includegraphics[width=0.40\linewidth]{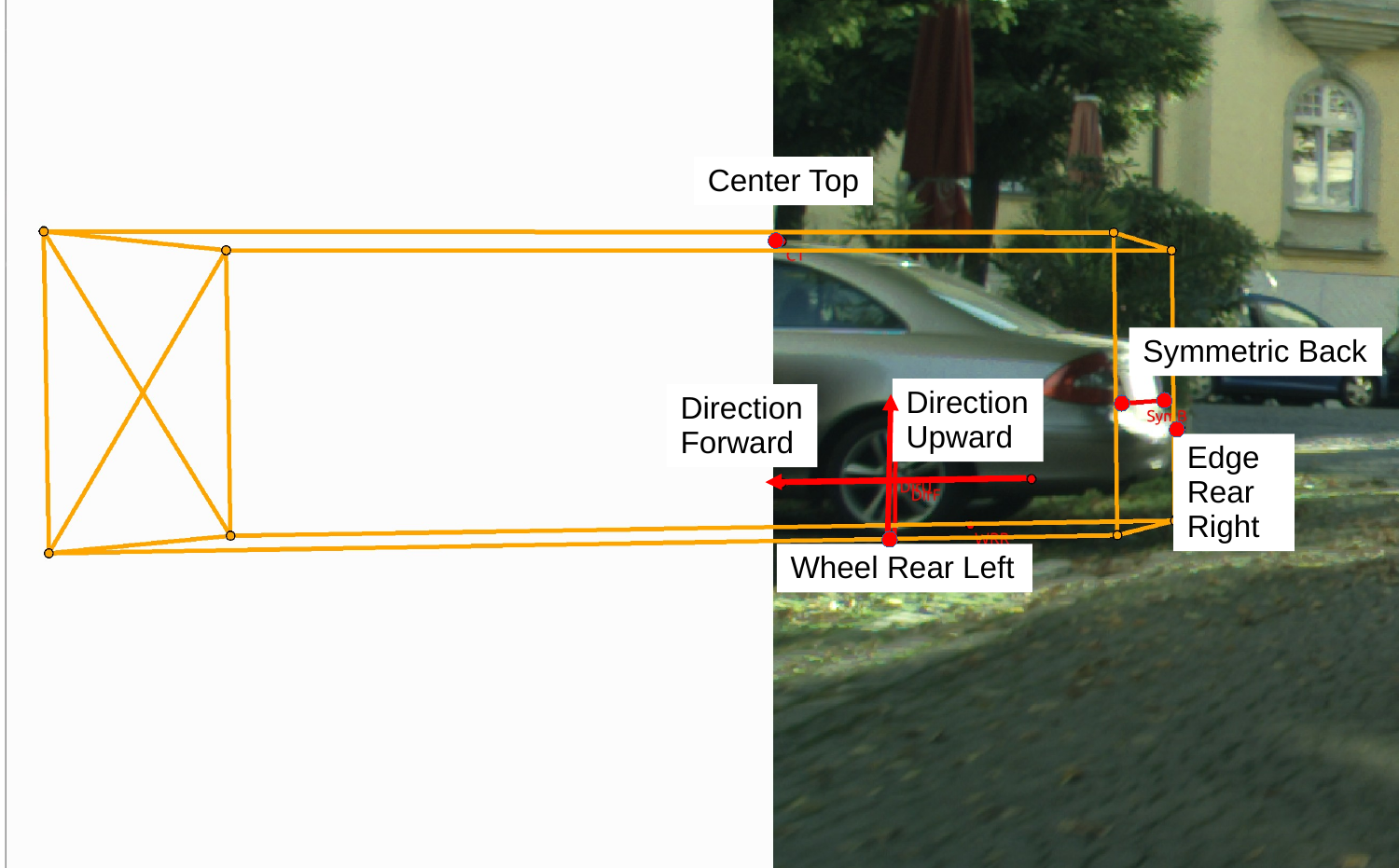} &
		\includegraphics[width=0.40\linewidth]{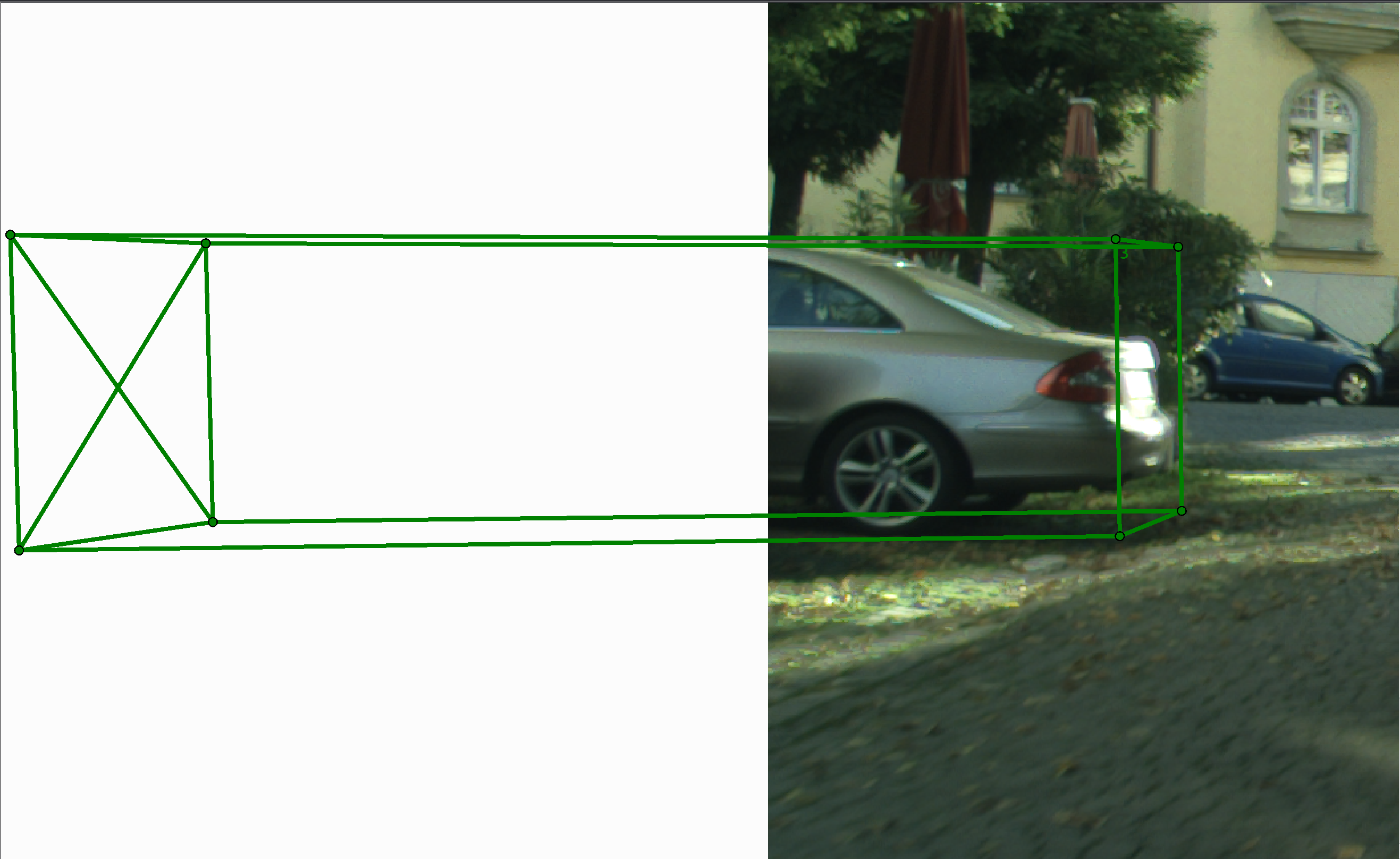} \\
		\includegraphics[width=0.40\linewidth]{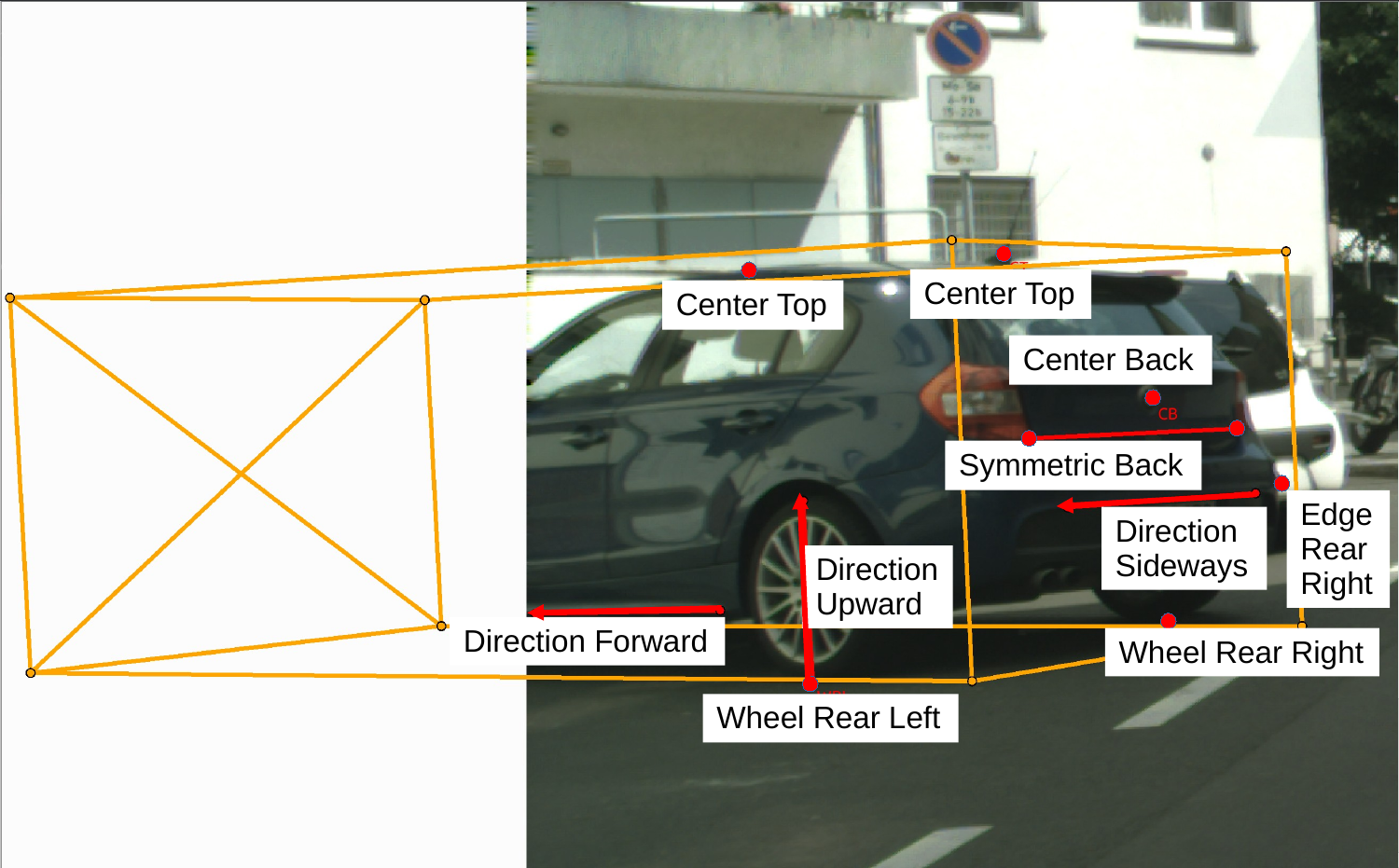} &
		\includegraphics[width=0.40\linewidth]{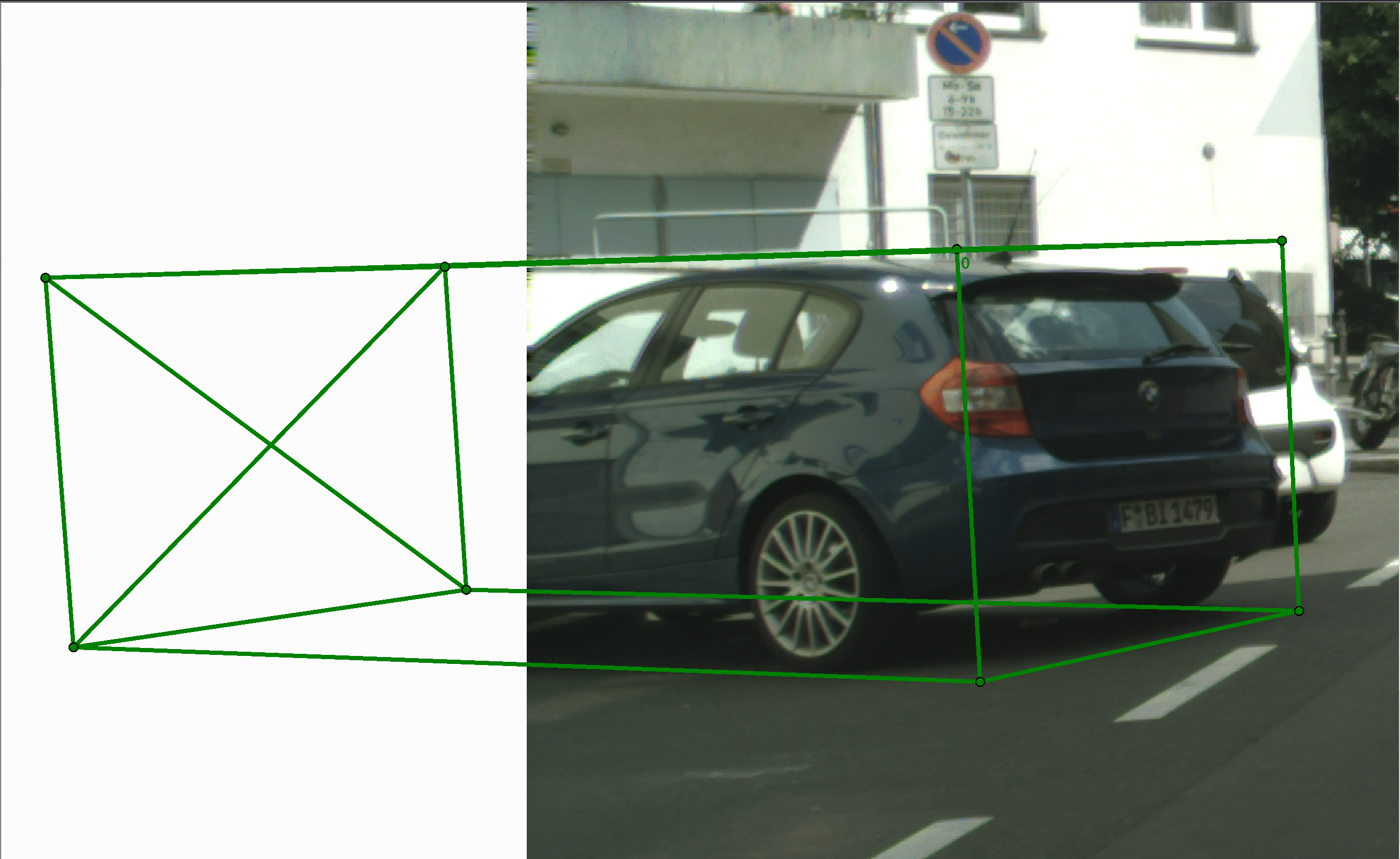}
	\end{tabular}
	\caption{Examples of cuboid annotations for occluded and truncated vehicles in the Cityscapes3D dataset. Left: Our annotations. Right: Cityscapes ground truth. Note that in the second row, the annotator have made a guess about the location of \emph{Edge-Front-Left} despite it not being directly visible. We generally avoid this practice unless it is absolutely necessary.}
	\label{fig:occlusion_Cityscapes}
\end{figure*}

Figs.~\ref{fig:occlusion_KITTI} and \ref{fig:occlusion_Cityscapes} show qualitative results on occluded and truncated vehicles in the KITTI and Cityscapes3D datasets. 
Our method successfully handles moderate to high levels of occlusion and truncation, as long as sufficient 2D vehicle features are visible.

\subsection{Discussion}
\label{sec:discussion}
Our approach demonstrates strong performance in estimating 8DoF cuboids (i.e., up to scale), even in the presence of occlusion or truncation. The primary limitation arises when an insufficient number of 2D features are visible, which typically occurs under severe occlusion, truncation, or when vehicles are too distant from the camera. 

The use of probabilistic size priors proves effective in resolving ambiguities related to unobserved dimensions. However, these priors are only moderately successful at addressing scale ambiguity, which is crucial for obtaining accurate 9DoF cuboid annotations. To reliably resolve this ambiguity, additional geometric cues are required—such as the known size of the vehicle, dimensions of identifiable features on the vehicle (e.g., license plates, tire diameter), or features on the surrounding ground plane (e.g., lane width, lane marking length). Incorporating such cues is essential for achieving high-precision 9DoF annotations.

\section{Conclusion and Future Work}
\label{sec:conclusion}
In this work, we introduced \emph{ToosiCubix}, a novel approach for 3D cuboid annotation that relies solely on a single monocular image and the camera's intrinsic calibration parameters. This method has the potential to significantly reduce the cost and complexity of data collection for autonomous driving datasets, removing the need for LiDAR or stereo camera systems. Moreover, it enables the generation of 3D annotations for existing 2D datasets, enhancing their utility in 3D perception tasks.

Numerous additional geometric cues can further constrain the 3D pose and dimensions of vehicles, presenting clear avenues for future improvement. One direction is to extend our method to video sequences, where object tracking can enforce temporal consistency across frames and yield more stable annotations. Another is to incorporate geometric relationships between vehicles, such as coplanarity constraints, as explored in \cite{song2019Apollocar3D}. Expanding the annotation capability to include a broader range of road users—such as two-wheelers, articulated buses, pedestrians, dumpsters, and traffic cones—would increase the practical applicability of our tool.

Furthermore, integrating semi-automatic landmark detection with user-guided interaction could greatly accelerate the annotation process. A detection model that identifies keypoints and symmetries could maintain accuracy while reducing manual workload. These features could also facilitate the training of supervised 3D vehicle detection models. Introducing learned priors for various vehicle types would offer additional benefits by improving both annotation precision and generalization.

A particularly exciting direction is exploring the use of geometric cues from our framework in weakly supervised 3D detection settings. This could enable high-quality 3D annotations with minimal reliance on fully annotated datasets, thereby reducing development costs while maintaining strong performance.

Our framework can also be used directly for 3D object detection. In this scenario, one would first apply a 2D keypoint detector to localize the relevant features on each vehicle, and then infer the full 3D cuboid using our geometric reasoning pipeline. This opens the door for a class of interpretable, geometry-based monocular 3D detectors that rely less on end-to-end learning and more on structured inference.

\bibliographystyle{IEEEtran}
\bibliography{refrences}
\begin{IEEEbiography}[{\includegraphics[width=1in,height=1.25in,clip,keepaspectratio]{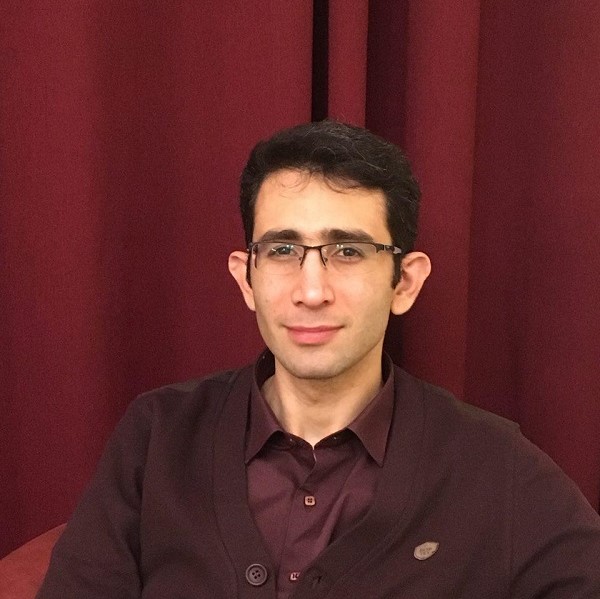}}]{Behrooz Nasihatkon}
	received his Ph.D. degree from the Australian National University in 2014. His dissertation focused on generalizing the theory of projective reconstruction in computer vision for the design and analysis of 3D reconstruction algorithms. He is currently an assistant professor at K. N. Toosi University of Technology, Tehran, Iran. His research interests include computer vision, multiple-view geometry, Advanced Driver Assistant Systems, and autonomous driving.
\end{IEEEbiography}

\begin{IEEEbiography}[{\includegraphics[width=1in,height=1.25in,clip,keepaspectratio]{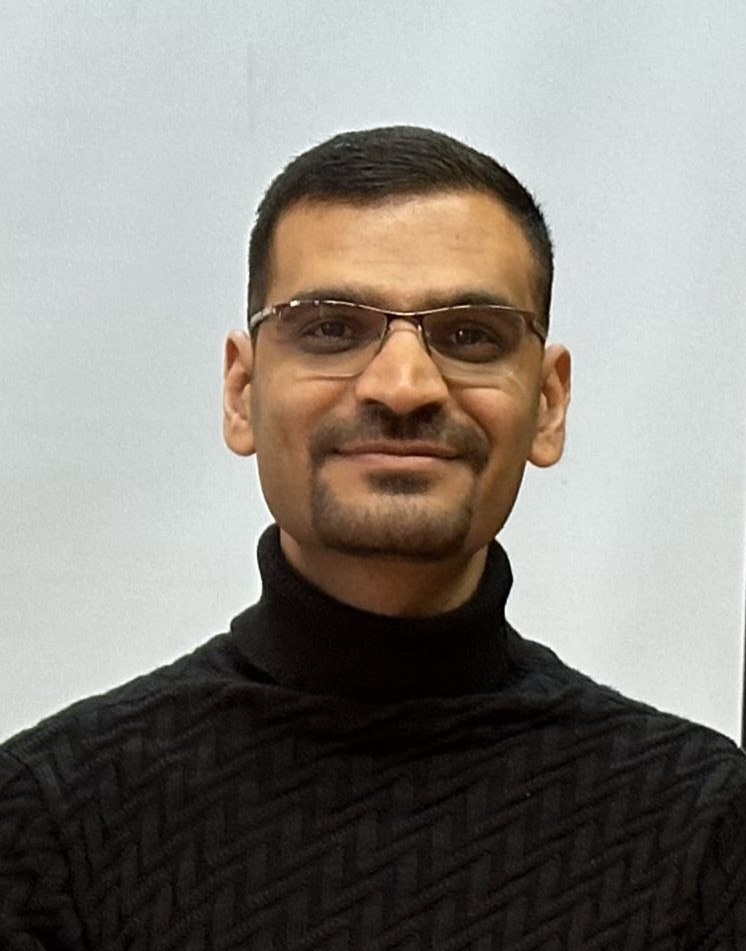}}]{Hossein Resani}
	Recived M.Sc. degree from the K. N. Toosi University of Technology in 2023. He has been collaborating with Behrooz Nasihatkon at the Computer Vision and 3D Geometry Group at K. N. Toosi University of Technology. His research interests include Continual Learning, 3D Computer Vision and Multimodal learning.    
\end{IEEEbiography}

\begin{IEEEbiography}[{\includegraphics[width=1in,height=1.25in,clip,keepaspectratio]{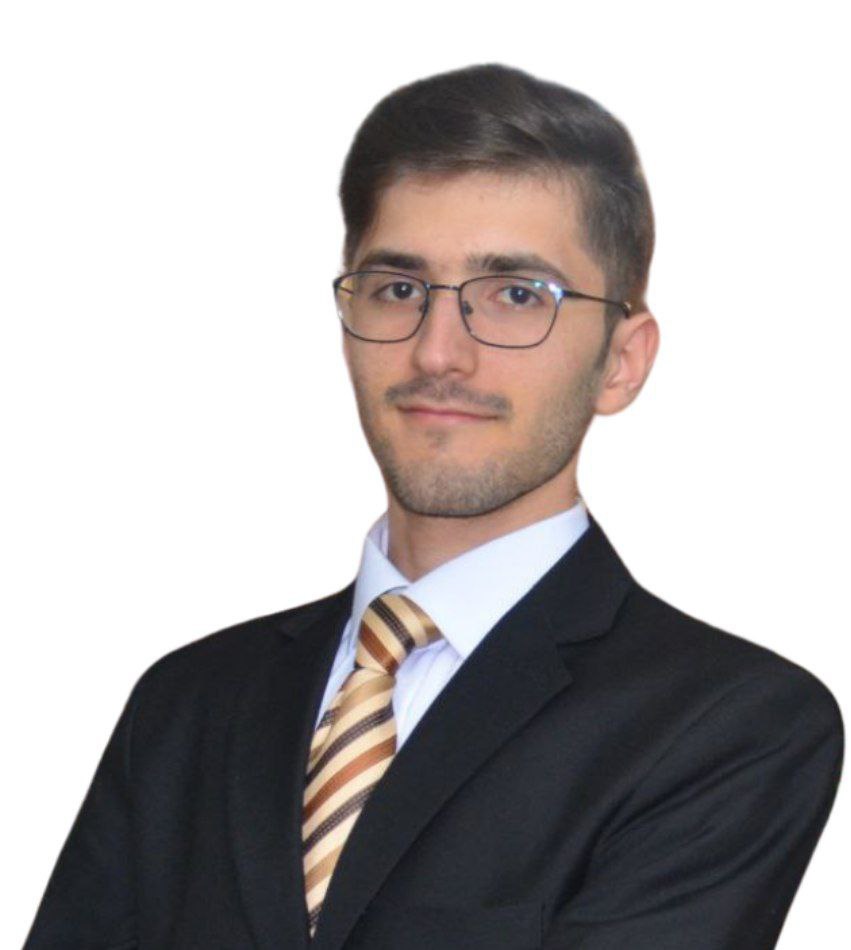}}]{Amirreza Mehrzadian}
	is an undergraduate student in Computer Engineering at K. N. Toosi University of Technology and a member of the Computer Vision and 3D Geometry Group under the supervision of Behrooz Nasihatkon. His research interests include computer vision, large multimodal models, and deep reinforcement learning.
\end{IEEEbiography}

\end{document}